\pdfoutput=1
\documentclass[10pt,compsoc,journal]{IEEEtran}

\usepackage[table, x11names, usenames,dvipsnames, cmyk]{xcolor}
\definecolor{salmon}{HTML}{EDF4F2} 
\usepackage{colortbl}
\usepackage{pgf}
\usepackage{highlight}
\usepackage{collcell}
\usepackage{ifthen}
\usepackage{xstring}

\usepackage{amsmath,amssymb,amsfonts}
\usepackage{array}
\usepackage{subcaption}
\usepackage{textcomp}
\usepackage{stfloats}
\usepackage{url}
\usepackage{verbatim}
\usepackage{graphicx}
\usepackage{cite}
\usepackage{longtable}

\usepackage[normalem]{ulem}
\useunder{\uline}{\ul}{}

\usepackage{bibentry}
\usepackage[numbers]{natbib}
\usepackage{xspace}

\newcommand{\tablecite}[2]{%
\ifcsname r@ht@#1\endcsname
  \hyperlink{cite:#1}{#2}~\cite{#1}%
\else
  #2~\cite{#1}
\fi
}

\makeatletter
\newcommand{\firstcite}[1]{%
  \ifcsundef{first@#1}{%
    \expandafter\gdef\csname first@#1\endcsname{}%
    \label{ht@#1}%
    \Hy@raisedlink{\hypertarget{cite:#1}{}}\cite{#1}%
  }{%
    \cite{#1}%
  }%
}
\makeatother

\usepackage{adjustbox}
\usepackage{multirow}

\usepackage{siunitx}
\usepackage{rotating}
\usepackage{booktabs}
\usepackage{enumitem} %
\usepackage{makecell}
\usepackage{wasysym} %

\def\MYTITLE{Event-based Stereo Depth Estimation: A Survey}

\usepackage{color}

\definecolor{eccvblue}{rgb}{0.12,0.49,0.85}
\usepackage[breaklinks=true,
    colorlinks=true,
    citecolor=eccvblue,
    bookmarksdepth=4
    bookmarksopen=true,
    bookmarksopenlevel=2
    ]{hyperref}
\hypersetup{
  pdftitle={\MYTITLE},
  pdfsubject={Computer Vision, Robotics, Deep Learning},
  pdfauthor={Suman Ghosh, Guillermo Gallego},
  pdfkeywords={Event Cameras, 3D reconstruction, Asynchronous Sensor, High Dynamic Range, High Temporal Resolution}
}

\usepackage[capitalize]{cleveref}
\newif\ifaisy
\aisyfalse %

\ifaisy
\crefname{section}{Section}{Sections}
\crefname{table}{Table}{Tables}
\crefname{figure}{Figure}{Figures}
\else 
\crefname{section}{Sec.}{Secs.}
\crefname{table}{Tab.}{Tabs.} 
\crefname{figure}{Fig.}{Figs.}
\fi
\Crefname{section}{Section}{Sections}
\Crefname{table}{Table}{Tables}
\Crefname{figure}{Figure}{Figures}

\newcommand\gframe[1]{{\color{lightgray}\frame{#1}}}

\newcommand{\revise}[1]{#1} %

\newcommand{\gred}[1]{#1} %
\newcommand{\gpass}[1]{#1}
\newcommand{\bg}[1]{#1}
\newcommand{\rquestion}[1]{#1}
\newcommand{\descript}[1]{#1}
\newcommand{\appl}[1]{#1}
\newcommand{\challeng}[1]{#1}
\newcommand{\opport}[1]{#1}
\newcommand{\findin}[1]{#1}
\newcommand{\trend}[1]{#1}
\newcommand{\proscons}[1]{#1}
\newcommand{\rammif}[1]{#1}
\newcommand{\organiz}[1]{#1}
\newcommand{\methodol}[1]{#1}

\def\bigbullet{\CIRCLE}
\def\bigcirc{\Circle}

\def\pol{p} %

\def\bx{\mathbf{x}}

\def\pol{p}

\NewDocumentCommand{\rot}{O{45} O{1em} m}{\makebox[#2][l]{\rotatebox{#1}{#3}}}%

\makeatletter
\long\def\@IEEEtitleabstractindextextbox#1{\parbox{0.922\textwidth}{#1}}
\makeatother

\usepackage{orcidlink}

\newif\ifarxiv
\arxivtrue %

\newif\ifclearlook

\newcommand{\cellcolorgradient}[1]{
    \pgfmathsetmacro{\value}{#1}
    \pgfmathsetmacro{\minvalue}{0}   %
    \pgfmathsetmacro{\maxvalue}{20} %
    \pgfmathsetmacro{\range}{\maxvalue-\minvalue}
    \pgfmathsetmacro{\percent}{(\value-\minvalue)/\range}
    \pgfmathsetmacro{\red}{1-\percent}
    \pgfmathsetmacro{\green}{\percent}
    \pgfmathsetmacro{\alpha}{0.1} %
    \xdef\tempcolor{\noexpand\cellcolor[rgb]{\red,\green,0}}
    \tempcolor #1
}

\ifarxiv
\usepackage[absolute]{textpos}
\fi

\begin{document}
\title{\MYTITLE}

\ifarxiv
\definecolor{somegray}{gray}{0.5}
\newcommand{\darkgrayed}[1]{\textcolor{somegray}{#1}}
\begin{textblock}{11}(2.5, 0.4)
\begin{center}
\darkgrayed{This paper has been accepted for publication at the\\
IEEE Transactions on Pattern Analysis and Machine Intelligence, 2025.
\copyright IEEE}
\end{center}
\end{textblock}
\fi

\author{Suman Ghosh$^{1}$\orcidlink{0000-0002-4297-7544}
and Guillermo Gallego$^{1,2}$\orcidlink{0000-0002-2672-9241}%
\IEEEcompsocitemizethanks{\IEEEcompsocthanksitem 
$^1$ TU Berlin and Robotics Institute Germany, Berlin, Germany. 
$^2$ Science of Intelligence Excellence Cluster and Einstein Center Digital Future, Berlin, Germany.
}%
}

\IEEEtitleabstractindextext{
\begin{abstract}
Stereopsis has widespread appeal in computer vision and robotics as it is the predominant way by which we perceive depth to navigate our 3D world. Event cameras are novel bio-inspired sensors that detect per-pixel brightness changes asynchronously, with very high temporal resolution and high dynamic range, enabling machine perception in high-speed motion and broad illumination conditions. The high temporal precision also benefits stereo matching, making disparity (depth) estimation a popular research area for event cameras ever since their inception. Over the last 30 years, the field has evolved rapidly, from low-latency, low-power circuit design to current deep learning (DL) approaches driven by the computer vision community. The bibliography is vast and difficult to navigate for non-experts due its highly interdisciplinary nature. Past surveys have addressed distinct aspects of this topic, in the context of applications, or focusing only on a specific class of techniques, but have overlooked stereo datasets. This survey provides a comprehensive overview, covering both instantaneous stereo and long-term methods suitable for simultaneous localization and mapping (SLAM), along with theoretical and empirical comparisons. It is the first to extensively review DL methods as well as stereo datasets, even providing practical suggestions for creating new benchmarks to advance the field. The main advantages and challenges faced by event-based stereo depth estimation are also discussed. Despite significant progress, challenges remain in achieving optimal performance in not only accuracy but also efficiency, a cornerstone of event-based computing. We identify several gaps and propose future research directions. We hope this survey inspires future research in depth estimation with event cameras and related topics, by serving as an accessible entry point for newcomers, as well as a practical guide for seasoned researchers in the community.

\end{abstract}

}
\maketitle

\IEEEdisplaynontitleabstractindextext

\ifarxiv
{
\hypersetup{linkcolor=blue}
\setcounter{tocdepth}{4}
{\scriptsize
\tableofcontents
}
}
\fi

\ifarxiv
\setcounter{section}{0}
\section*{Supplementary}
\label{sec:links}
Large tables (spreadsheets) on stereo \href{https://docs.google.com/spreadsheets/d/1DfmVXdg3H9iaLpkXNm5ygB6ald9dK0ggO0rUDXEDTXE}{methods}
and \href{https://docs.google.com/spreadsheets/d/1DfmVXdg3H9iaLpkXNm5ygB6ald9dK0ggO0rUDXEDTXE/edit#gid=1539773438&range=A1}{datasets}.
\cleardoublepage
\fi

\ifclearlook\cleardoublepage\fi \section{Introduction}

\bg{Depth estimation or 3D reconstruction from images acquired by traditional cameras is a topic of paramount interest in computer vision and robotics 
because it tries to mimic the same functionality of the human brain (i.e., inverting the operation of perspective projection) 
and has innumerable applications (since we operate in a 3D world).
}

\bg{
Event cameras \cite{Gallego20pami} are novel bio-inspired sensors that, mimicking the transient visual pathway of the human visual system, output pixel-wise intensity changes asynchronously instead of intensity frames at a fixed rate\footnote{An animation of the principle of operation of event cameras can be found here: \url{https://youtu.be/LauQ6LWTkxM?t=28}.}.
Since the seminal work \cite{Lichtsteiner08ssc} (2008) they have gained increasing interest due to their appealing properties, which allow them to perform well in challenging scenarios for traditional cameras, such as high-speed motion, high dynamic range (HDR) illumination, and low-power consumption.
Hence, recent years have witnessed how this hardware technology moved from university laboratories into startups that have been acquired or are participated by leading companies in the camera industry (SONY, OmniVision, Samsung, etc.).
The advantages of event cameras are being leveraged to tackle a variety of tasks in computer vision and robotics, such as optical flow estimation, pattern recognition, video synthesis, \revise{novel view synthesis}, AR/VR and SLAM
\cite{Gallego20pami}.
}
\ifarxiv \else \begin{figure}[t]
    \centering
    {\includegraphics[trim={0.5cm 1.2cm 0.3cm 1.5cm},clip,width=0.49\linewidth]{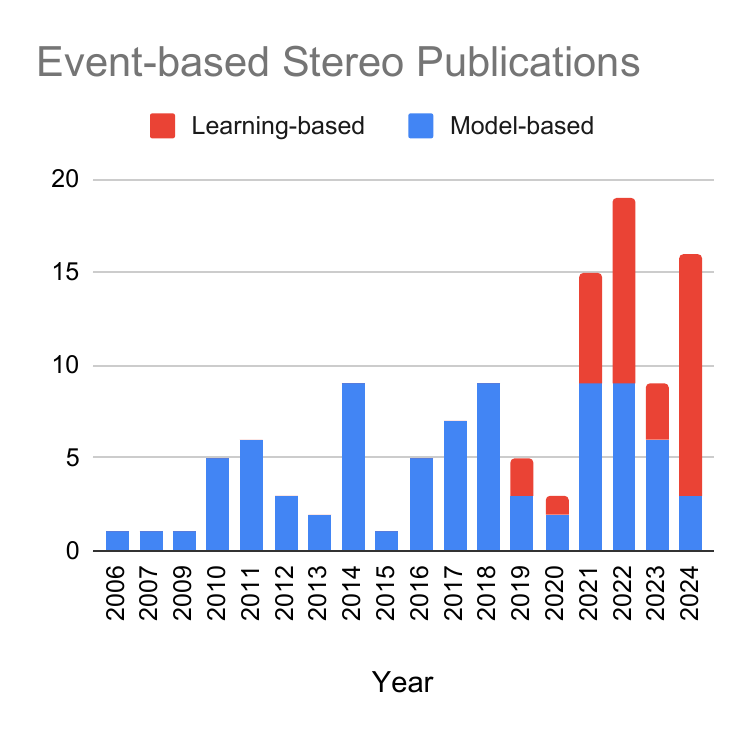}}
    {\includegraphics[trim={0.5cm 1.2cm 0.3cm 1.5cm},clip,width=0.49\linewidth]{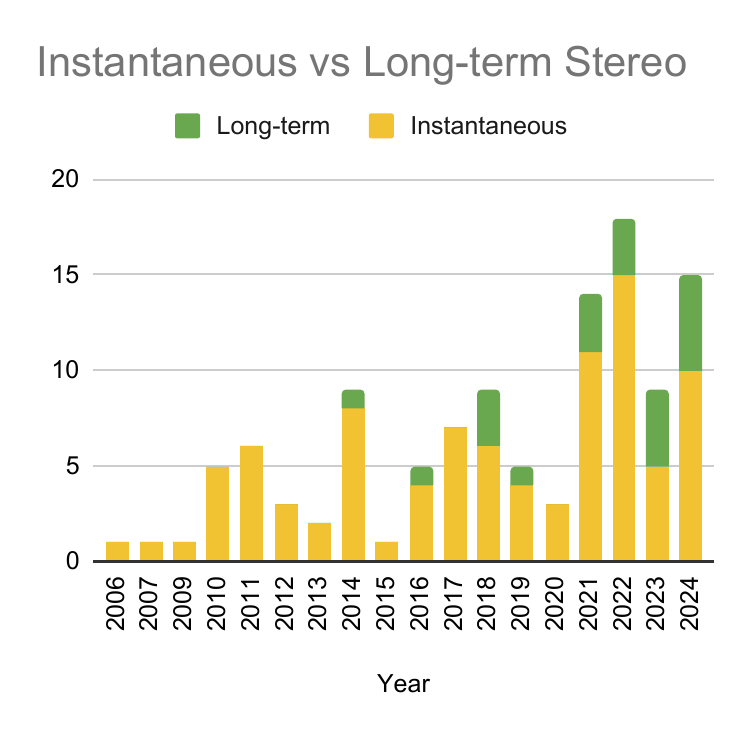}}
    \ifarxiv \else \vspace{-1ex} \fi 
    \caption{
    \gpass{
    Publications on event-based stereo depth estimation in the last two decades,
    classified according to criteria \#1: whether they are model-based (i.e., hand-crafted) or learning-based (i.e., data-driven) methods (left), and 
    \#2: whether they produce instantaneous depth outputs or have long-term motion consistency (right).
    Plots created from a \href{https://docs.google.com/spreadsheets/d/1DfmVXdg3H9iaLpkXNm5ygB6ald9dK0ggO0rUDXEDTXE}{compiled spreadsheet} of growing number of papers.
    }
    }
    \ifarxiv \else \vspace{-1ex} \fi
    \label{fig:publications:modelvslearning}
\end{figure}
 \fi

\trend{
The literature about event-based depth estimation is recent but continuously growing (\cref{fig:publications:modelvslearning} and \cref{tab:stereomethods}).}
\methodol{It can be organized according to diverse criteria, 
which are often related to the type of hardware setup, 
the assumptions or constraints on the scenario, and the target task (i.e., input/output), as in the columns of \cref{tab:stereomethods}.}
\begin{enumerate}
\item \methodol{The first categorization criterion is the number of event cameras considered: one (i.e., monocular) vs. stereo (or, in general, multi-camera).}

\item 
\methodol{Another classification criterion is whether depth \revise{is estimated over ``short'' or ``long'' intervals.
The latter requires knowledge of the camera motion (as an external input or concurrently estimated) for proper data assimilation.} 
This is different from whether the method has been tested on a moving platform.}

\item \methodol{A third criterion is whether the method is model-based (i.e., hand-crafted) or learning-based (i.e., data-driven). 
Early works fall in the former paradigm, whereas new approaches are dominated by the latter (see \cref{fig:publications:modelvslearning}).}

\item \methodol{A fourth criterion pertains to the type of output: 
depth on a per-event basis (i.e., at each asynchronous brightness change), 
or on a per-pixel basis at specified times (e.g., when using grid-like representations of events). 
Output can be dense (for every pixel) or semi-dense / sparse (e.g., at object contours or event data locations).
}
\end{enumerate}

\ifarxiv \begin{figure}[t]
    \centering
    {\includegraphics[trim={0.5cm 1.2cm 0.3cm 1.5cm},clip,width=0.49\linewidth]{images/2024/learning_v_model.pdf}}
    {\includegraphics[trim={0.5cm 1.2cm 0.3cm 1.5cm},clip,width=0.49\linewidth]{images/2024/instant_v_long.pdf}}
    \ifarxiv \else \vspace{-1ex} \fi 
    \caption{
    \gpass{
    Publications on event-based stereo depth estimation in the last two decades,
    classified according to criteria \#1: whether they are model-based (i.e., hand-crafted) or learning-based (i.e., data-driven) methods (left), and 
    \#2: whether they produce instantaneous depth outputs or have long-term motion consistency (right).
    Plots created from a \href{https://docs.google.com/spreadsheets/d/1DfmVXdg3H9iaLpkXNm5ygB6ald9dK0ggO0rUDXEDTXE}{compiled spreadsheet} of growing number of papers.
    }
    }
    \ifarxiv \else \vspace{-1ex} \fi
    \label{fig:publications:modelvslearning}
\end{figure}
 \fi 

\ifarxiv \else \begin{figure}[t]
    \centering
    {\includegraphics[trim={0 0.4cm 0 0cm},clip,width=0.8\linewidth]{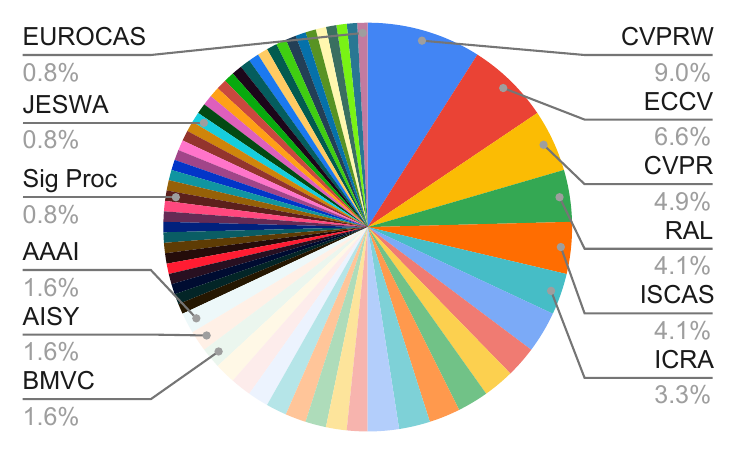}}
    \caption{Pie chart for publication venues of \gred{more than 135} publications in event-based stereo.}
    \label{fig:publications:venues}
\end{figure}
 \fi
\methodol{
Criterion \#2 can be formulated as ``instantaneous'' vs. ``long temporal baseline'' stereo depth estimation. 
``Instanteneous'' refers to stereo methods that estimate depth using only the event data available in a short time interval,
for which no information about camera motion (other than the extrinsic calibration) is needed. 
Thus, they can be used generally for any scenario involving two synchronized event cameras.}
\proscons{They shine in scenarios where the cameras are stationary and observing independently moving objects.
}

\methodol{
On the other hand, ``long temporal baseline'' stereo refers to cases involving ego-motion where the scene depth is estimated more accurately by fusing multiple consecutive depth observations of the same 3D structure. 
Such fusion requires knowledge of the camera motion to properly associate observations.} 
\findin{Using camera motion for temporal aggregation, the depth estimated by these methods is more accurate and consistent over time.
This is for example the case of visual odometry (VO) and SLAM}, \bg{which generate a reliable map of the scene for accurate localization, a fundamental technology enabling autonomous robot navigation.}

\proscons{
The temporal duration of the data aggregated and the knowledge (or absence) of the camera motion are related to the dynamicity of the scene: 
``instantaneous'' methods tend to handle better dynamic scenes containing independent motion, whereas ``long time baseline'' methods are better suited for persistent, i.e., stationary, parts of the scene.
Dynamicity is not binary; it depends on the relative speed difference between the camera and the scene (e.g., ``long time baseline'' methods may work well for parts of the scene that move little compared to the camera motion).}
\trend{In the literature, we observe that only a small portion of the stereo methods deal with the additional complexity of temporal aggregation that is suitable for SLAM (\cref{fig:publications:modelvslearning} \gred{right}).
}

\methodol{The above criteria represent orthogonal axes of variation. 
However, a linear order is needed to present the literature;
hence we focus on the stereo methods (criterion \#1) 
and follow the structure indicated by criterion \#2 (type of problem addressed), with subsections by criterion \#3 (see \cref{sec:stereomethods}).
We comment on the type of output (criterion \#4) when needed.
}

\findin{
Finally, our analysis of the literature in \cref{fig:publications:venues} shows that, due its interdisciplinary nature,
event-based stereo methods have been published in multiple venues of diverse scope (vision, robotics, neuroscience, machine learning, circuit design, instrumentation, etc.), without a dominant publication venue.} 
\trend{This trend is changing recently, as the computer vision and robotics communities are concentrating the latest efforts on depth estimation, which coincides with a time when the technology has become more widespread (and more research labs are jumping into it), along with the increased availability of several public datasets for benchmarking and training neural networks. %
}
\ifarxiv \begin{figure}[t]
    \centering
    {\includegraphics[trim={0 0.4cm 0 0cm},clip,width=0.8\linewidth]{images/2024/venue.pdf}}
    \caption{Pie chart for publication venues of \gred{more than 135} publications in event-based stereo.}
    \label{fig:publications:venues}
\end{figure}
 \fi

\bg{\emph{Remark: Depth and disparity} are often used interchangeably.
While they are different (depth refers to the component of the 3D scene along the camera's optical axis ($Z$) and disparity refers to the image-based displacement between the projections of a 3D point), 
they are related by the camera parameters and geometric configuration.
In canonical stereo configuration with cameras of focal length $f$ separated by a baseline distance $b$, 
depth can be computed using the formula $Z = (b \cdot f) / \Delta x$, 
where $\Delta x$ is the disparity.
}

\ifclearlook\cleardoublepage\fi \section{Previous surveys on the topic}
\ifarxiv \else

\begin{table}[t]
\centering
\caption{
\gpass{
Surveying the review papers on event-based stereo depth estimation.
This work is the last row.
}
}
\label{tab:surveys}
\begin{adjustbox}{max width=\linewidth}
\setlength{\tabcolsep}{2pt}
\begin{tabular}{llccccc}
\toprule
  \textbf{Ref}. & \textbf{Topic} & \textbf{Inst.} &\textbf{Long-} & \textbf{Model} & \textbf{Learning} & \textbf{Dataset}\\
  & & & \textbf{term} & \textbf{based} & \textbf{based} & \textbf{review} \\  
  \midrule
  
  2019 \cite{Steffen19fnbot} & Bio-inspired stereo vision with event & \CIRCLE & \Circle & \CIRCLE & \Circle & \Circle \\
  & cameras; focus on cooperative networks. \\[0.5ex]
  
  2022 \cite{Gallego20pami} & General survey on event-based vision. & \CIRCLE & \CIRCLE & \CIRCLE & \Circle & \Circle \\[0.7ex]

  2022 \cite{Furmonas22sensors} & Event-based monocular and stereo depth. & \CIRCLE & \Circle & \CIRCLE & \Circle & \Circle\\[0.7ex]
  
  2023 \cite{Zheng23arxiv} & Deep-learning--based event-vision \\
  & applications; benchmarking on image & \CIRCLE & \Circle & \Circle & \CIRCLE & \Circle \\
  & reconstruction and optical flow.\\[0.7ex]

  2023 \cite{huang23arxiv} & Event-based SLAM.    & \Circle & \CIRCLE & \CIRCLE & \Circle & \Circle \\[0.7ex]
  
  2025 & Event-based stereo methods and datasets. & \CIRCLE & \CIRCLE & \CIRCLE & \CIRCLE & \CIRCLE \\
    \bottomrule
\end{tabular}
\end{adjustbox}
\vspace{-2ex}
\end{table}
 \fi

\revise{\trend{
Due to the relevance of the topic, several survey articles have been published recently in the event-based vision literature about stereo depth estimation.}}
\methodol{They are summarized in \cref{tab:surveys}, indicating whether they cover instantaneous (Inst.) or long-term, model-based or learning-based methods.
}

\descript{
Steffen et al \cite{Steffen19fnbot} review various bio-inspired aspects of event cameras and the stereo algorithms that may benefit from their novel data. 
The article focuses on neuromorphic strategies for depth estimation, particularly revolving around cooperative stereo networks and their variants. 
Concurrent work, albeit published later, is the survey about Event-based Vision by Gallego et al \cite{Gallego20pami}. 
It comprehensively collates works in event-based literature from all aspects of event-based processing, applications as well as provides details about the various sensors available in the market. 
It also discusses existing literature in 3D monocular and stereo reconstruction in one of its sections (in about a 1.5 pages).
}

\descript{
Recently, Furmonas et al published a review article \cite{Furmonas22sensors} that dives into various event-based depth estimation methods, both monocular and stereo. 
A significant portion of the article is spent on discussing hardware-based efficient implementations of event-based processing for depth estimation.} 
\proscons{Although recent deep learning methods for monocular depth estimation are presented, only one learning-based stereo method is mentioned.
}

\descript{
A survey on deep learning methods for event-based vision by Zheng et al. \cite{Zheng23arxiv} dedicates a section to stereo depth estimation methods. 
It discusses learning-based methods that use stereo events, as well as methods using both events and intensity images.} 
\proscons{However, the primary focus of \cite{Zheng23arxiv} is on learning methods for image reconstruction and optical flow estimation.}
\descript{Also, a survey on SLAM using event cameras by Wang et al. \cite{huang23arxiv} discusses depth estimation and camera tracking using monocular and stereo setups;} 
\proscons{the main focus is on the monocular setup. 
\revise{Among stereo methods, only long-term ones are discussed.}
}

\ifarxiv

\begin{table}[t]
\centering
\caption{
\gpass{
Surveying the review papers on event-based stereo depth estimation.
This work is the last row.
}
}
\label{tab:surveys}
\begin{adjustbox}{max width=\linewidth}
\setlength{\tabcolsep}{2pt}
\begin{tabular}{llccccc}
\toprule
  \textbf{Ref}. & \textbf{Topic} & \textbf{Inst.} &\textbf{Long-} & \textbf{Model} & \textbf{Learning} & \textbf{Dataset}\\
  & & & \textbf{term} & \textbf{based} & \textbf{based} & \textbf{review} \\  
  \midrule
  
  2019 \cite{Steffen19fnbot} & Bio-inspired stereo vision with event & \CIRCLE & \Circle & \CIRCLE & \Circle & \Circle \\
  & cameras; focus on cooperative networks. \\[0.5ex]
  
  2022 \cite{Gallego20pami} & General survey on event-based vision. & \CIRCLE & \CIRCLE & \CIRCLE & \Circle & \Circle \\[0.7ex]

  2022 \cite{Furmonas22sensors} & Event-based monocular and stereo depth. & \CIRCLE & \Circle & \CIRCLE & \Circle & \Circle\\[0.7ex]
  
  2023 \cite{Zheng23arxiv} & Deep-learning--based event-vision \\
  & applications; benchmarking on image & \CIRCLE & \Circle & \Circle & \CIRCLE & \Circle \\
  & reconstruction and optical flow.\\[0.7ex]

  2023 \cite{huang23arxiv} & Event-based SLAM.    & \Circle & \CIRCLE & \CIRCLE & \Circle & \Circle \\[0.7ex]
  
  2025 & Event-based stereo methods and datasets. & \CIRCLE & \CIRCLE & \CIRCLE & \CIRCLE & \CIRCLE \\
    \bottomrule
\end{tabular}
\end{adjustbox}
\vspace{-2ex}
\end{table}
 \fi
\rquestion{
As \cref{tab:surveys} shows, our survey covers the \emph{entire body of work of event-based stereo} depth estimation, 
both instantaneous and long-term, %
model-based and learning-based. 
We not only describe the methods, but also benchmark their performance on common datasets.
Moreover, we also survey the related datasets due to their importance in advancing the field, as data fuels the development of learning-based methods, which is becoming the dominant paradigm.
}

\ifclearlook\cleardoublepage\fi \section{Event Camera Working Principle}

\bg{
Over the years, several event camera designs have been investigated inspired by biological retinas. 
Three main designs are presented in \cite{Posch14ieee}: the Dynamic Vision Sensor (\textbf{DVS}), the Dynamic and Active-Pixel Vision Sensor (\textbf{DAVIS}) and the Asynchronous time-based image sensor (\textbf{ATIS}).
They produce so-called ``change-detection'' events $e_k \doteq (\bx_k,t_k,\pol_k)$ as soon as the logarithmic brightness at a pixel $\bx_k = (x_k,y_k)^\top$ changes by a predefined amount $C$ (i.e., 20\%) \cite{Gallego20pami}:
\begin{equation}
    L(\bx,t_k) - L(\bx,t_k - \Delta t_k) = \pol_k C,
    \label{eq:EGM}
\end{equation}
where the event polarity $\pol_k \in \left\{ +1,-1 \right\}$ indicates the sign of the brightness change, and $\Delta t_k$ is the time elapsed since the last event at the same pixel.
}

\proscons{
This bio-inspired working principle confers advantages (low latency, HDR, low-power consumption, temporal redundancy suppression, minimal motion blur, etc.)
and poses challenges, such as dealing with noise, the lack of absolute brightness information, and the unfamiliar asynchronous nature of event data.}
\findin{
Hence, new algorithms are needed to leverage the advantages of event cameras \cite{Gallego20pami}. 
Among them, stereo algorithms play a prominent role, as they have been explored ever since the first cameras where prototyped by pioneer M.~Mahowald in the nineties (\cref{fig:stereosystem:Mahowald}).}
\bg{See \cite{Posch14ieee,Gallego20pami} for more details on the working principle of event cameras.
}

\ifarxiv \else \begin{table}[!ht]
\setlength\extrarowheight{1.22pt}
\centering
\caption{\label{tab:stereomethods}
\methodol{
Event-based stereo methods.
The columns indicate whether the sensor rig moves (``Ego''-motion) or not, 
the method is instantaneous (I) or long-term (LT),
output depth (or disparity) is sparse (S) or dense (D),
the method is model-based (M) or learning-based (L),
learning is supervised (SL) or unsupervised (UL), 
and the type of sensors used (E = event camera, F = frame-based camera, 
train = at training time).
\href{https://tinyurl.com/4vywjm55}{Link to spreadsheet with stereo methods}.
}
}
\rowcolors{2}{salmon}{white}
\begin{adjustbox}{max width=0.99\linewidth}
\begin{tabular}{llrllllll}
\toprule
\textbf{Methods}  & \textbf{Venue}  & \textbf{Year}  & \textbf{Ego?}  & \textbf{I/LT}  & \textbf{S/D}  & \textbf{L/M}  & \textbf{SL/UL}  & \textbf{Sensors} \\
\midrule 
\tablecite{deOliveira25derdnet}{DERD-Net} & arXiv &2025 &\CIRCLE  & LT  &S &L &SL &2E, 1E\\
\tablecite{Niu24esvo2}{ESVO2} &TRO &2025 &\CIRCLE  & LT  &S &M & &2E + IMU \\
\tablecite{Jiang24tim}{EV-MGDispNet} & TIM &2025 &\CIRCLE &I &D &L &SL &2E \\
\tablecite{Lou24neurips}{ZEST} & NeurIPS &2024 &\CIRCLE  & I  &D &L &UL &1E (+1F train) \\
\tablecite{Ghosh24eccvw}{ES-PTAM} &ECCVW &2024 &\CIRCLE  & LT  &S &M & &2E (N-ocular) \\
\tablecite{Zhao24eccv}{Zhao et al.} & ECCV & 2024 &\CIRCLE &I &D &L &SL &2E + 2F\\
\tablecite{Bartolomei24eccv}{Bartolomei et al.} & ECCV & 2024 &\CIRCLE &I &D &L &SL &2E + LiDAR\\
\tablecite{Cho24eccv}{StereoFlow-Net} &ECCV &2024 &\CIRCLE &I &D &L &SL &2E \\
\tablecite{Shiba24pami}{Shiba et al. } &TPAMI &2024 &\CIRCLE  & LT  &D &M & &2E \\
\tablecite{Chen24spl}{Chen et al. } &SPL &2024 &\CIRCLE &I &D &L &SL &2E \\
\tablecite{Ghosh24jeswa}{Ghosh et al. } &JESWA &2024 &\CIRCLE &I &D &L &SL &2E \\
\tablecite{Chen24wacv}{SAFE } &WACV &2024 &\CIRCLE  & LT  &D &L &SL &1E + 1F \\
\tablecite{Soliman24tiv}{DH-PTAM} & TIV &2024 &\CIRCLE  & LT  &S &M + L &SL &2E + 2F \\
\tablecite{Ding24tmm}{SEVFI-Net} &TMM &2024 &\CIRCLE &I &D &L &SL &1E + 1F \\
\tablecite{Jianguo23iecon}{ASNet } &IECON &2023 &\CIRCLE &I &D &L &SL &2E \\
\tablecite{Elmoudni23itsc}{El Moudni et al. } &ITSC &2023 &\CIRCLE  & LT  &S &M & &2E \\
\tablecite{Liu23aisy}{T-ESVO } &AISY &2023 &\CIRCLE  & LT  &S &M & &2E \\
\tablecite{Cho23cvpr}{ADES } &CVPR &2023 &\CIRCLE &I &D &L &UL &2E (+2F train) \\
\tablecite{Liu23sensors}{ESVIO} (direct) &Sensors &2023 &\CIRCLE  & LT  &S &M & &2E + IMU \\
\tablecite{Lin25tcsvt}{St-EDNet} &arXiv &2023 &\CIRCLE &I &D &L &UL &1E + 1F \\
\tablecite{Chen23ral}{ESVIO} (indirect) &RAL &2023 &\CIRCLE  & LT  &S &M & &2E (+ 2F) + IMU \\
\tablecite{Zhang22eccv2}{Zhang et al. } &ECCV &2022 &\CIRCLE &I &D &L &SL &1E + 1F \\
\tablecite{Cho22eccv}{SCS-Net} &ECCV &2022 &\CIRCLE &I &D &L &SL &2E + 2F \\
\tablecite{Uddin22tcsvt}{N. Uddin et al. } &TCSVT &2022 &\CIRCLE &I &D &L &UL &2E + 2F \\
\tablecite{Ghosh22aisy}{MC-EMVS} &AISY &2022 &\CIRCLE  & LT  &S &M & &2E \gred{(N-ocular)} \\
\tablecite{Nam22cvpr}{Conc-Net } &CVPR &2022 &\CIRCLE &I &D &L &SL &2E (+ 2F optional) \\
\tablecite{Zhang22cvpr}{DTC-SPADE} &CVPR &2022 &\CIRCLE &I &D &L &SL &2E \\
\tablecite{Liu22icra}{Liu et al. } &ICRA &2022 &\CIRCLE &I &S &M + L &UL &2E \\
\tablecite{Ilani22iaoc}{Ilani et al.} &IAOC &2022 &\Circle &I &S &M & &1E \\
\tablecite{Kim22access}{Kim et al. } &Access &2022 &\Circle &I &S &M & &2E \\
\tablecite{Gu22jist}{SIES} &JIST &2022 &\CIRCLE &I &D &L &UL &1E + 1F \\
\tablecite{Kim22ral}{HSM} &RAL &2022 &\CIRCLE &I &S &M & &1E + 1F \\
\tablecite{rancon22access}{StereoSpike} &Access &2022 &\CIRCLE &I &D &L &SL &2E \\
\tablecite{Ghosh22srep}{3D-saliency} &Sci. Rep. &2022 &\CIRCLE &I &S &M & &2E \\
\tablecite{Wang21iros}{SHEF} &IROS &2021 &\CIRCLE &I &S &M + L &SL &1E + 1F \\
\tablecite{Ahmed21aaai}{EIT-Net } &AAAI &2021 &\CIRCLE &I &D &L &SL &2E \\
\tablecite{Zhou20tro}{ESVO} &TRO &2021 &\CIRCLE  & LT  &S &M & &2E \\
\tablecite{Mostafavi21iccv}{EIS } &ICCV &2021 &\CIRCLE &I &D &L &SL &2E + 2F \\
\tablecite{Risi21iscas}{Risi et al.} &ISCAS &2021 &\Circle &I &S &M & &2E \\
\tablecite{Zuo21iros}{HDES} &IROS &2021 &\CIRCLE &I &D &L &SL &1E + 1F \\
\tablecite{Muglikar21threedv}{ESL} &3DV &2021 &\Circle &I &D &M & & 1E + light projector\!\!\!\\
\tablecite{Kweon21cvprw}{CES-Net} &CVPRW &2021 &\CIRCLE &I &S &L &SL &2E \\
\tablecite{Hadviger21advro}{Hadviger et al.} &Adv. Rob. &2021 &\CIRCLE &I &S &M & &1E + 2F \\
\tablecite{Risi20fnbot}{Risi et al.} &FNBOT &2020 &\Circle &I &S &M & &2E \\
\tablecite{Wang20eccv}{Wang et al.} &ECCV &2020 &\Circle &I &S &M & &2E \\
\tablecite{Tulyakov19iccv}{DDES} &ICCV &2019 &\CIRCLE &I &D &L &SL &2E \\
\tablecite{Hadviger19ecmr}{Hadviger et al. } &ECMR &2019 &\Circle &I &S &M & &2E \\
\tablecite{Kaiser18icann}{Kaiser et al.} &ICANN &2018 &\CIRCLE &I &S &M & &2E \\
\tablecite{Everding18thesis}{Everding} &Ph.D. Thesis &2018 &\CIRCLE &I &S &M & &2E \\
\tablecite{Andreopoulos18cvpr}{Andreopoulos et al.} &CVPR &2018 &\Circle &I &S &M & &2E \\
\tablecite{Ieng18fnins}{GTS} &FNINS &2018 &\Circle &I &S &M & &2E \\
\tablecite{Zhu18eccv}{TSES} &ECCV &2018 &\CIRCLE  & LT  &S &M & &2E \\
\tablecite{Zhou18eccv}{Zhou et al.} &ECCV &2018 &\CIRCLE  & LT  &S &M & &2E \\
\tablecite{Xie18ijars}{SGM} &IJARS &2018 &\Circle &I &S &M & &2E \\
\tablecite{CamunasMesa18tnnls}{Camuñas-Mesa et al.} &TNNLS &2018 &\Circle &I &S &M & &2E \\
\tablecite{Eibensteiner17radioe}{Eibensteiner et al.} &RADIOELEK &2017 & \Circle &I & S &M & & 2E \\
\tablecite{Piatkowska17cvprw}{Piatkowska et al. } &CVPRW &2017 &\Circle &I &S &M & &2E \\
\tablecite{Dikov17cbbs}{Dikov et al.} &CBBS &2017 &\Circle &I &S &M & &2E \\
\tablecite{Osswald17srep}{Osswald et al.} &Sci. Rep. &2017 &\Circle &I &S &M & &2E \\
\tablecite{Zou17bmvc}{Zou et al.} &BMVC &2017 &\CIRCLE &I &D &M & &2E \\
\tablecite{Xie17fns}{BPN} &FNINS &2017 &\Circle &I &S &M & &2E \\
\tablecite{Ieng17fnins}{Ieng et al.} &FNINS &2017 &\Circle &I &S &M & &2E \\
\tablecite{kogler17thesis}{Kogler } &Ph.D. Thesis &2016 &\Circle &I &S &M & &2E \\
\tablecite{Firouzi16npl}{Firouzi et al.} &NPL &2016 &\Circle &I &S &M & &2E \\
\tablecite{Zou16icip}{Zou et al.} &ICIP &2016 &\CIRCLE &I &S &M & &2E \\
\tablecite{Schraml16tie}{Scharml et al.} &TIE &2016 &\CIRCLE &I &S &M & &2E \\
\tablecite{Schraml15cvpr}{Scharml et al.} &CVPR &2015 &\CIRCLE &I &S &M & &2E \\
\tablecite{CamunasMesa14biocas}{Camuñas-Mesa et al.} &BioCAS &2014 &\Circle &I &S &M & &2E \\
\tablecite{Eibensteiner14cvprw}{Eibensteiner et al.} &CVPRW &2014 &\Circle &I &S &M & &2E \\
\tablecite{carneiro14thesis}{Carneiro} &Ph.D. Thesis &2014 &\Circle &I &S &M & &3E (N-ocular) \\
\tablecite{Lee14tnnls}{Lee et al.} &TNNLS &2014 &\Circle &I &S &M & &2E \\
\tablecite{Piatkowska14msci}{Piatkowska et al.} &M. Sci. Tech. &2014 &\Circle &I &S &M & &2E \\
\tablecite{CamunasMesa14fns}{Camuñas-Mesa et al.} &FNINS &2014 &\Circle &I &S &M & &2E \\
\tablecite{CamunasMesa14iscas}{Camuñas-Mesa et al.} &ISCAS &2014 &\Circle &I &S &M & &2E \\
\tablecite{Kogler14jei}{Kogler et al.} &JEI &2014 &\gred{\Circle} &I &\gred{S} &M & &2E \\
\tablecite{Carneiro13nn}{Carneiro et al.} &NN &2013 &\Circle &I &S &M & &\gred{3E (N-ocular)} \\
\tablecite{Piatkowska13iccvw}{Piatkowska et al.} &ICCVW &2013 &\Circle &I &S &M & &2E \\
\tablecite{Litzenberger12wcits}{CARE} &WCITS &2012 &\Circle &I &S &M & &2E \\
\tablecite{humenberger12cvprw}{Humenberger et al.} &CVPRW &2012 &\Circle &I &S &M & &2E \\
\tablecite{Rogister12tnnls}{Rogister et al. } &TNNLS &2012 &\Circle &I &S &M & &2E \\
\tablecite{Kogler11isvc}{Kogler et al. } &ISVC &2011 &\Circle &I &S &M & &2E \\
\tablecite{Kogler11book}{Kogler et al.} &ATASV &2011 &\Circle &I &S &M & &2E \\
\tablecite{sulzbachner11cvprw}{Sulzbachner et al.} &CVPRW &2011 &\Circle &I &S &M & &2E \\
\tablecite{eibensteiner11eurocast}{Eibensteiner et al.} &EUROCAST &2011 &\Circle &I &S &M & &2E \\
\tablecite{belbachir11cvprw}{Belbachir et al.} &CVPRW &2011 &\Circle &I &S &M & &2E \\
\tablecite{Benosman11tnn}{Benosman et al.} &TNNLS &2011 &\Circle &I &S &M & &2E \\
\tablecite{Schraml10iscas}{Schraml et al.} &ISCAS &2010 &\Circle &I &S &M & &2E \\
\tablecite{schraml10cvprw2}{Schraml et al.} &CVPRW &2010 &\Circle &I &S &M & &2E \\
\tablecite{belbachir10cvprw}{Belbachir et al.} &CVPRW &2010 &\Circle &I &S &M & &2E \\
\tablecite{sulzbachner10elmar}{Sulzbachner et al.} &ELMAR &2010 &\Circle &I &S &M & &2E \\
\tablecite{Kogler09icvs}{Kogler et al.} &ICVS &2009 &\Circle &I &S &M & &2E \\
\tablecite{Schraml07visapp}{Schraml et al.} &VISAPP &2007 &\Circle &I &S &M & &2E \\
\tablecite{Hess06thesis}{Hess} &Sem. Thesis &2006 &\Circle &I &S &M & &2E \\
\tablecite{Mahowald92thesis}{Mahowald } &Ph.D. Thesis &1992 &\Circle &I &S &M & &2E \\
\bottomrule
\end{tabular}
\end{adjustbox}
\end{table}
 \fi 

\ifclearlook\cleardoublepage\fi \section{Stereo (and Multi-camera) Methods}
\label{sec:stereomethods}
\begin{figure}[t]
    \centering
    \includegraphics[trim={0 0 0 100px},clip,width=.6\linewidth]{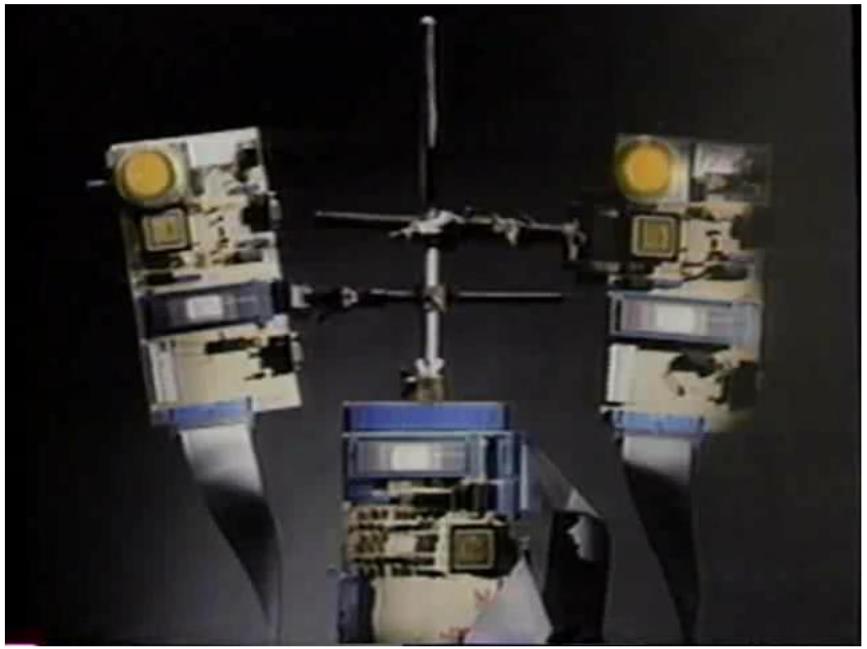} 
    \caption{
    \descript{
    The first event-based stereo matching system, developed in Michelle (Misha) Mahowald's 
    PhD thesis \cite{Mahowald92thesis,Mahowald89vlsi,Mahowald94book}. 
    Binocular silicon retinas (in yellow, early prototypes of event cameras) were used to estimate depth with analog circuits (the stereo processing chip is at the bottom).
    Image adapted from \href{https://vimeo.com/109984367}{Silicon Vision}.
    }
    }
    \label{fig:stereosystem:Mahowald}
\end{figure}

\descript{
This section presents algorithms for depth estimation using multiple event and/or frame-based cameras in stereo configuration (\cref{tab:stereomethods}).}
\findin{
Most works follow the classical (i.e., frame-based) paradigm that decouples the stereo depth estimation problem in two sequential steps:
first, establishing stereo correspondences across image planes (``stereo matching''), 
and then back-projecting the correspondences to compute the associated 3D point (``triangulation'') \cite{Trucco98book}.
}
\bg{
The first subproblem (stereo matching) is more difficult than the second one. 
The geometric configuration of the problem is often exploited to avoid searching for correspondences over the entire image plane: the epipolar constraint reduces the search to the epipolar line \cite{Hartley03book}.} 
\findin{
Additionally, due to unique properties of event cameras, such as their spatially sparse and temporally quasi-continuous output, pixels can be \emph{stereo-matched using precise event timestamps}. 
The usual assumption is that a moving object triggers simultaneous / co-occurrent / coincident events on both camera views.}
\trend{
This idea has inspired many approaches ever since the inception of event cameras (\cref{fig:stereosystem:Mahowald}).
}

\ifarxiv  \fi

\ifclearlook\cleardoublepage\fi \subsection{Instantaneous Stereo}
\label{sec:stereomethods:inst}

\organiz{
This section deals with scenarios where a time-evolving depth map is estimated using only the most recent stereo events, without any explicit knowledge of camera motion.}
\descript{
These methods can be used to estimate depth of independently moving objects in the scene along with the static parts.}
\organiz{
We first discuss model-based and then more recent deep-learning--based techniques.
}

\subsubsection{Model-based Methods}
\label{sec:stereomethods:inst:model}

\paragraph{\textbf{Time-based Stereo Matching}}
\findin{The main line of research in event-based stereo matching is focused on \emph{temporal matching}.}
\trend{Initial works~\firstcite{Schraml07visapp,Kogler09icvs,Schraml10iscas,DominguezMorales11icspma,Belbachir12iscas} accumulated events into %
frames for compatibility with standard binocular vision techniques.}
\proscons{However, temporal information is quantized during accumulation.}
\descript{\challeng{To emphasize the benefits of accurate time information},~\firstcite{Kogler11isvc} proposed a purely event-driven matching procedure using time and polarity-based correlation of the events, without aggregation. 
}

\proscons{
However, matching events solely based on time is not reliable because of inherent \emph{ambiguities}} \bg{(e.g., the visual stimuli may generate multiple events simultaneously on multiple cameras) and noise (e.g., jitter delay~\firstcite{Muglikar21threedv}). 
The former happens when several objects move in the scene on overlapping epipolar lines \firstcite{Benosman11tnn}.} 
\challeng{To reduce false stereo correspondences} and therefore enhance matching quality, temporal matching was aided by \emph{additional cues} 
such as epipolar and ordering constraints~\firstcite{Rogister12tnnls}. %
\descript{Carneiro et al.~\firstcite{Carneiro13nn} demonstrated that adding cameras to the stereo setup 
(N-ocular vision) 
could disambiguate and produce more reliable matches.
}

\descript{
To further remove noise and ambiguities, a generalized framework for time-based stereo event matching (\textbf{GTS}) was proposed in~\firstcite{Ieng18fnins}, by combining four types of constraints: epipolar, temporal, luminance and motion. 
The \emph{luminance constraints} are applicable for ATIS event cameras, %
which also output intensity-encoded (i.e., exposure-measurement ``EM'') events.}
\proscons{
However, the low quality of EM events, especially in dark regions, means that they are hardly used in event-based processing. 
\bg{In fact, they have been abandoned after Prophesee's 3rd generation camera model~\firstcite{Gallego20pami}.}
}

\descript{The \emph{motion constraint} in GTS is implemented by matching patches of \emph{time surfaces} %
across cameras\ifarxiv ~(\cref{fig:GTS:motionconstratints})\fi.
} 
\bg{A time surface is a grid-like representation where each pixel stores the timestamp of its latest event~\firstcite{Lagorce17pami}. 
The trails of object edges in a time surface depict historical footprints of how that object moved.} %
\descript{If the object underwent similar motions on both camera views, we observe similar edge trails in time surfaces.} 
\findin{Similar to how frame-based stereo relies on the principle of photometric consistency across views to establish matches, 
event-based stereo leverages the \emph{time-consistency principle}~\firstcite{Muglikar21threedv} using similarity time surfaces.
Time surface matching is also used for finding stereo matches in event-based methods for stereo VO/SLAM like~\cite{Zhou18eccv,Zhou20tro}.}

\descript{
\emph{Belief propagation networks} (\textbf{BPN}) were proposed as an alternative way to extend the influence of epipolar and temporal co-incidence constraints~\firstcite{Xie17fns},}
\findin{where the goal was to improve matching by making information consistent over larger (``global'') space-time neighborhoods, as opposed to exploiting only local information.}  
\descript{Camuñas-Mesa et al.~\firstcite{CamunasMesa18tnnls} used time coincidence to estimate \emph{object-level} disparity and track 3D clusters in the scene, especially to address occlusions. 
\revise{By conducting tests with objects at different depths, they showed advantages of event-based sensing over frame-based methods for tracking fast moving objects in 3D.} %
}
\ifarxiv
\begin{figure}[t]
    \centering
    \includegraphics[width=.8\linewidth]{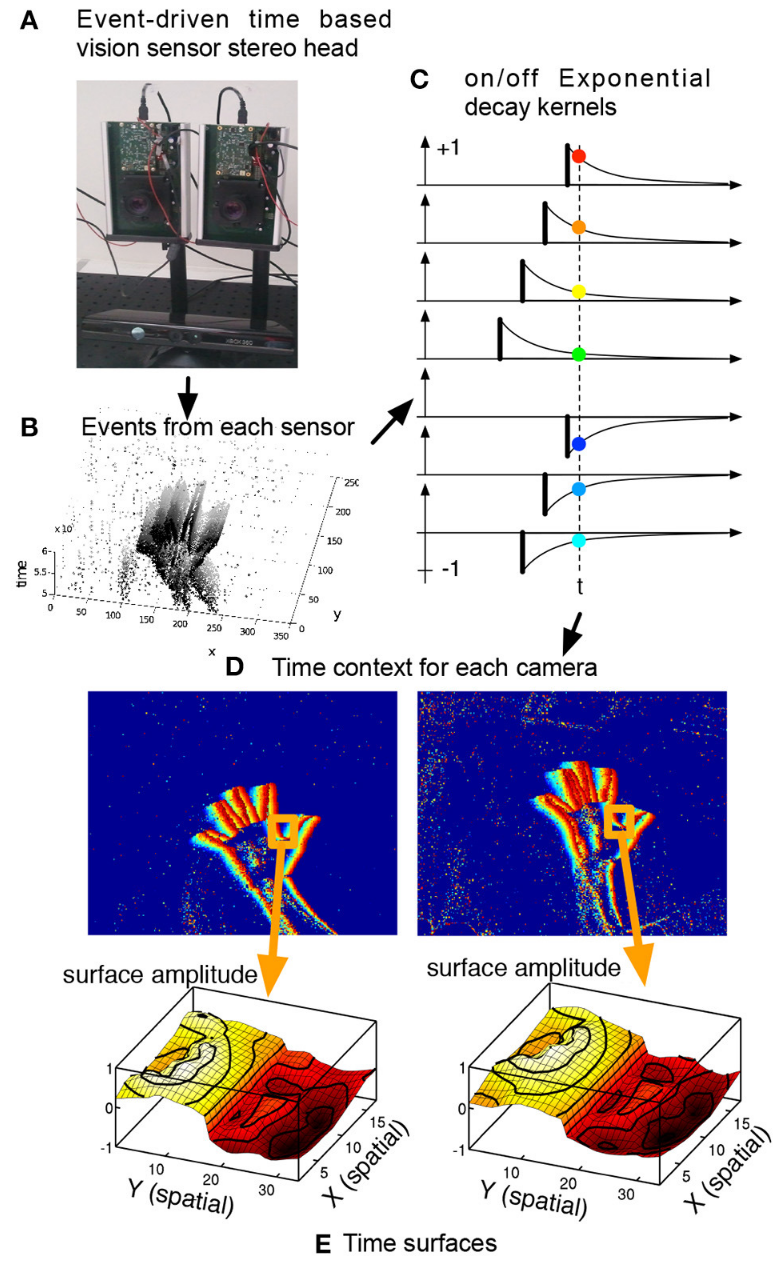}
    \caption{
    \descript{
    Time-surface matching implied by the motion constraint in GTS \cite{Ieng18fnins}.
    Event timestamp information or context (d) visualized as an elevation map leads to the name ``time surface'' (e). 
    Edge trails of time surfaces encode motion information 
    and are used to match patches across event-camera views.
    Image courtesy of \cite{Ieng18fnins}.
    }
    }
    \label{fig:GTS:motionconstratints}
\end{figure}

\fi

\findin{
Most of the instantaneous methods described above were evaluated on scenes with static cameras.}
\challeng{
Scenes with \emph{moving cameras} are more challenging due to the large number of events generated by all (static and dynamic) object contours in the scene.}
\descript{
To tackle this, a line-based feature matching algorithm for depth estimation was proposed in~\firstcite{Everding18thesis} which showed promising results on short real-world scenarios with moving cameras\ifarxiv ~(\cref{fig:stereo:linebased})\fi.
Zou et al.~\firstcite{Zou17bmvc} also demonstrated decent depth estimation results on sequences with camera motion by stereo-matching sharp event images obtained by events over their lifetime~\firstcite{Mueggler15icra}. 
Event lifetime was estimated from time surface gradients by measuring the amount of time it takes for each edge to move by 1 pixel. 
From the sharp stereo event images, classical feature descriptors were extracted and matched, producing a sparse depth map.
They also proposed the first method in the literature to estimate \emph{dense} depth (i.e., for every pixel) from events by inpainting sparse depth maps.}
\ifarxiv
\begin{figure}[t]
    \centering
    \includegraphics[trim={0 21cm 0 0},clip,width=.9\linewidth]{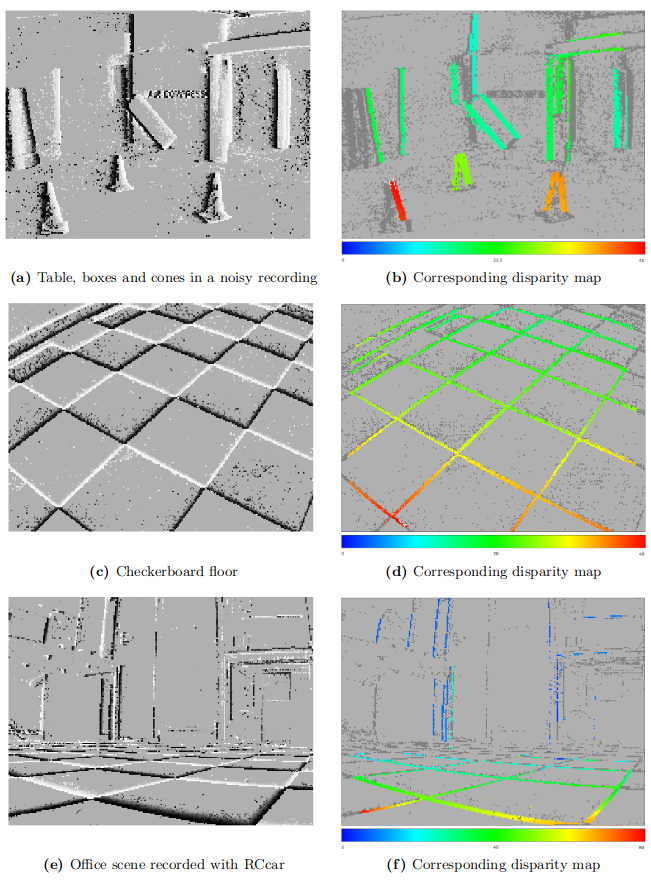}
    \caption{\descript{Results of line-based stereo matching proposed in \cite{Everding18thesis}.} 
    \findin{It produces sparse depth maps in structured environments with small camera motion.}
    Image courtesy of \cite{Everding18thesis}.
    }
    \label{fig:stereo:linebased}
\end{figure}

\fi

\begin{figure}[t]
    \centering
    \includegraphics[width=\linewidth]{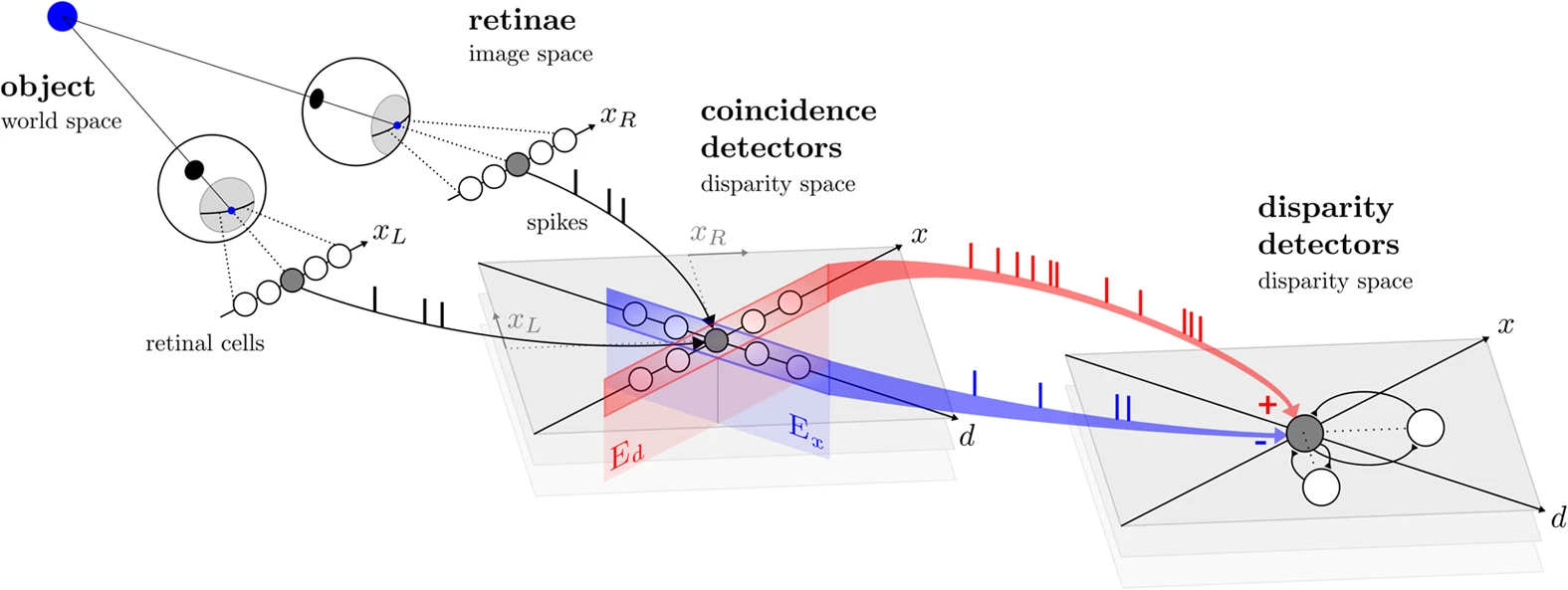}
    \caption{
    \gpass{
    \descript{Asynchronous cooperative stereo method in \cite{Osswald17srep}.} 
    \trend{This architecture was also subsequently adapted and implemented in neuromorphic hardware
platforms \cite{Risi20fnbot,Risi21iscas}, and on FPGA \cite{Kim22access}.}
Image courtesy of \cite{Osswald17srep}.
}
}
    \label{fig:cooperative:spiking}
\end{figure}

\paragraph{\textbf{Cooperative Stereo}}
\descript{
While the methods discussed so far were designed to run on conventional computers (von Neumann machines), an entirely distinct class of stereo algorithms known as ``cooperative stereo'' uses a dynamic network-based computational model of binocular stereopsis for finding correspondences \firstcite{Marr76Science}, 
making it highly efficient for sparse asynchronous event-driven processing on specialized neuromorphic hardware 
(\cref{fig:cooperative:spiking}).
}

\descript{The dynamic network models 3D space: its nodes encode the belief in matching a corresponding pair of pixels from the stereo camera. 
Driven by the input data, local distributed computations in the network interact with each other to provide, upon stabilization, a global solution (akin to a BPN). 
Local excitatory and inhibitory operations encourage disparity smoothness and matching uniqueness, respectively.}

\trend{
Cooperative networks for stereo event data can be traced back to the early days of event-based vision in 1989, when Mahowald and Delbruck~\firstcite{Mahowald89vlsi} showed results on single-line event sensors.}
\descript{Later, Piatkowska et al.~\firstcite{Piatkowska13iccvw} implemented a cooperative technique for depth computation using a winner-take-all mechanism to match temporally-close and spatially-constrained events. 
It was improved via window-based matching~\firstcite{Piatkowska17cvprw}.
Firouzi et al.~\firstcite{Firouzi16npl} also proposed a cooperative algorithm for events, which employed an additional second pattern of inhibitory connections %
to suppress ambiguous matches. %
Besides timestamp and polarity, the orientation of object edges using Gabor filters has also been used as a supplementary signal 
for stereo matching~\firstcite{CamunasMesa14fns}.}
\findin{Overall, cooperative stereo has been a popular technique thanks to its ability to process events with low latency (i.e., without buffering), leading to downstream applications like 3D saliency estimation in humanoid robots~\firstcite{Ghosh22srep}.}

\paragraph{\textbf{Hardware Implementations}}
\label{sec:methods:hardware}

\findin{
Cooperative networks for event-based stereo matching are popular because they can be realized as Spiking Neural Networks (\textbf{SNN}s), which can be implemented on highly efficient neuromorphic hardware} \bg{like SpiNNaker~\firstcite{Furber14ieee}, ROLLS~\firstcite{Qiao15fns}, Loihi~\firstcite{Davies18micro}, DYNAP~\firstcite{Moradi18tbcas} and TrueNorth~\firstcite{Akopyan15tcad}.} 
\trend{Recent works have proven this idea on small-scale scenes with stationary cameras~\firstcite{Osswald17srep,Dikov17cbbs,Risi21iscas,Risi20fnbot}.
}

\descript{
For example, in~\firstcite{Osswald17srep}, the cooperative method is realized with a hierarchical SNN architecture of coincidence and disparity detector layers (\cref{fig:cooperative:spiking}). 
This SNN architecture was also implemented in a prototype mixed signal processor DYNAP and tested on synthetic data~\firstcite{Risi20fnbot}. 
Risi et al~\firstcite{Risi21iscas} further demonstrated the efficiency and performance of the DYNAP implementation of the cooperative network in~\firstcite{Osswald17srep} by evaluating it on complex real-world scenes from the DHP19 3D human pose tracking dataset. 
On the other hand, Dikov et al.~\firstcite{Dikov17cbbs} extended their previous cooperative method~\firstcite{Firouzi16npl} by implementing it on SpiNNaker digital processor boards.
Recently, Kim et al~\firstcite{Kim22access} showed \proscons{improved power and hardware efficiency by trading off latency for accuracy} by implementing a SNN for stereo matching~\firstcite{Osswald17srep} entirely on Field Programmable Gate Arrays (\textbf{FPGA}).
}

\descript{
The methods discussed so far employed some pre-processing steps (like stereo rectification) on standard computing hardware.
In contrast, Andreopoulos et al.~\firstcite{Andreopoulos18cvpr} implemented the first fully event-driven stereo matching pipeline on IBM TrueNorth digital neuromorphic processors, thereby taking advantage of the sparsity and asynchronous nature of events to produce low power 3D reconstructions.
Moreover, they contextualized their work within the event-driven stereo literature by comparing power and latency metrics using data reported in the original articles on different evaluation datasets, as depicted in \cref{tab:tableAndreopoulos18cvpr}. 
}
\begin{table}[t!]
    \centering
    \caption{
    \descript{
    Comparison of instantaneous model-based stereo methods (Sec.~\ref{sec:stereomethods:inst:model}) adapted from \cite{Andreopoulos18cvpr} (2018). 
    `\bigcirc{}' means feature is not present, 
    `-' means unknown, and \bigbullet{} denotes the presence of the respective feature. 
    Performance metrics (bottom half of the table) are only illustrative, as the methods have not been tested on the same data and platform.    
    }
    }
    \label{tab:tableAndreopoulos18cvpr}    
    \rowcolors{2}{salmon}{white}
    \begin{adjustbox}{max width=\linewidth}
    \setlength{\tabcolsep}{2pt}
\begin{tabular}{lcccccccccc}
\toprule
 Approaches & 
 \rotatebox{90}{Andreopoulos \cite{Andreopoulos18cvpr}} & 
 \rotatebox{90}{Osswald \cite{Osswald17srep}} & 
 \rotatebox{90}{Dikov \cite{Dikov17cbbs}} & 
 \rotatebox{90}{Schraml \cite{Schraml07visapp}} & 
 \rotatebox{90}{Schraml \cite{Schraml15cvpr}} & 
 \rotatebox{90}{Piatkowska \cite{Piatkowska17cvprw}} & 
 \rotatebox{90}{Eibensteiner \cite{Eibensteiner14cvprw}} & 
 \rotatebox{90}{Mahowald \cite{Mahowald92thesis}} & 
 \rotatebox{90}{Rogister \cite{Rogister12tnnls}} & 
 \rotatebox{90}{Camunñas \cite{CamunasMesa14fns}} \\
 \multicolumn{11}{l}{\emph{1. Features of disparity algorithm and implementation}} \\
 Fully neuromorphic computation &\bigbullet{} &\bigcirc{}&\bigcirc{}&\bigcirc{}&\bigcirc{}&\bigcirc{}&\bigcirc{}&\bigbullet{} &\bigcirc{}&\bigcirc{}\\
 Neuromorphic rectification of data &\bigbullet{} &\bigcirc{}&\bigcirc{}&\bigcirc{}&\bigcirc{}&\bigcirc{}&\bigcirc{}&\bigcirc{}&\bigcirc{}&\bigcirc{}\\
 Real time depth from live sensor &\bigbullet{} &\bigcirc{}&\bigcirc{}&\bigbullet{} &\bigbullet{} &\bigcirc{}&\bigcirc{}&\bigcirc{}&\bigbullet{} &\bigbullet{} \\
 Multi-resolution computation &\bigbullet{} &\bigcirc{}&\bigcirc{}&\bigcirc{}&\bigcirc{}&\bigcirc{}&\bigcirc{}&\bigbullet{} &\bigcirc{}&\bigcirc{}\\
 Bidirectional consistency check &\bigbullet{} &\bigcirc{}&\bigcirc{}&\bigbullet{} &\bigbullet{} &\bigcirc{}&\bigbullet{} &\bigcirc{}&\bigbullet{} &\bigcirc{}\\
 Scene-agnostic throughput &\bigbullet{} &\bigcirc{}&\bigcirc{}&\bigbullet{} &\bigbullet{} &\bigcirc{}&\bigbullet{} &\bigcirc{}&\bigbullet{} &\bigbullet{} \\
 Uses event polarity compatibility &\bigbullet{} &\bigbullet{} &\bigcirc{}&\bigcirc{}&\bigcirc{}&\bigcirc{}&\bigbullet{} & NA &\bigbullet{} &\bigbullet{} \\
 Tested on dense RDS data &\bigbullet{} &\bigbullet{} &\bigcirc{}&\bigcirc{}&\bigcirc{}&\bigcirc{}&\bigcirc{}&\bigbullet{} &\bigcirc{}&\bigcirc{}\\
 Tested on fast and slow motions &\bigbullet{} &\bigcirc{}&\bigbullet{} &\bigcirc{}&\bigcirc{}&\bigcirc{}&\bigbullet{} &\bigcirc{}&\bigbullet{} &\bigcirc{}\\
 \midrule
 \multicolumn{11}{l}{\emph{2. Implementation metrics}} \\
 Hardware implementation & Neuro & FPGA & CPU & DSP & CPU & CPU & FPGA & ASIC & CPU & FPGA \\
 Energy consumption (mW/px) & 0.058 & - & 16 & 0.30 & - & - & - & - & - & - \\
 Disparity maps per second & 400 & 151 & 500 & 200 & - & - & 1140 & 40 & 3333 & 20 \\
 System latency (ms) & 9 & 6.6 & 2 & 5 & - & - & 0.87 & 25 & 0.3 & 50 \\
 Sensor size, in real-time (px) & 10800 & 32400 & 11236 & 16384 & 1.4M & - & 16384 & 57 & 16384 & 16384 \\
 Disparity levels in real-time & 21 & 30 & 32 & - & - & - & 36 & 9 & 128 & - \\
\bottomrule
\end{tabular}
    \end{adjustbox}
\end{table}

\ifarxiv
\paragraph{\textbf{Stereo with a single event camera}}
\bg{
This is a rare case in the literature, but worth surveying.}
\descript{\challeng{To eliminate the cost of using two separate event cameras for stereo matching},~\firstcite{Ilani22iaoc} mounted a stereo lens on a single event camera (\cref{fig:stereolens}). 
This setup splits the field of view (\textbf{FOV}) horizontally, with each half of the camera sensor array dedicated to one view. 
The stereo disparity between the views was used to reconstruct plasma sparks in 3D. 
Although the methodology is not described in detail, it seems that a standard frame-based stereo matching algorithm on accumulated events was used.}
\proscons{Besides the reduction in cost and effort in building a stereo mount, such a setup also has the benefit that both halves of the sensor have similar bias and noise properties which could make stereo matching easier. 
However, with such a setup, spatial resolution and FOV decreases (by half), the baseline is fixed and the lens still needs to be calibrated (for minor focus adjustment).
}
\begin{figure}[ht]
    \centering
    \includegraphics[trim={2cm 2cm 20cm 4cm},clip,width=.55\linewidth]{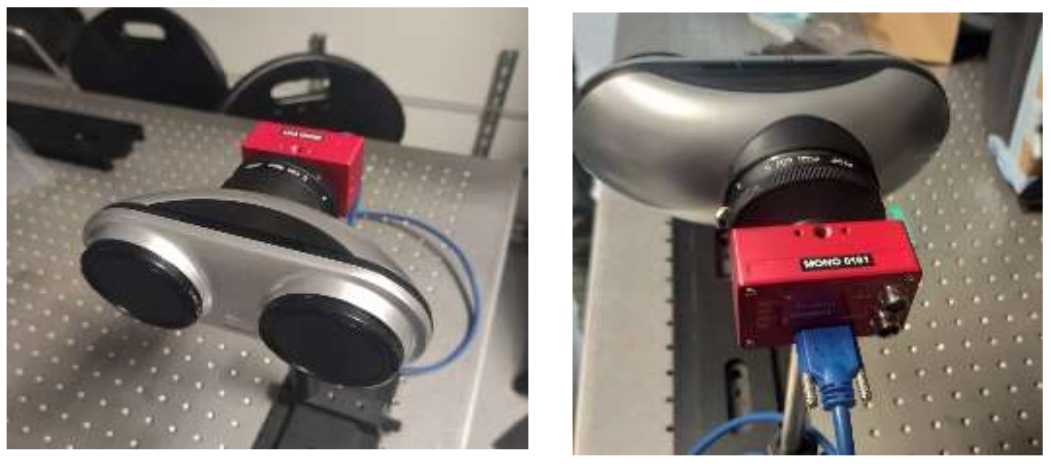}
    \caption{\descript{Setup using a stereo lens and an event camera~\cite{Ilani22iaoc}.}}
    \label{fig:stereolens}
\end{figure}

\fi

\ifarxiv
\def\figWidth{\linewidth}
\begin{figure}[t!]
\centering
\begin{subfigure}{\figWidth}
  \centering
  {\includegraphics[trim={0 0 0 5cm},clip,width=\linewidth]{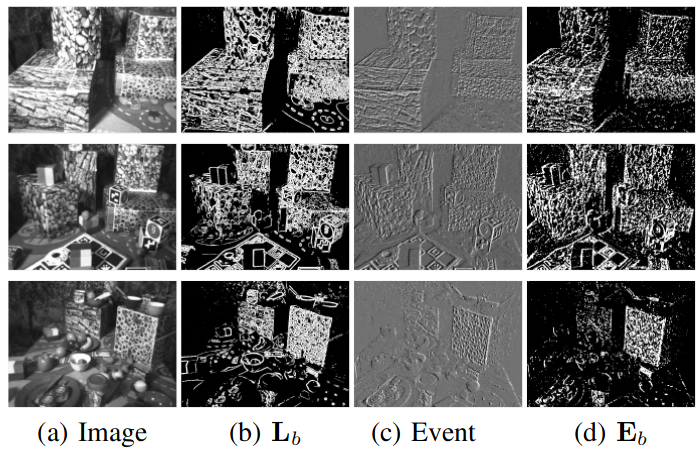}}
  \caption{
  \descript{
  The columns named $\mathbf{L}_b$ and $\mathbf{E}_b$ depict edge maps derived from intensity frames and events, respectively.\label{fig:shef:images}
  }
  }
\end{subfigure}\\[1ex]
\begin{subfigure}{\figWidth}
  \centering
  {\includegraphics[trim={0 0 0 8cm},clip,width=.95\linewidth]{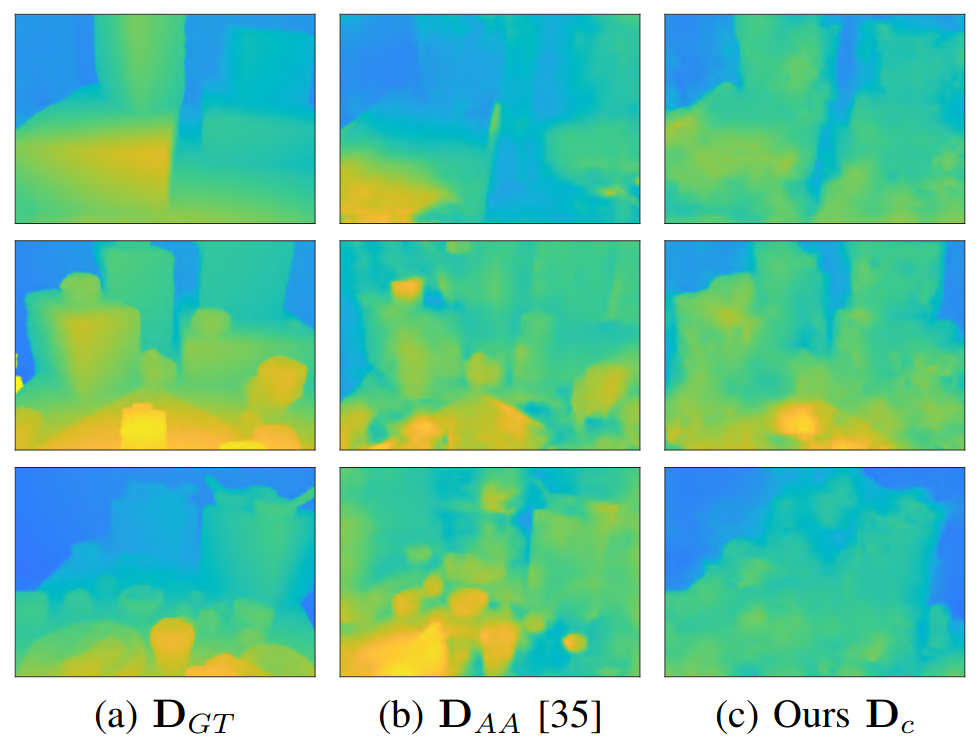}}
    \caption{
    \descript{
    Results of dense depth estimation using SHEF. 
    The left column depicts the sparse depth maps initially estimated by the algorithm. 
    The densification coarsely resembles the GT depth. 
    The column $\mathbf{D}_{AA}$ depicts the result of frame-based depth estimation from reconstructed frames.
  \label{fig:shef:depth}
    }
    }
\end{subfigure}
    \caption{
    \descript{
    Stereo hybrid matching using edge maps extracted from images and events. 
    Image courtesy of SHEF \cite{Wang21iros}.
    }
    }
    \label{fig:shef}
\end{figure}

\fi
\paragraph{\textbf{Stereo from Events and Intensity frames}}
\trend{
An increasing number of publications combine the complementary characteristics of frame-based and event-based cameras for stereo depth estimation.}
\descript{Hadviger et al~\firstcite{Hadviger21advro} calculated disparity using intensity information from standard stereo frames, and used the optical flow computed from a single event camera to interpolate disparity in between frames, leading to a high frame rate depth map. 
}
\descript{In contrast, a Stereo Hybrid Event-Frame (\textbf{SHEF}) matching method between a frame-based and an event camera was proposed in~\firstcite{Wang21iros}, where sparse disparity was computed by cross-correlating \emph{edge maps} from both sensors \ifarxiv ~(\cref{fig:shef})\fi. 
This produced semi-dense disparity maps that were fed into a disparity completion network (DCNet) to densify the final output,
\proscons{but the reported dense depth maps are not as accurate as those from the end-to-end learning methods in \cref{sec:stereomethods:inst:learning}.}
}

\ifclearlook\cleardoublepage\fi \subsubsection{Learning-based Methods}
\label{sec:stereomethods:inst:learning}

\trend{
Since 2019 there has been a steady rise of literature on using deep learning models for stereo depth estimation (\cref{fig:publications:modelvslearning}).}
\findin{
Most of these methods approach the problem by proposing novel representations (``embeddings'') for converting events to fixed-size image-like arrays that are compatible with modern image-based deep neural networks (\textbf{DNNs}). 
These representations tend to show clear edge features, with minimal blur and high dynamic range. 
They are subsequently used for finding correspondences across stereo pairs by forming and processing 3D cost volumes, i.e., 3D arrays that encode the similarity of candidate stereo pixel pairs according to different disparity values.
}

\organiz{
This section discusses deep-learning methods for stereo depth estimation using events, possibly in combination with additional inputs, such as intensity frames.}
\descript{
Most of the methods discussed (except \cite{Cho23cvpr,Uddin22tcsvt,Liu22icra, Gu22jist, Lin25tcsvt}) employ supervised learning for stereo depth estimation, using depth maps acquired by a LiDAR as ground truth (GT) labels. 
They output dense disparity/depth maps.
}

\paragraph{\textbf{Event-only methods (2E)}} [as indicated in the last column of \cref{tab:stereomethods}].
\label{sec:stereomethods:inst:learning:2E}
\descript{
Deep Dense Event Stereo (\textbf{DDES})~\firstcite{Tulyakov19iccv} was the first event-based deep stereo method 
which learned \emph{event sequence embeddings} via continuous fully connected layers.}  
\proscons{However, the resulting dense depth maps presented poor details around object edges and local structures.} 
\trend{Many subsequent methods tried to overcome this by including intensity information explicitly (either by reconstructing images from events or using co-captured frames).
}

\descript{
For instance, the Event-Image-Translation-Network (\textbf{EIT-Net})~\firstcite{Ahmed21aaai} improves upon DDES~\firstcite{Tulyakov19iccv} by reconstructing images from events and using them as a guiding signal for stereo matching.
Instead of fusing events and (reconstructed) frames directly, a Convolution Neural Network (CNN)-based module, called stacked dilated SPatially-Adaptive DEnormalization (\textbf{SPADE}), modulates the event features using the reconstructed image features. 
}

\findin{
A fast and lightweight deep stereo network is proposed in the Discrete Time Convolution (\textbf{DTC}) SPADE (\textbf{DTC-SPADE}) framework~\firstcite{Zhang22cvpr} that includes temporal dynamics during event embedding step, by processing spatio-temporal event voxel grids with discrete-time CNNs \ifarxiv ~(\cref{fig:dtcspade})\fi.
} 
The output from the DTC embedding layers is fed to a spatial embedding module, followed by the matching and regularization modules similar to DDES~\firstcite{Tulyakov19iccv}. 
As in EIT-Net~\firstcite{Ahmed21aaai}, SPADE is also used before stereo matching to inject semantic information.
\ifarxiv
\begin{figure}[t]
    \centering
    \includegraphics[width=\linewidth]{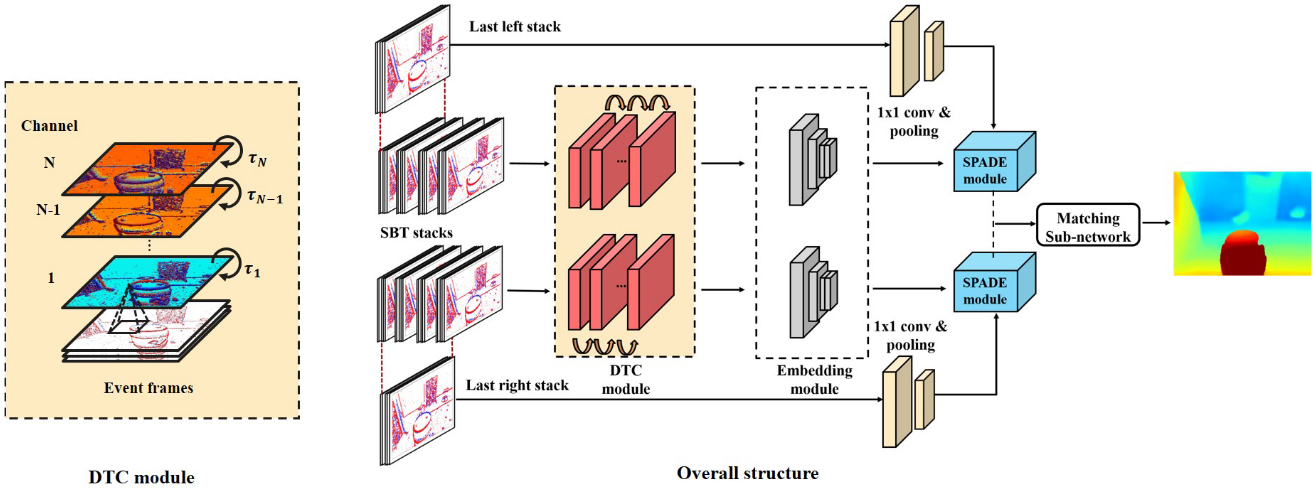}
    \caption{DTC-SPADE architecture. 
    Image courtesy of \cite{Zhang22cvpr}.}
    \label{fig:dtcspade}
\end{figure}

\fi
Ghosh et al.~\firstcite{Ghosh24jeswa} further improved upon the SPADE framework by using a novel two-stage coarse-to-fine strategy for stereo matching.

\descript{While the previous works focused on learning embeddings from spatio-temporal event volumes, the Adaptive Stacks Depth Estimation Network (\textbf{ASNet}) \firstcite{Jianguo23iecon} proposed the use of \emph{event-histogram stacks} as input to an image-based stereo matching network that follows the architecture of MobileStereoNet~\cite{Shamsafar22wacv}.} 
\findin{Their key contribution was to adaptively modify event-histogram stack lengths using the event rate in order to produce sharp images for improved matching.} 
\descript{
Another approach of directly extracting features from stereo events was proposed in the ConvLSTM Event Stereo Network (\textbf{CES-Net})~\firstcite{Kweon21cvprw}, which achieved the best results in mean average disparity error (MAE) in the DSEC disparity benchmark among the event-only methods at the CVPR Event Vision Workshop 2023. 
}

\descript{
Recently, Chen et al.~\firstcite{Chen24spl} used \emph{cross-attention} mechanisms to provide additional temporal and spatial context to voxelized event inputs, leading to highly improved performance.
Attention was also used in the Motion-Guided Event-Based Stereo Disparity Estimation Network (\textbf{EV-MGDispNet})~\firstcite{Jiang24tim}, where stacked event volumes (called MES) 
are fused with time surfaces (which the authors term motion confidence maps) to generate ``edge-aware'' event frames, from which multi-scale features are extracted using deformable attention layers. 
The features are then used for stereo matching and disparity regression.
}

Additional \emph{temporal context} is also utilized by \descript{\textbf{StereoFlow-Net}~\firstcite{Cho24eccv}, which achieves high accuracy} \findin{by aggregating 
event features and cost volumes from the past into the current time.}
\descript{
On top of a stereo matching network, it uses an optical flow prediction network that learns by warping GT disparity maps from the past to the current time. 
\findin{This is a way of learning 3D scene flow using supervision from disparity maps without the need for GT optical flow} \challeng{(which is hard to obtain)}. 
}
\proscons{
This type of temporal aggregation, which explicitly models motion, preserves sharp edges better (even during small relative motion) and is more lightweight than the recurrent structures used in frameworks like DTC-SPADE~\firstcite{Zhang22cvpr} discussed above. 
}

\descript{
The methods discussed so far relied on GT depth from LiDAR or depth sensors for supervision, which limits applicability to out-of-distribution scenarios without re-training.
An \textbf{unsupervised} method for estimating sparse depth maps was proposed in~\firstcite{Liu22icra}. 
It is a hybrid approach that combines machine learning with model-based optimization.
Separate Siamese DNNs were trained to find efficient and accurate event representations from stereo cameras. 
Feature descriptors from these representations were then used to find the optimal disparity which minimizes matching cost.
}
\descript{
Steffen et al.~\firstcite{Steffen19icar} presented another unsupervised approach for finding stereo correspondences between two event cameras using \emph{Self-Organizing Maps} \cite{Kohonen01book}, where
2D event data from both views was mapped to a latent 3D representation similar to a disparity space image grid, without requiring knowledge of the camera parameters.}
\proscons{The qualitative results of reconstructing a moving cube observed by stationary cameras in simulation showed limited success.}

\begin{figure}[t]
    \centering
    \includegraphics[width=\linewidth]{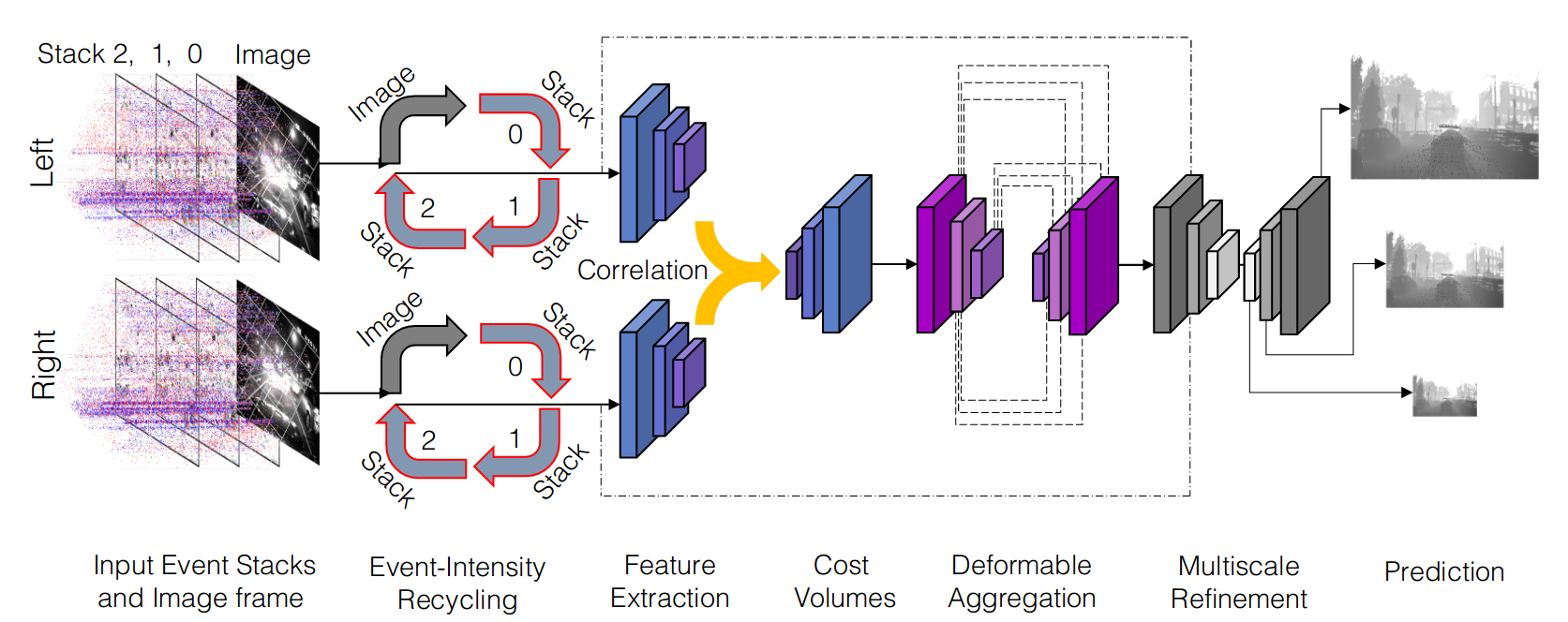}
    \caption{
    \descript{Network structure of EI-Stereo \cite{Mostafavi21iccv}.    
    Frames and events are stacked and combined using a recycling network to form image-like representations. 
    These are subsequently used for stereo matching using feature extraction, cost volume computation, deformable aggregation and multiscale refinement}, \findin{which are mainstays in frame-based stereo deep-learning architectures.}
    Image courtesy of \cite{Mostafavi21iccv}.
    }
    \label{fig:eistereo}
\end{figure}

\paragraph{\textbf{Events-and-Intensity methods (2E+2F)}}
\label{sec:stereomethods:inst:learning:2E+2F}
\descript{
The first learning-based stereo depth estimation using both frames and events was proposed in 
Event-Intensity Stereo (\textbf{EIS}) \firstcite{Mostafavi21iccv}.   
It combines intensity frames and event voxel grids into a blur-free HDR image-like representation using a recurrent recycling network \firstcite{Mostafavi21pami}.
\descript{From the HDR representations, their stereo matching network extracts depth via: 
(1) the feature extraction module, 
(2) the cost volume module, 
(3) the deformable aggregation module, and 
(4) the multiscale refinement module.}
\findin{This architecture (\cref{fig:eistereo}) is standard among stereo matching frameworks and is also used in other works, with the main difference being at the feature extraction stage.}
}
\proscons{
It aims to learn from multi-modal data (events and frames), and can substitute one modality by the other based on availability. 
For instance, even when frames are missing, it can still output disparity from events.
}

\descript{
Stereo Cross-Modality network (\textbf{SCM-Net})~\firstcite{Cho22aaai} is a similar work that uses an event-intensity fusion network to combine intensity features and event embeddings learned from the DDES method~\firstcite{Tulyakov19iccv}. 
}
\proscons{Similar to EIS, it uses an embedding that prioritizes most recent events, as is often done for video frame interpolation and super-resolution.} 

\descript{
By contrast, the Selection and Cross Similarity network (\textbf{SCS-Net})~\firstcite{Cho22eccv} proposes an alternative approach that selects "relevant" event stacks for feature extraction
by using a novel event-image embedding for stereo matching. 
The relevance of events depends on the camera motion -- 
it discards time slices containing insufficient motion or when the viewing perspectives have little overlap. 
This relevance-based embedding, along with a cross-similarity feature loss that matches local features between events and frames, produces disparity maps with sharp edges.}

\def\figWidth{\linewidth}
\begin{figure}[t!]
\centering
\begin{subfigure}{\figWidth}
  \centering
  {\includegraphics[width=\linewidth]{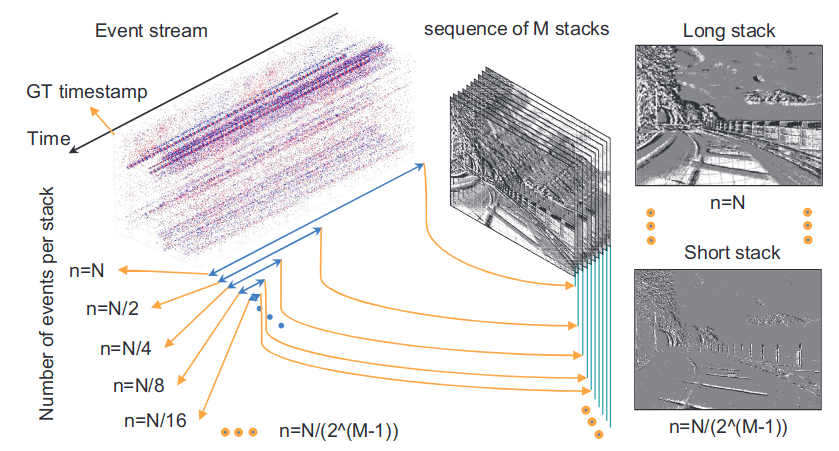}}
  \caption{\label{fig:concentnet:diagram}How image-like stacks are built from event data.}
\end{subfigure}\\[1ex]
\begin{subfigure}{\figWidth}
  \centering
  {\includegraphics[trim={0 16.3cm 0 0},clip,width=\linewidth]{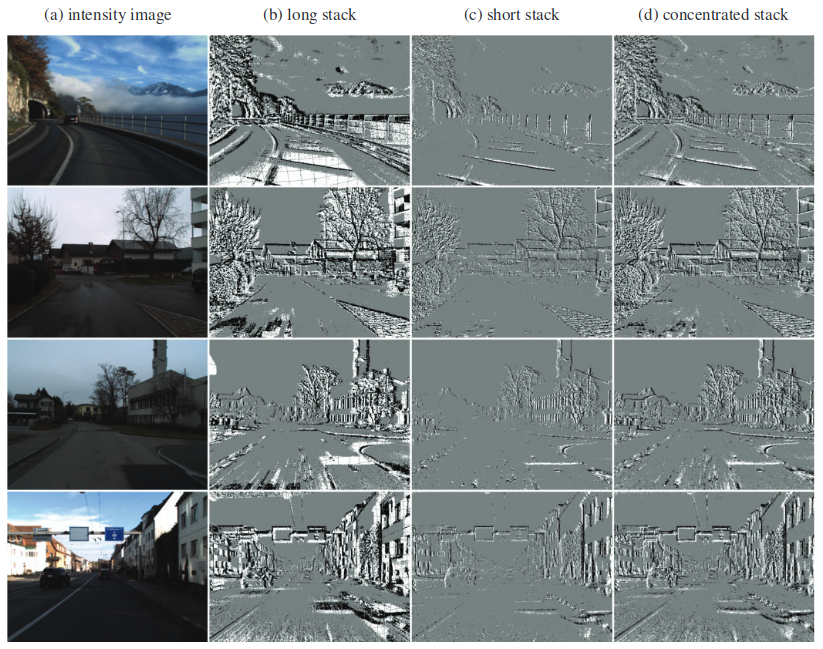}}
  \caption{\label{fig:concentnet:results}Comparison of various image-like event stacks.}
\end{subfigure}
    \caption{
    \descript{
    Conc-Net in \cite{Nam22cvpr} determines the optimal event stack size for each pixel in order to form sharp HDR image-like representations of the scene, called ``concentrated stacks''. 
    Image courtesy of \cite{Nam22cvpr}.
    }
    }
    \label{fig:concentnet}
\end{figure}

\trend{
Similar to the event-selection network’s role of selecting relevant events in SCS-Net~\firstcite{Cho22eccv}, the Concentration Network (\textbf{Conc-Net})~\firstcite{Nam22cvpr} aims to select optimal event window sizes to generate a sharp image-like array from events.} 
\proscons{As shown in \cref{fig:concentnet}, Conc-Net attempts to form an event histogram in which the contribution of events is weighted per pixel, rather than having a fixed number of past events as in ASNet~\firstcite{Jianguo23iecon}, which makes the former more flexible.}
\descript{It also has a variant (called Conc-Net + I) that combines intensity frames with the concentrated event stacks before the feature extraction stage.}
Recently, Zhao et al.~\firstcite{Zhao24eccv} explicitly used edges extracted from intensity frames to sharpen edges in event-intensity stereo depth estimation, producing state-of-the-art (\textbf{SOTA}) performance.

\descript{
While the methods discussed above rely on supervision from GT depth,
a \textbf{self-supervised} approach for learning stereo depth was proposed in~\firstcite{Uddin22tcsvt}. 
It is a follow-up work from the event-only supervised method EIT-Net~\firstcite{Ahmed21aaai}. %
A stereo matching network processes stereo events to predict dense disparity maps,
which are used to project the right intensity image onto the left camera. 
The similarity between the warped and original left intensity frames is used as supervisory signal.} 
\proscons{However, since it relies on intensity frames, %
its performance is limited in challenging situations where event cameras excel, like HDR and fast motion.}

\challeng{To circumvent the need for large amounts of good quality GT depth data for training a supervised network that generalizes well}, \descript{an \textbf{unsupervised domain adaptation} approach called Adaptive Dense Event Stereo (\textbf{ADES}) is adopted in~\firstcite{Cho23cvpr}. 
By training on the source domain of learning stereo depth from intensity images from abundant frame-based datasets with GT depth, and using self-supervision from video-to-event and event-to-video reconstructions, the network is able to adapt to the target task of estimating dense depth from just stereo events.
} 

\paragraph{\textbf{Hybrid stereo (1E+1F)}}
\label{sec:stereomethods:inst:learning:1E+1F}
\descript{
This section discusses methods that estimate depth from a single frame-based camera and a single event camera in stereo configuration.
Hybrid Disparity EStimation network (\textbf{HDES})~\firstcite{Zuo21iros} learns stereo depth via spatio-temporal representations similar to DDES~\firstcite{Tulyakov19iccv}, along with a novel hybrid pyramid attention module.}
\proscons{While the model-based model SHEF~\cite{Wang21iros} uses a separate neural network to densify its sparse depth output, HDES directly outputs dense depth maps.} 

\descript{Deviating from recurrent networks, a \emph{Transformer}-based framework for pixel-level feature matching between intensity frames and event batches was proposed in~\firstcite{Zhang22eccv2}. It showed good performance in finding correspondences between overlapping views in small stereo baseline setups.
}
To tackle overfitting due to limited availability of ground truth for training with event data,
\descript{Self-supervised Intensity-Event Stereo Matching (\textbf{SIES}) uses an approach
where events are first converted to a reconstructed image that is stereo-matched with the image from the traditional camera \firstcite{Gu22jist}.
\proscons{The computed disparity can facilitate downstream tasks like video frame interpolation.}
Its self-supervised loss is based on similarity of the two images and their derivatives (edge maps), along with a smoothness term for regularization.}

On the other hand, Zero-shot Event-intensity asymmetric STereo (\textbf{ZEST})~\firstcite{Lou24neurips} proposed a completely \textbf{unsupervised} visual prompting scheme that re-utilizes monocular and stereo depth estimation models pre-trained purely on intensity frames. It reconciles the two modalities by converting events and intensity frames to brightness change images following the event generation model in ~\cref{eq:EGM}. 

\descript{
Recent works like ``Event-based motion Deblurring with STereo event and intensity camera'' (\textbf{St-EDNet}) \firstcite{Lin25tcsvt} and ``Stereo Event-based Video Frame Interpolation network'' (\textbf{SEVFI-Net})~\firstcite{Ding24tmm} approach the problem of hybrid stereo depth estimation for the specialized downstream tasks of frame deblurring and video frame interpolation, respectively. 
In St-EDNet, blurred images are simulated from sharp images, which are then used as GT for supervision. 
SEVFI-Net outputs high frame rate video and interpolated disparity frames, thus overcoming the \challeng{frame rate bottleneck of intensity frames in hybrid intensity-event setups}.
}

\begin{figure}[t]
    \centering
    \begin{subfigure}{0.25\linewidth}
    \centering
    \includegraphics[trim={2.1cm 0 1.1cm 0cm},clip,width=\linewidth]{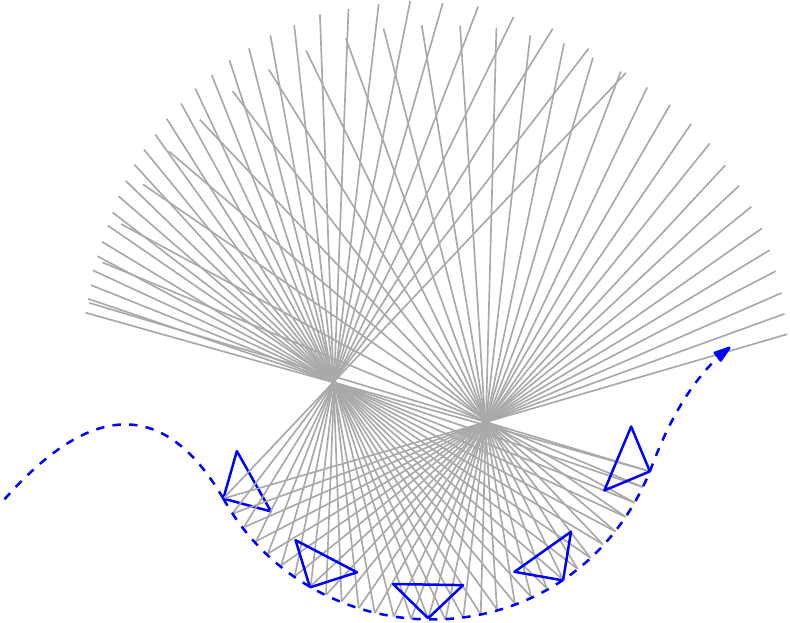}
    \caption{Event rays\label{fig:mcemvs:rays}}
    \end{subfigure}
    \begin{subfigure}{0.74\linewidth}
    \centering
    \includegraphics[trim={6cm 11.2cm 0 0.7cm},clip,width=\linewidth]{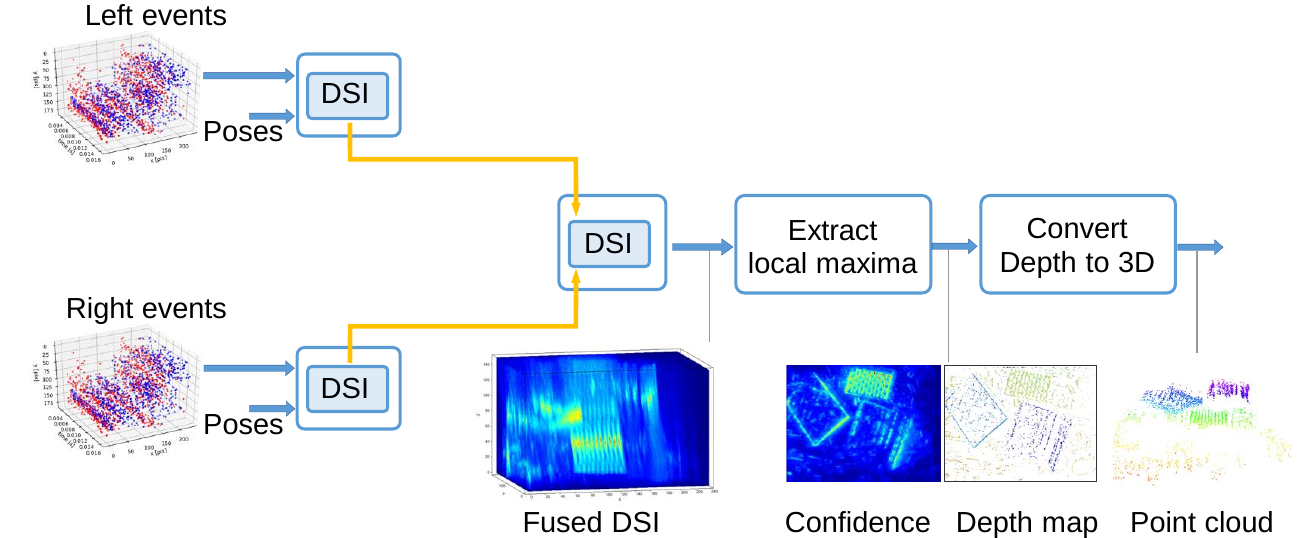}
    \caption{Pipeline\label{fig:mcemvs:pipeline}}
    \end{subfigure}
    \caption{
    \descript{
    Stereo fusion method proposed in MC-EMVS (b). %
    Camera poses are needed as input, to back-project events through space (event rays (a)).
    Images courtesy of \cite{Ghosh22aisy,Rebecq18ijcv}.}
    \label{fig:mcemvs}
    }
\end{figure}

\ifarxiv
\begin{figure}[t]
    \centering
    \includegraphics[width=\linewidth]{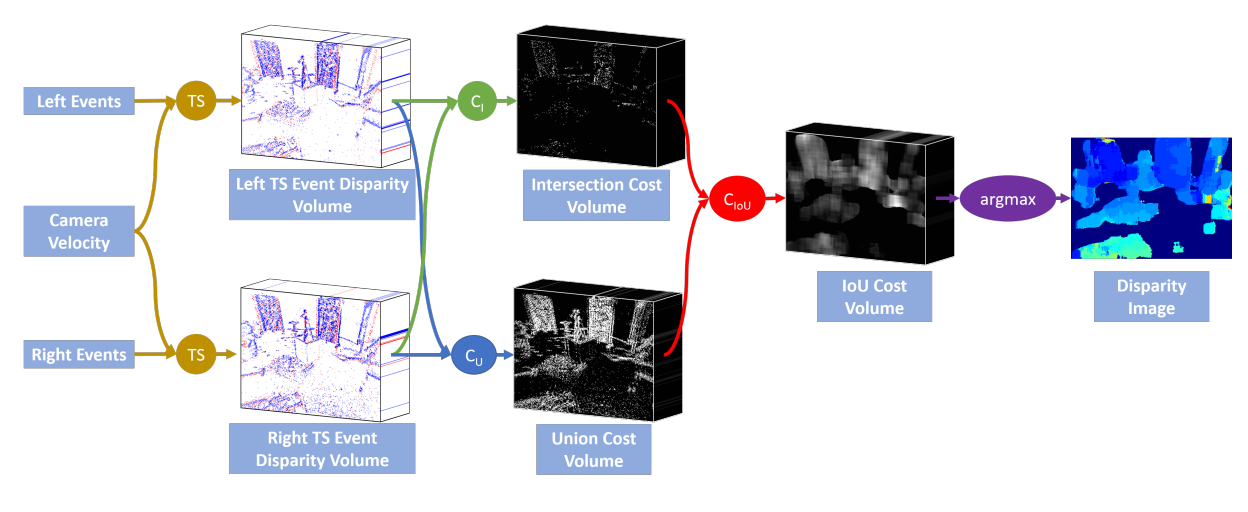}
    \caption{
    \descript{
    Stereo fusion architecture employed in TSES \cite{Zhu18eccv}. 
    Camera velocity is needed as input.
    Image courtesy of \cite{Zhu18eccv}.
    }
    }
    \label{fig:Zhu18eccv}
\end{figure}

\fi
\paragraph{\textbf{Stereo Events-LiDAR Fusion (2E + LiDAR)}}
The first work combining stereo events with LiDARs for dense disparity estimation was proposed in~\firstcite{Bartolomei24eccv}. 
It used sparse depth measurements from the LiDAR to inject ``hallucinated'' stereo pairs of events either in the raw stream or in stacked tensor representations. 
These hallucinated events are not related to real-world edges or brightness changes; they are random patterns used for injecting features for stereo matching in event-less regions where LiDAR data is available. 
Since LiDAR operates at a much slower rate (e.g., 10Hz) than events, temporal misalignment can
cause hallucinated stereo events to be injected at inaccurate coordinates.
Although the depth estimation error grows with increased misalignment, this LiDAR augmentation strategy exhibits robustness to small temporal shifts (up to 100 ms).

\paragraph{\textbf{Spiking Neural Network (SNN)}}
\bg{
The sparsity of event data is ideal for efficient processing using SNNs, yet most stereo matching SNNs are not data-driven since \challeng{they are notoriously hard to train}.}
\revise{While Sec.~\ref{sec:methods:hardware} discusses SNN methods where the weights are pre-defined (hand-tuned) by the user, let us present data-driven approaches, whose SNN weights are automatically learned.
\textbf{StereoSpike}~\firstcite{rancon22access} is currently the only SNN that effectively uses deep learning for stereo depth estimation; it leverages surrogate gradients for backpropagation.} 
Instead of explicit stereo matching, data from stereo cameras are combined into a single array from which depth is learned in an end-to-end manner using GT within a U-Net like architecture. 
The same architecture is used for both monocular and stereo depth estimation, producing comparable performance to non-spiking event stereo methods. 
\proscons{StereoSpike's biggest advantage is a fully spiking architecture that benefits from sparsity (which can reduce energy costs) and reduces the number of learned parameters. 
However, it should be noted that DTC-SPADE~\firstcite{Zhang22cvpr} has a comparable parameter size with a better stereo depth estimation accuracy.}

\ifclearlook\cleardoublepage\fi \subsection{Longer temporal baseline Methods}
\label{sec:stereomethods:long}

\organiz{
The previous section discussed stereo methods that produce an instantaneous depth ``snapshot'' of a possibly dynamic scene.}
\findin{By contrast, event-based stereo 3D reconstruction methods for VO/SLAM assume a static world and known camera motion (e.g., from a tracking method) to assimilate events over longer time intervals in order to produce more accurate depth maps (i.e., by increasing parallax). 
Most methods in this category are still model-based.
}

\subsubsection{Model-based Methods}
\label{sec:stereomethods:long:model}
\paragraph{\textbf{Event-only Methods}}
\descript{
Assuming known camera poses, the monocular method Event-based Multi-View Stereo (\textbf{EMVS}) \firstcite{Rebecq18ijcv} casts rays through event pixels in a 3D space (\cref{fig:mcemvs:rays}). 
This collection of rays, computed using a space-sweep approach \firstcite{Collins96cvpr}, is referred to as a Disparity Space Image (\textbf{DSI}), which is represented by a discrete voxel grid. 
The 3D structure of the scene can be recovered from the points of highest ray density, i.e., largest number of intersections of back-projected rays. 
This has connections with the Contrast Maximization (\textbf{CMax}) framework \firstcite{Gallego18cvpr,Gallego19cvpr}: 
the depth slices of the DSI can be interpreted as Images of Warped Events (\textbf{IWEs}) 
and depth is estimated by finding the slice whose IWE has maximum contrast.
EMVS has been extended to the stereo setup in Multi-Camera (\textbf{MC})-\textbf{EMVS} \firstcite{Ghosh22aisy}, where DSIs computed from individual cameras are fused using element-wise operations like the harmonic mean, and the local maxima of the fused DSI yields the depth map (\cref{fig:mcemvs:pipeline}).} 
\proscons{MC-EMVS produced state-of-the-art results in sparse stereo depth estimation,
and it has been employed in \firstcite{Elmoudni23itsc} and further improved in Event Stereo Parallel Tracking and Mapping
(\textbf{ES-PTAM)} \firstcite{Ghosh24eccvw} as the mapping module for SLAM systems.}

\descript{
Instead of camera poses, Time Synchronized Event-based Stereo (\textbf{TSES}) \firstcite{Zhu18eccv} uses known and constant camera velocity to warp events during a short time interval. 
Events are warped using a range of candidate disparities, producing binary IWEs that are stacked into a 3D grid. 
The warping and 3D grid structure are analogous to the space-sweep and DSI in EMVS, respectively.
The volumes of warped events from each camera are then fused using Intersection over Union (IoU) over a spatial neighborhood. 
The final depth map is obtained by maximizing the IoU \ifarxiv ~(\cref{fig:Zhu18eccv})\fi.
}

\begin{figure}[t]
    \centering
    \includegraphics[width=\linewidth]{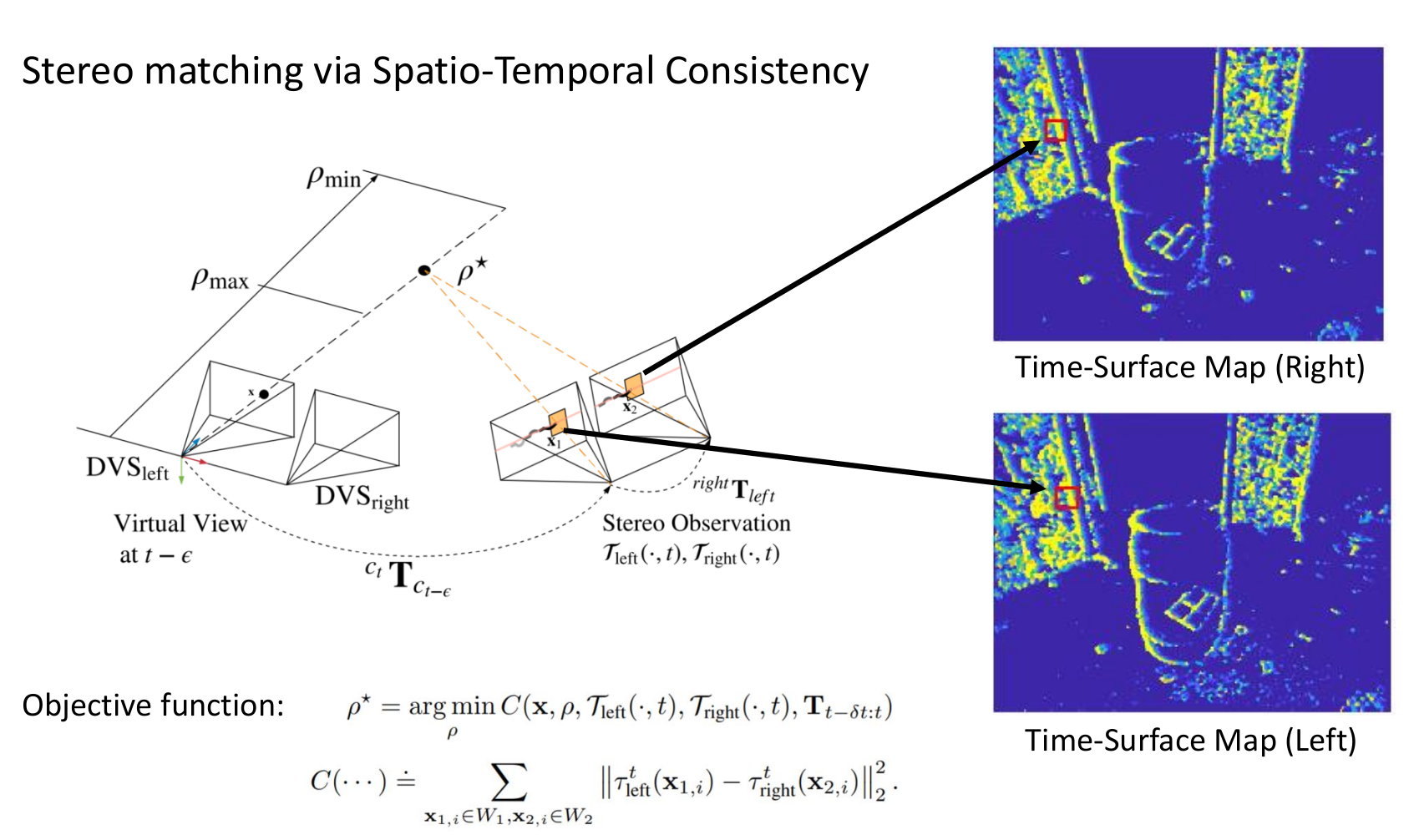}
    \caption{
    \descript{Stereo correspondences in ESVO \cite{Zhou20tro} are obtained by matching time-surface patches across cameras.
    Image courtesy of \cite{Zhou20tro}.}
    }
    \label{fig:esvo}
\end{figure}

\ifarxiv
\begin{figure}[t]
    \centering
    \includegraphics[width=\linewidth]{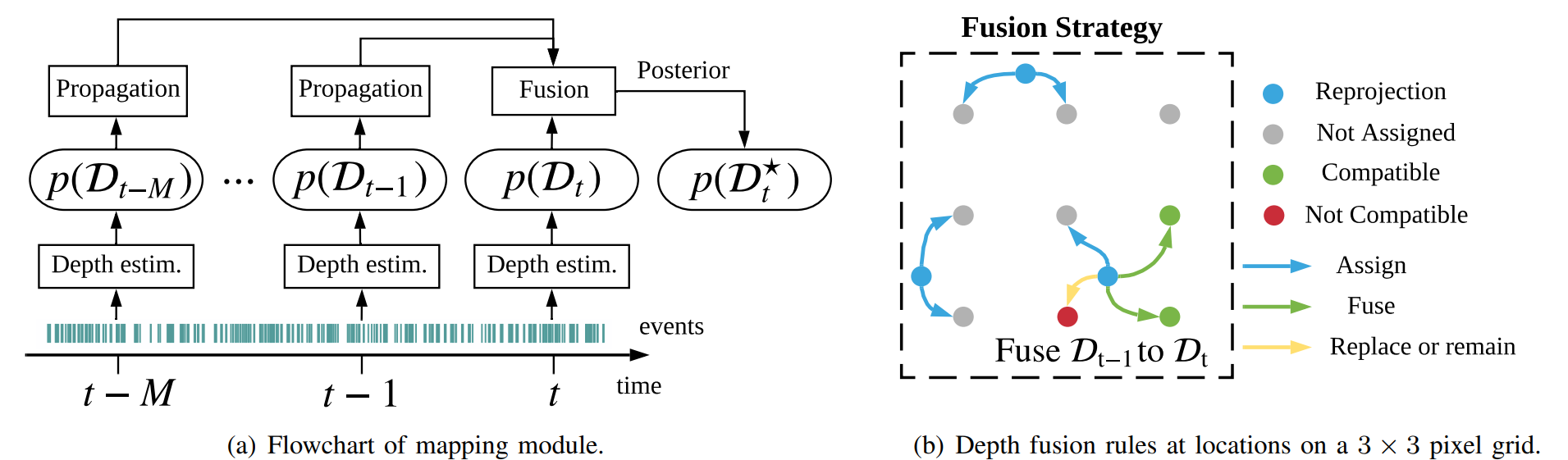}
    \caption{
    \descript{
    Probabilistic fusion of depth predictions through time in ESVO.} 
    \findin{For persistent scenes, fusion can help recover cleaner and denser depth estimates.}
    Image courtesy of \cite{Zhou20tro}.
    }
    \label{fig:esvo:fusion}
\end{figure}

\fi

\trend{
For long-term stereo, Event-based Stereo Visual Odometry (\textbf{ESVO}) \firstcite{Zhou20tro} is an established framework, which has inspired many other works~\cite{Liu23aisy,Wang23sensors,Liu23sensors,Niu24esvo2}.} 
\descript{Its mapping module is an improved version of the method proposed in \firstcite{Zhou18eccv}, 
where instantaneous depth is estimated by matching time surfaces across stereo cameras (\cref{fig:esvo}), which are then fused using Student-t filters and known camera poses over a longer duration \ifarxiv ~(\cref{fig:esvo:fusion})\fi.
Depth fusion is a critical component for ESVO's success as it confers consistency to the 3D map.
\challeng{To robustify against varying camera speeds}, \mbox{T-ESVO} \firstcite{Liu23aisy} proposed using adaptive time surfaces whose time constant (decay rate) is determined by the input event rate.
}

\descript{
Recently, Shiba et al. \firstcite{Shiba24pami} extended the CMax framework \challeng{to simultaneously estimate depth and ego-motion using an optical flow warp, which is a challenging task.}
Their qualitative results demonstrate \proscons{good overall stereo depth estimation, 
albeit with artefacts due to the short time window 
where the optical flow warp model is a valid approximation to the non-linear motion in the scene (as in TSES).}
}

\paragraph{\textbf{Hybrid Stereo Methods (1E + 1F)}}
\descript{
\firstcite{Kim22ral} proposed a long-term stereo matching algorithm using a frame-based and an event camera. 
It first estimates an initial depth by matching edge maps from both cameras using normalized cross-correlation. 
Using this initial depth, camera poses are estimated over a short time, which are used to fine-align edge maps from both cameras. 
Then, motion-compensated edge maps are used to estimate more accurate depth. 
}

\ifarxiv
\paragraph{\textbf{Events + IMU Methods (``ESVIO'')}}
\trend{
Recent works have also fused events with inertial measurement unit (\textbf{IMU}) data} \challeng{to build more robust stereo VO systems.}
\descript{
For example, Wang et al. \firstcite{Wang23sensors} improved upon ESVO \firstcite{Zhou20tro} by fusing IMU data with estimated camera poses using an Extended Kalman Filter, yielding better depth maps over long periods.
Feature-based event-based stereo visual-inertial odometry (\textbf{ESVIO}) \firstcite{Chen23ral} combines corner feature tracking and IMU readings to estimate odometry. 
Stereo event time surfaces are first motion-compensated using angular velocity from the IMU (while ignoring the translation component). 
Then, corners are tracked on the motion-compensated time surfaces using Arc$^\ast$ \firstcite{Alzugaray18ral}, and used to triangulate for instantaneous stereo matching, as well as for temporal camera tracking. 
The system is augmented with standard frame-based stereo matching \challeng{for robustness}, and is tested on several challenging and} \challeng{low-light sequences}. %
\proscons{Since this method works using feature tracking, focus is on ego-motion recovery and no explicit depth maps are output.}

\descript{
\bg{Event-only feature extraction and tracking is still not accurate and stable.}
Compared with traditional cameras, frames constructed from raw events have \challeng{low resolution, high noise, and fewer texture details due to the working principle and current hardware limitations of event cameras}. 
Hence, Liu et al \firstcite{Liu23sensors} have proposed a direct method of state estimation using stereo events and IMU without explicit feature tracking, referred to as \textbf{direct-ESVIO}. 
It is highly based on ESVO \firstcite{Zhou20tro}. 
They improve upon the time surface matching of ESVO by using an adaptive decay rate which depends on the number of events in the considered time window. 
This formulation helps in recovering features better under \challeng{low-speed camera conditions. 
To reduce computational load}, the depth map is further filtered by downsampling from connected contours.
}

\descript{
The \gred{latest work} in this category, \textbf{ESVO2} \firstcite{Niu24esvo2}, is a follow-up from the authors of the ESVO paper.}
\proscons{They have improved ESVO by adding IMU integration, reducing latency and remodeling depth estimation of horizontal edges.}
\fi

\begin{figure}[t]
    \centering
    \includegraphics[width=\linewidth]{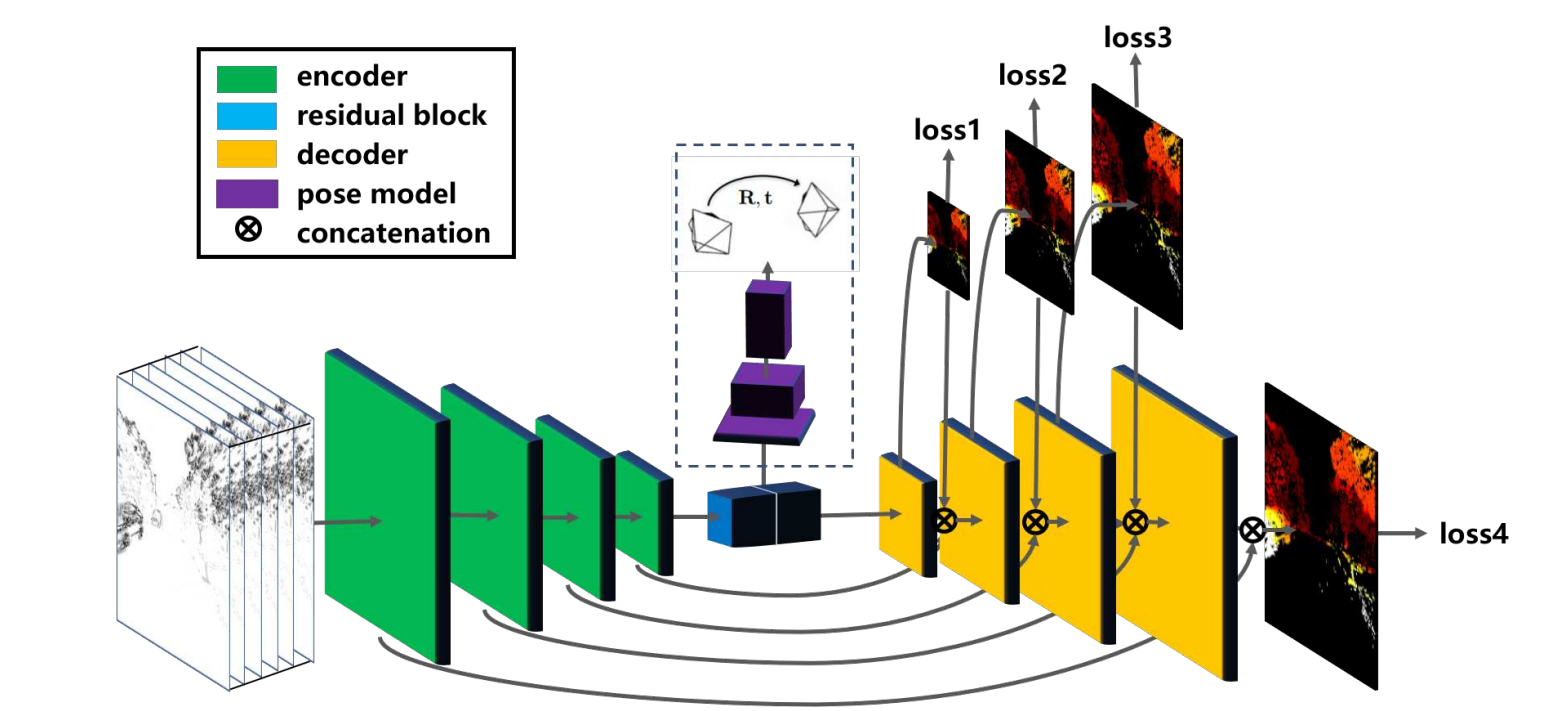}
    \caption{
    \descript{
    Unsupervised depth and ego-motion estimation deep-learning model proposed in \cite{Zhu19cvpr}. 
    The depth-pose network (in purple) jointly estimates the depth and camera motion whose motion field maximizes sharpness of the IWE of a static scene.
    Image courtesy of \cite{Zhu19cvpr}.
    }
    }
    \label{fig:unsuperviseddepthandegomotion}
\end{figure}

\subsubsection{Learning-based Methods}
\label{sec:stereomethods:long:learning}
\findin{
We have not found extensive literature on learning-based methods for stereo depth estimation over long time windows.} 
\descript{One relevant work in this category is by Zhu et al. \firstcite{Zhu19cvpr} which proposes an unsupervised, deep-learning solution for estimating optical flow, and depth with ego-motion, using two separate DNNs based on CMax. 
The depth-pose network (\cref{fig:unsuperviseddepthandegomotion}) learns to predict the correct camera motion and scene depth that minimizes motion blur (maximizes sharpness) of the IWE. 
A supervised stereo disparity loss is incorporated within a composite training loss function that also includes temporal aggregation.
}

\descript{
Recently, a Stereo Asymmetric Frame-Event (\textbf{SAFE}) system \firstcite{Chen24wacv} was proposed as a divide-and-conquer approach for estimating depth in hybrid event-intensity setups. 
Edges containing both event and intensity information are matched using an ``instantaneous'' stereo matching network. 
To estimate dense depth in other image regions, separate structure-from-motion (SfM) DNNs aggregate temporal information from multiple views of both event and frame-based cameras, estimating relative camera pose and a depth cost volume. 
Finally, the cost volumes from three networks are fused in a fourth network to predict dense depth. 
\findin{Thus, it combines instantaneous and long-term multi-view stereo.} 
It also uses temporal fusion \challeng{to handle occlusions},
and produces SOTA performance (\cref{tab:quant:deepmethods} last row).
}

\descript{
Deep Hybrid Parallel Tracking and Mapping (\textbf{DH-PTAM}) \firstcite{Soliman24tiv} uses a hybrid approach, combining model-based estimation with deep learning, for solving SLAM with stereo pairs of event and frame-based cameras. 
It is a full SLAM system that includes bundle adjustment and loop closure. 
Events and frames are converted into a single image-like representation called E3CT, from which features like Superpoints and R2D2 are extracted using deep learning, followed by model-based stereo matching. 
\proscons{Despite using a sophisticated architecture, the camera tracking accuracy obtained by combining frames and events is worse or comparable than with just stereo frames.
Hence, the results do not highlight the benefits of event data for SLAM.
The estimated depth maps reported look worse qualitatively than monocular methods like EDS~\firstcite{Hidalgo22cvpr}.}
}

Recently, \textbf{DERD-Net} \firstcite{deOliveira25derdnet} demonstrated SOTA performance by learning semi-dense depth from DSIs generated using events and input camera poses. 
It achieves low computational memory and inference costs by splitting the DSI into multiple parts and processing them in parallel.

\color{black}

\ifclearlook\cleardoublepage\fi \begin{figure}[t]
\scriptsize
    \centering
            \begin{tabular}{*{5}{>{\centering\arraybackslash}p{0.15\linewidth}}} 
             Events (on frames) & DDES \cite{Tulyakov19iccv} & DTC-SPADE~\cite{Zhang22cvpr} & StereoFlow-Net~\cite{Cho24eccv} & GT\\
        \end{tabular}
    \includegraphics[trim={0cm 11.9cm 0cm 0cm},clip,width=\linewidth]{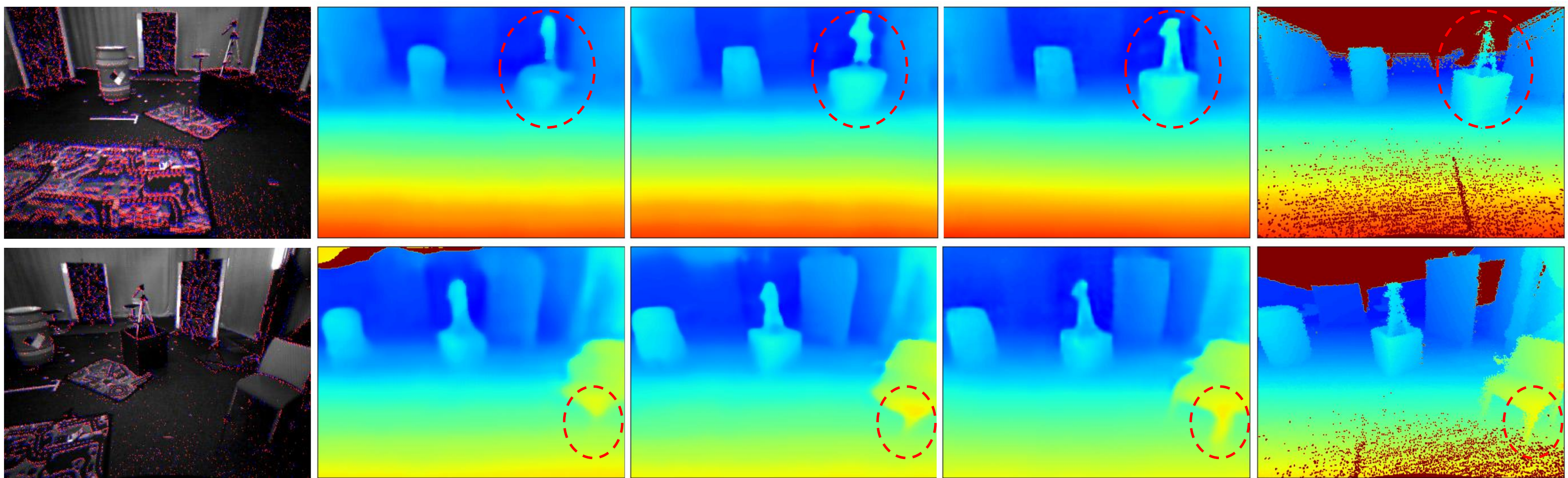}
    \caption{
    \descript{Depth estimated by learning-based methods on MVSEC \emph{indoor\_flying} data. 
    Adapted from \cite{Cho24eccv}.}}
    \label{fig:results:mvsec}
\end{figure}

\begin{figure}[t]
    \centering
        \scriptsize
        \begin{tabular}{*{5}{>{\centering\arraybackslash}p{0.15\linewidth}}} 
             Events & DTC-SPADE~\cite{Zhang22cvpr} & Conc-Net \cite{Nam22cvpr} & StereoFlow-Net~\cite{Cho24eccv} & Color image\\
        \end{tabular}
    \includegraphics[trim={0 0 0 0.7cm}, clip, width=\linewidth]{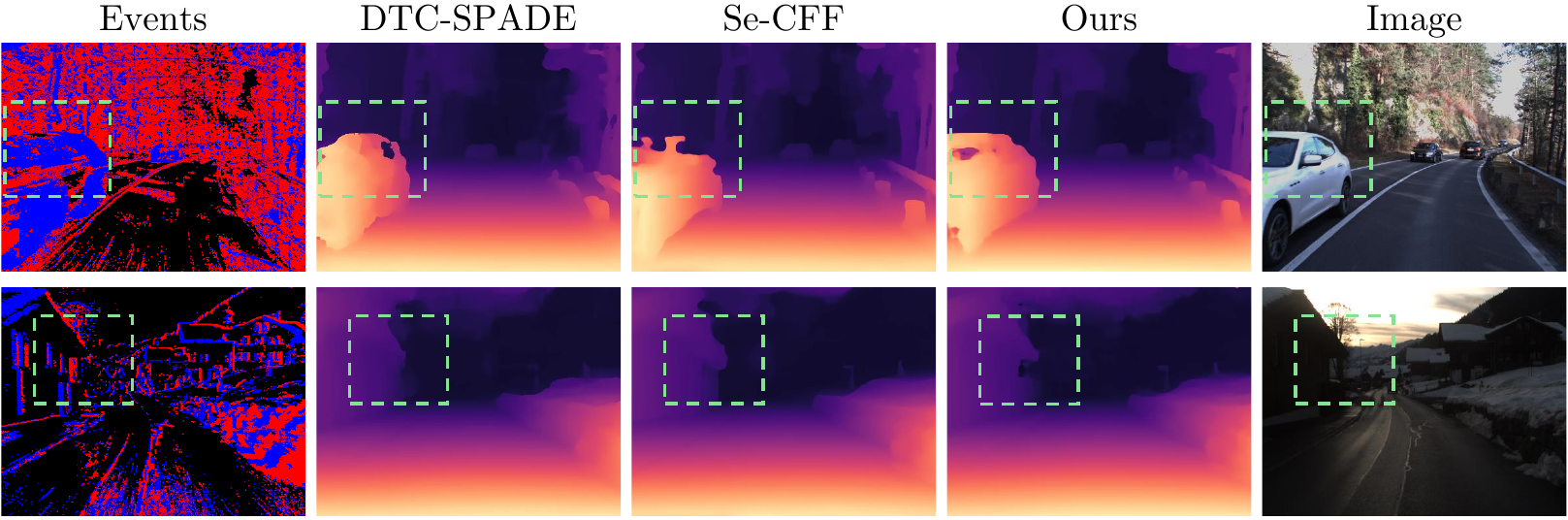}
    \caption{\descript{Disparity output of learning-based methods on DSEC test set (no available GT). 
    Adapted from \cite{Cho24eccv}.}}
    \label{fig:results:dsec}
\end{figure}

\subsection{Evaluation of Stereo Depth Estimation Methods}
\label{sec:eval}
\trend{
Early event-based depth estimation methods were evaluated on self-recorded data (and on specific hardware) that demonstrated the novel advantages of event cameras.
Despite the flourishing of much subsequent work, evaluation is still today largely ad-hoc.} 
\revise{
Using reported results, we compare methods of each (sub-)category described above:
\begin{itemize}
    \item \Cref{tab:tableAndreopoulos18cvpr} compares instantaneous model-based methods (Sec.\ref{sec:stereomethods:inst:model}) on latency and power consumption metrics \cite{Andreopoulos18cvpr}. 
    Since they have not been evaluated on any common datasets and public code is not available in many cases, it is not possible to benchmark their accuracy.
    \item \Cref{tab:quant:deepmethods} compares learning-based methods from Sec.\ref{sec:stereomethods:inst:learning} and \ref{sec:stereomethods:long:learning}. 
    They do not report latency or power consumption metrics. 
    \item \Cref{tab:model_eval_mvsec,tab:model_eval_dsec} compare long-term model-based methods (Sec. \ref{sec:stereomethods:long:model}). 
    They have been shown to produce real-time performance with DAVIS346 (QVGA resolution) cameras on standard CPUs.  
\end{itemize}
}

\subsubsection{Evaluation of Learning-based Methods}
\label{sec:learning_eval_instant}

\newcommand{\gi}[1]{
\IfDecimal{#1}{
\gradientcelld{#1}{9.7}{21}{32}{green}{yellow}{red}{40}}{\textbf{#1}}
}
\newcommand{\gii}[1]{
\IfDecimal{#1}{
\gradientcelld{#1}{49}{72}{95.1}{red}{yellow}{green}{40}}{\textbf{#1}}
}
\newcommand{\giii}[1]{
\IfDecimal{#1}{
\gradientcelld{#1}{0.3}{0.55}{0.8}{green}{yellow}{red}{40}}{\textbf{#1}}
}
\newcommand{\giv}[1]{
\IfDecimal{#1}{
\gradientcelld{#1}{4.8}{12}{20}{green}{yellow}{red}{40}}{\textbf{#1}}
}
\newcommand{\gv}[1]{
\IfDecimal{#1}{
\gradientcelld{#1}{0.8}{3.1}{5.4}{green}{yellow}{red}{40}}{\textbf{#1}}
}
\newcommand{\gvi}[1]{
\IfDecimal{#1}{
\gradientcelld{#1}{0.8}{1.25}{1.7}{green}{yellow}{red}{40}}{\textbf{#1}}
}

\begin{table*}[!ht]
\centering
\caption{
\descript{
Quantitative comparison of deep stereo methods (Sec.~\ref{sec:stereomethods:inst:learning} and \ref{sec:stereomethods:long:learning}) on MVSEC \emph{indoor\_flying} and DSEC datasets.} 
\methodol{
The column Modality indicates the type of input: 
stereo events (2E), stereo events and frames (2E + 2F), or stereo hybrid (1E + 1F).} 
\descript{
The figures of merit are color-coded from \colorbox{green!40}{good} to \colorbox{red!40}{poor}. 
Cells with ``-'' indicate data unavailability. 
Methods marked with $^*$ are unsupervised, whereas all other methods are supervised.
$^\dagger$ indicates long-term stereo method, whereas the rest are instantaneous.
$^\spadesuit$ represents Spiking Neural Network.
Metrics are collected from original articles.
}
}
\label{tab:quant:deepmethods}

\begin{adjustbox}{max width=\linewidth}

\begin{tabular}{lll|
>{\collectcell\gi}r<{\endcollectcell}
>{\collectcell\gi}r<{\endcollectcell}
>{\collectcell\gi}r<{\endcollectcell}|
>{\collectcell\gii}r<{\endcollectcell}
>{\collectcell\gii}r<{\endcollectcell}
>{\collectcell\gii}r<{\endcollectcell}
|l|
>{\collectcell\giii}r<{\endcollectcell}
>{\collectcell\giv}r<{\endcollectcell}
>{\collectcell\gv}r<{\endcollectcell}
>{\collectcell\gvi}r<{\endcollectcell}}

\toprule
& & &\multicolumn{6}{c|}{\textbf{MVSEC \emph{indoor\_flying}}} & & & & & \\
\cmidrule{4-9}
& & & Split 1 & Split 2 & Split 3 & Split 1 & Split 2 & Split 3 & &\multicolumn{4}{c}{\textbf{DSEC}} \\
\cmidrule{4-9}
\cmidrule{11-14}
\textbf{Method} & \textbf{Section} & \textbf{Modality} &\multicolumn{3}{c|}{\textbf{Mean depth error [cm] $\downarrow$}} &\multicolumn{3}{c|}{\textbf{One pixel error [\%] $\uparrow$}} 
& &{MAE [px] $\downarrow$} &{1PE [\%] $\downarrow$} &{2PE [\%] $\downarrow$} &{RMSE [px] $\downarrow$} \\
\cmidrule{1-9}
\cmidrule{11-14}
\tablecite{Tulyakov19iccv}{DDES} &\ref{sec:stereomethods:inst:learning:2E} &2E & 16.7 &29.4 &27.8 &89.8 &61 &74.8 & &0.576 &10.915 &2.905 &1.386 \\
\tablecite{Ahmed21aaai}{EIT-Net} &\ref{sec:stereomethods:inst:learning:2E} &2E &14.2 &- &19.4 &92.1 &- &89.6 & &- &- &- &- \\
\tablecite{Mostafavi21iccv}{ES} &\ref{sec:stereomethods:inst:learning:2E} &2E &13.27 &25.18 &25.72 &80.6 &73 &68.3 & &0.529 &9.958 &2.645 &1.222 \\
\tablecite{Zhang22cvpr}{DTC-SPADE} &\ref{sec:stereomethods:inst:learning:2E} &2E &13.5 &- &17.1 &93 &- &89.7 & &0.526 &9.270 &2.405 &1.285 \\
\tablecite{Nam22cvpr}{Conc-Net} &\ref{sec:stereomethods:inst:learning:2E} &2E &- &- &- &- &- &- & &0.520 &9.580 &2.620 &1.230 \\
\tablecite{Kweon21cvprw}{CES-Net} &\ref{sec:stereomethods:inst:learning:2E} &2E &- &- &- &- &- &- & &0.510 &9.366 &2.408 &1.170 \\
\tablecite{Liu22icra}{Liu et al} * &\ref{sec:stereomethods:inst:learning:2E} &2E &20 &25 &31 &85.1 &76.8 &81.4 & &- &- &- &- \\
\tablecite{rancon22access}{StereoSpike} $^\spadesuit$ & \ref{sec:stereomethods:inst:learning:2E} & 2E & 16.5 &- &18.4 &- &- &- & &- &- &- &-\\
\tablecite{Jianguo23iecon}{ASNet} & \ref{sec:stereomethods:inst:learning:2E} & 2E &20.46 &28.74 &22.15 &- &- &- & &- &- &- &-\\
\tablecite{Ghosh24jeswa}{Ghosh et al} &\ref{sec:stereomethods:inst:learning:2E} &2E &12.1 &- &15.6 &94.7 &- &92.9 & &- &- &- &- \\
\tablecite{Chen24spl}{Chen et al} & \ref{sec:stereomethods:inst:learning:2E} &2E &13.9 &- &14.6 &91.5 &- &91.9 & &0.499 &9.199 &2.343 &1.142 \\
\tablecite{Jiang24tim}{EV-MGDispNet} & \ref{sec:stereomethods:inst:learning:2E} &2E	&-	&-	&-	&-	&-	&-	& &0.514	&9.625	&2.512	&1.187\\
\tablecite{Cho24eccv}{StereoFlow-Net} & \ref{sec:stereomethods:inst:learning:2E} &2E &13	&-	&15	&92.9	&-	&92.6 & &0.493	&8.662	&2.259	&1.172\\
\tablecite{deOliveira25derdnet}{DERD-Net} $^\dagger$ & \ref{sec:stereomethods:long:learning} &2E & 11.69 & 11.11 & 12.28 &- &- &- & &- &- &- &- \\
\cmidrule{1-9}
\cmidrule{11-14}
\tablecite{Mostafavi21iccv}{EIS} & \ref{sec:stereomethods:inst:learning:2E+2F} &2E + 2F &13.74 &18.43 &22.36 &89 &85.2 &88.1 & &0.396 &5.814 &1.055 &0.905 \\
\tablecite{Cho22aaai}{SCM-Net} &\ref{sec:stereomethods:inst:learning:2E+2F} &2E + 2F &11.2 &- &14.5 &94.3 &- &92 & &- &- &- &- \\
\tablecite{Nam22cvpr}{Conc-Net}+ I & \ref{sec:stereomethods:inst:learning:2E+2F} &2E + 2F &- &- &- &- &- &- & &0.364 &4.844 &0.840 &0.818 \\
\tablecite{Cho22eccv}{SCS-Net} &\ref{sec:stereomethods:inst:learning:2E+2F} &2E + 2F &11.4 &- &13.5 &94.7 &- &94 & &0.390 &5.670 &0.990 &0.850 \\
\tablecite{Uddin22tcsvt}{N. Uddin et al} * &\ref{sec:stereomethods:inst:learning:2E+2F} &2E + 2F &19.7 &- &26.4 &87.4 &- &84.49 & &- &- &- &- \\
\tablecite{Cho23cvpr}{ADES} * &\ref{sec:stereomethods:inst:learning:2E+2F} &2E (+2F train) &- &- &- &- &- &- & &0.771 &18.370 &5.360 &1.698 \\
\tablecite{Zhao24eccv}{Zhao et al.} &\ref{sec:stereomethods:inst:learning:2E+2F} & 2E + 2F &9.7	&-	&11.1	&95.1	& &93.4 & &- &- &- &-\\
\cmidrule{1-9}
\cmidrule{11-14}
\tablecite{Zuo21iros}{HDES} &\ref{sec:stereomethods:inst:learning:1E+1F} &1E + 1F &16 &28 &18 &86.41 &49.7 &80.08 & &0.698 &19.020 &4.970 &1.307 \\
\tablecite{Wang21iros}{SHEF} + DCNet &\ref{sec:stereomethods:inst:learning:1E+1F} &1E + 1F &- &- &- &- &- &- & &0.587 &13.050 &2.850 &1.193 \\
\tablecite{Zhang22eccv2}{Zhang et al} &\ref{sec:stereomethods:inst:learning:1E+1F} &1E + 1F &15.8 &31.8 &19.7 &88.1 &50.3 &77.4 & &- &- &- &- \\
\tablecite{Chen24wacv}{SAFE} $^\dagger$ &\ref{sec:stereomethods:long:learning} &1E + 1F &- &- &- &- &- &- & &0.543 &10.970 &2.230 &1.175 \\
\bottomrule
\end{tabular}
\end{adjustbox}
\end{table*}

\descript{
\Cref{tab:quant:deepmethods} summarizes the performance comparison of learning-based methods.
Given their lack of explainability, their merit is mostly judged empirically (in this case, with prediction accuracy metrics).
Multi-modal datasets with ground truth depth acquired on moving platforms (like MVSEC and DSEC) are used for evaluation.}
\findin{
Interestingly, while such datasets are designed to foster research in VO/SLAM to enable autonomous robots (which implies long-term depth estimation), they have also been used to define benchmarks for instantaneous depth prediction (which are the majority of methods in \cref{tab:quant:deepmethods}).
}

\methodol{The evaluation on MVSEC (central columns of \cref{tab:quant:deepmethods}) is based on the protocol proposed in \cite{Tulyakov19iccv}.} 
\descript{Three indoor\_flying sequences are used in a three-fold cross validation or ``splits'' after removing the landing and take-off sections. 
Split 1 means that the method is tested on sequence 1 and trained on the other two sequences. 
Depth errors are computed using estimations from undistorted rectified events within 0.05s before and after each GT depth map obtained from the LiDAR (at 20Hz). 
Many of the methods do not report results on split~2, citing \challeng{poor generalization} because of the difference in dynamic characteristics in training and testing events on that split, as mentioned in \cite{Tulyakov19iccv,Ahmed21aaai}.}

\methodol{
The evaluation on the right part of \cref{tab:quant:deepmethods} is based on the DSEC disparity benchmark} \bg{introduced as a competition during the 2021 CVPR Event-based Vision Workshop.}
\descript{Dense disparity outputs are evaluated on GT disparity maps available at 10Hz, obtained using LiDAR scans.}

\methodol{
The figures of merit (FOM) used for comparison are:
\begin{itemize}[wide, labelwidth=!, labelindent=0pt]
\item Mean depth error.
\item One pixel error: \% of pixels whose calculated disparity error is less than one pixel.
\item MAE: Mean absolute error of the disparity.
\item 1PE: 1-px-error, \% of GT pixels with disparity error $>$~1~px.
\item 2PE: 2-px-error, \% of GT pixels with disparity error $>$~2~px.
\item RMSE: Root mean square error of the disparity.
\end{itemize}
}

\methodol{The methods in \cref{tab:quant:deepmethods} are organized according to the type of sensors used for estimation (``Modality'' column) and chronologically.}
\descript{
StereoFlow-Net \cite{Cho24eccv} and Chen et al \cite{Chen24spl}
produce the best results among event-only methods (top part of \cref{tab:quant:deepmethods}), 
whereas Conc-Net with Intensity frames \cite{Nam22cvpr}, SCS-Net \cite{Cho22eccv} and Zhao et al.~\cite{Zhao24eccv} produce the best results for sensor fusion (middle rows of \cref{tab:quant:deepmethods}). 
\findin{As expected, there is a significant improvement in the FOMs when frames are used in combination with events, since they provide information in regions with no event data.} 
Among the hybrid stereo methods in our comparison (bottom part of \cref{tab:quant:deepmethods}), the stereo method SAFE \cite{Chen24wacv} produces the best performance in the DSEC benchmark \findin{by inferring camera motion and aggregating temporal information}.
}

\findin{
Overall, methods that aggregate temporal information while maintaining consistency \cite{Cho24eccv,Chen24spl,Chen24wacv} seem to perform better, implying there is value in looking at larger temporal contexts.}
\opport{We also observe a gap in the performance of unsupervised methods \cite{Liu22icra,Uddin22tcsvt,Cho23cvpr} compared to the best supervised methods, indicating potential for future research.}

\descript{
\Cref{fig:results:mvsec} illustrates sample depth maps estimated by recent data-driven methods in the literature on MVSEC data. 
Comparing the outputs of DDES~\cite{Tulyakov19iccv} (2019) to StereoFlow-Net~\cite{Cho24eccv} (2024) we observe a \trend{notable improvement in depth accuracy and sharpness of object boundaries through the years}.
Furthermore, disparity maps estimated on the DSEC benchmark, shown in \cref{fig:results:dsec}, depict similar progress in learning-based dense stereo methods on driving scenarios. 
\trend{Recent DNNs are getting better at resolving fine details in HDR conditions (bottom row).} 
However, \proscons{the estimation quality drops off significantly} \challeng{in pixels with low event counts} (image center in forward driving), which \opport{is still an open problem in event-based regression tasks.}
}

\subsubsection{Evaluation of Model-based Methods}
\newcommand{\mi}[1]{
\IfDecimal{#1}{
\gradientcelld{#1}{18}{40}{700}{green}{yellow}{red}{40}}{\textbf{#1}}
}
\newcommand{\mii}[1]{
\IfDecimal{#1}{
\gradientcelld{#1}{9}{12.3}{46}{green}{yellow}{red}{40}}{\textbf{#1}}
}
\newcommand{\miii}[1]{
\IfDecimal{#1}{
\gradientcelld{#1}{1.3}{3.8}{39}{green}{yellow}{red}{40}}{\textbf{#1}}
}
\newcommand{\miv}[1]{
\IfDecimal{#1}{
\gradientcelld{#1}{1.7}{4.2}{75}{green}{yellow}{red}{40}}{\textbf{#1}}
}
\newcommand{\mv}[1]{
\IfDecimal{#1}{
\gradientcelld{#1}{7.7}{12.7}{103}{green}{yellow}{red}{40}}{\textbf{#1}}
}
\newcommand{\mvi}[1]{
\IfDecimal{#1}{
\gradientcelld{#1}{13}{20.7}{89}{green}{yellow}{red}{40}}{\textbf{#1}}
}
\newcommand{\mvii}[1]{
\IfDecimal{#1}{
\gradientcelld{#1}{49}{85}{95}{red}{yellow}{green}{40}}{\textbf{#1}}
}
\newcommand{\mviii}[1]{
\IfDecimal{#1}{
\gradientcelld{#1}{62}{95}{98}{red}{yellow}{green}{40}}{\textbf{#1}}
}
\newcommand{\mix}[1]{
\IfDecimal{#1}{
\gradientcelld{#1}{69}{98}{99.2}{red}{yellow}{green}{40}}{\textbf{#1}}
}
\newcommand{\mx}[1]{
\IfDecimal{#1}{
\gradientcelld{#1}{0.06}{1.27}{14.5}{red}{yellow}{green}{40}}{\textbf{#1}}
}

\begin{table*}[ht]
\caption{
\descript{
Quantitative comparison of long-term model-based stereo methods (Sec. \ref{sec:stereomethods:long:model}) on MVSEC \emph{indoor\_flying} data.
Mean depth error values are provided individually for \emph{flying1}, \emph{flying2} and \emph{flying3} sequences, as in \cite{Tulyakov19iccv}.
The other metrics are averaged from the three sequences.
The methods are evaluated on 200 s of data (110 million events and 4000 GT depth maps). 
Each estimated depth map is computed using 1 s of event data ($\approx$0.55 million events).
Values for CopNet~\cite{Piatkowska17cvprw} and TSES~\cite{Zhu18eccv} are taken from \cite{Zhu18eccv}, whereas the others are from \cite{Ghosh22aisy}.
Each metric is independently color-coded with a gradient of \colorbox{green!40}{best} - \colorbox{yellow!40}{median} - \colorbox{red!40}{worst}.}
}
\label{tab:model_eval_mvsec}
\centering
\begin{adjustbox}{width=\linewidth}

\begin{tabular}{l|
>{\collectcell\mi}r<{\endcollectcell}
>{\collectcell\mi}r<{\endcollectcell}
>{\collectcell\mi}r<{\endcollectcell}|
>{\collectcell\mii}r<{\endcollectcell}
>{\collectcell\miii}r<{\endcollectcell}
>{\collectcell\miv}r<{\endcollectcell}
>{\collectcell\mv}r<{\endcollectcell}
>{\collectcell\mvi}r<{\endcollectcell}
>{\collectcell\mvii}r<{\endcollectcell}
>{\collectcell\mviii}r<{\endcollectcell}
>{\collectcell\mix}r<{\endcollectcell}
>{\collectcell\mx}r<{\endcollectcell}}

\toprule 
 Method & 
 \multicolumn{3}{c|}{\textbf{Mean depth error [cm]} $\downarrow$}
 & \text{Median Err} & \text{bad-pix} & \text{SILog Err} & \text{AErrR} & \text{log RMSE} & \text{$\delta < 1.25$} & \text{$\delta < 1.25^2$} & \text{$\delta < 1.25^3$} & \text{\#Points}\\

& \text{Split 1} & \text{Split 2} & \text{Split 3} & \text{[cm] $\downarrow$} & \text{[\%] $\downarrow$} & \text{$\times 100 \downarrow$} & \text{[\%] $\downarrow$} & \text{$\times 100 \downarrow$} & \text{[\%] $\uparrow$} & \text{[\%] $\uparrow$} & \text{[\%] $\uparrow$} & \text{[million] $\!\uparrow$}\\
\midrule

 EMVS \cite{Rebecq18ijcv} (monocular)  & 39.37 & 31.42 & 30.54 & 14.35 & 3.83 & 4.20 & 12.73 & 20.72 & 84.75 & 94.86 & 97.98 & 1.27\\
 CopNet~\cite{Piatkowska17cvprw} & 61.00 & 100.00 & 64.00 & - &- &- &- &- &- &- &- &- \\
 TSES~\cite{Zhu18eccv} & 36.00 & 44.00 & 36.00 & - &- &- &- &- &- &- &- &-  \\
  GTS~\cite{Ieng18fnins} & 700.37 & 167.14 & 299.48 & 45.42 & 38.44 & 74.47 & 102.92 & 89.07 & 49.55 & 62.18 & 69.36 & 0.06\\
 SGM (Time Surf.)~\cite{Hirschmuller08pami}  & 35.45 & 32.94 & 37.86 & 12.34 & 6.38 & 8.45 & 16.16 & 29.48 & 85.34 & 93.05 & 96.03 & 14.46\\
 ESVO~\cite{Zhou20tro}    & 23.39 & 20.42 & 24.29 & 9.83 & 2.83 & 3.02 & 9.58 & 17.53 & 91.82 & 96.49 & 98.38 & 1.56\\
 MC-EMVS~\cite{Ghosh22aisy} & 22.53 & 18.20 & 19.49 & 9.53 & 1.35 & 1.71 & 7.79 & 13.23 & 95.03 & 98.07 & 99.21 & 0.81\\

\bottomrule
\end{tabular}
\end{adjustbox}
\end{table*}

\newcommand{\dsi}[1]{
\IfDecimal{#1}{
\gradientcelld{#1}{289}{350}{517}{green}{yellow}{red}{40}}{\textbf{#1}}
}
\newcommand{\dsii}[1]{
\IfDecimal{#1}{
\gradientcelld{#1}{66}{82}{163}{green}{yellow}{red}{40}}{\textbf{#1}}
}
\newcommand{\dsiii}[1]{
\IfDecimal{#1}{
\gradientcelld{#1}{5.9}{9.1}{14}{green}{yellow}{red}{40}}{\textbf{#1}}
}
\newcommand{\dsiv}[1]{
\IfDecimal{#1}{
\gradientcelld{#1}{6.5}{11}{33}{green}{yellow}{red}{40}}{\textbf{#1}}
}
\newcommand{\dsv}[1]{
\IfDecimal{#1}{
\gradientcelld{#1}{11.6}{21}{60}{green}{yellow}{red}{40}}{\textbf{#1}}
}
\newcommand{\dsvi}[1]{
\IfDecimal{#1}{
\gradientcelld{#1}{25}{33.4}{60}{green}{yellow}{red}{40}}{\textbf{#1}}
}
\newcommand{\dsvii}[1]{
\IfDecimal{#1}{
\gradientcelld{#1}{82}{87.3}{92}{red}{yellow}{green}{40}}{\textbf{#1}}
}
\newcommand{\dsviii}[1]{
\IfDecimal{#1}{
\gradientcelld{#1}{87}{93}{95}{red}{yellow}{green}{40}}{\textbf{#1}}
}
\newcommand{\dsix}[1]{
\IfDecimal{#1}{
\gradientcelld{#1}{89}{95.4}{97}{red}{yellow}{green}{40}}{\textbf{#1}}
}
\newcommand{\dsx}[1]{
\IfDecimal{#1}{
\gradientcelld{#1}{1.6}{2.34}{9.4}{red}{yellow}{green}{40}}{\textbf{#1}}
}

\begin{table*}[ht]
\caption{\descript{
Quantitative comparison of long-term model-based stereo methods (Sec.~\ref{sec:stereomethods:long:model}) on DSEC \emph{zurich\_city\_04a} sequence with a maximum range of 50~\si{\meter}. 
Values are taken from \cite{Ghosh24eccvw}, 
which evaluates on 35s of stereo data, consisting of 635 million events and 350 GT depth maps.
Each depth map is estimated using 0.2s of event data ($\approx3.5$ million events).
Each metric is independently color-coded with a gradient of \colorbox{green!40}{best} - \colorbox{yellow!40}{median} - \colorbox{red!40}{worst}.
}}
\label{tab:model_eval_dsec}
\centering
\begin{adjustbox}{width=\textwidth}
\begin{tabular}{l|
>{\collectcell\dsi}r<{\endcollectcell}
>{\collectcell\dsii}r<{\endcollectcell}
>{\collectcell\dsiii}r<{\endcollectcell}
>{\collectcell\dsiv}r<{\endcollectcell}
>{\collectcell\dsv}r<{\endcollectcell}
>{\collectcell\dsvi}r<{\endcollectcell}
>{\collectcell\dsvii}r<{\endcollectcell}
>{\collectcell\dsviii}r<{\endcollectcell}
>{\collectcell\dsix}r<{\endcollectcell}
>{\collectcell\dsx}r<{\endcollectcell}}
\toprule 
 Method & \text{Mean Err} & \text{Median Err} & \text{bad-pix} & \text{SILog Err} & \text{AErrR} & \text{log RMSE} & \text{$\delta < 1.25$} & \text{$\delta < 1.25^2$} & \text{$\delta < 1.25^3$} & \text{\#Points}\\

& \text{[cm] $\downarrow$} & \text{[cm] $\downarrow$} & \text{[\%] $\downarrow$} & \text{$\times 100 \downarrow$} & \text{[\%] $\downarrow$} & \text{$\times 100 \downarrow$} & \text{[\%] $\uparrow$} & \text{[\%] $\uparrow$} & \text{[\%] $\uparrow$} & \text{[million] $\!\uparrow$}\\
\midrule

EMVS~\cite{Rebecq18ijcv} (monocular) & 517.24 &	98.97 & 13.96 &	33.67 &	59.92 &	59.97 &	82.76 &	87.41 &	89.14 &	1.61\\
ESVO~\cite{Zhou20tro} &	393.15 &	162.45 & 10.53 & 8.30 &	17.65 &	28.90 &	84.36 &	92.80 &	96.04 &	9.39\\
MC-EMVS~\cite{Ghosh22aisy} & 313.18 & 66.16 & 7.68 &	14.07 &	24.55 & 37.84 & 90.37 & 93.27 & 94.84 &	2.38\\ 
ES-PTAM~\cite{Ghosh24eccvw} &	289.71 &	67.50 & 5.97 &	6.51 &	11.60 & 25.54 &	91.54 &	94.84 &	96.55 &	2.31\\

\bottomrule
\end{tabular}
\end{adjustbox}
\end{table*}

\bg{Hand-crafted methods for event-based depth estimation have been around since the 1990's \cite{Mahowald92thesis} (\cref{tab:stereomethods}).
Some of these methods (up to 2018) were compared theoretically and empirically in \cref{tab:tableAndreopoulos18cvpr}. 
\proscons{However, due to the lack of common public benchmarks at that time, most of them were evaluated on diverse self-collected datasets, often with very limited ego-motion.} %
\trend{Since the availability of datasets like MVSEC and DSEC, it has become easier to compare depth estimation methods.} 
\Cref{tab:model_eval_mvsec,tab:model_eval_dsec} summarize the evaluation of recent model-based methods on these standard datasets.}
\findin{Interestingly, most of the methods in these tables are designed for VO/SLAM and produce a semi-dense output depth map, using camera motion as an additional input.
Such model-based long-term depth estimation methods have been developed in parallel to data-driven instantaneous ones and have defined their own benchmarks.}

\descript{
\Cref{tab:model_eval_mvsec} collects the evaluation on the MVSEC indoor\_flying sequences.
As suggested in ESVO, 1s observation windows are used to propagate and fuse depth obtained by instantaneous methods like GTS and SGM.
The table shows a comprehensive set of ten standard metrics borrowed from the rich body of literature on visual depth prediction.
\findin{There is an accuracy-completion trade-off, so it is important to use FOMs that report both.}
\methodol{The table reports: 
mean and median errors between predicted and GT depths (median errors are robust to outliers), 
the number of reconstructed points, 
the number of outliers (bad-pix \cite{Geiger12cvpr}), 
the scale invariant depth error (SILog Err), 
the sum of absolute value of relative differences in depth (AErrR), 
and $\delta$-accuracy values on the percentage of points whose depth ratio with respect to GT is within some threshold (see \cite{Ye19arxiv}).}
Regarding the ranking, MC-EMVS performs overall best, closely followed by ESVO. 
SGM on time surfaces reconstructs \proscons{many more points (i.e., higher completion) but is less accurate}. 
More details are provided in \cite{Ghosh22aisy}.
}

\descript{
\Cref{tab:model_eval_dsec} shows the corresponding evaluation on one sequence of the DSEC dataset, using the same ten metrics as in \cref{tab:model_eval_mvsec}. 
Given the large amount of events produced by the VGA resolution cameras, the observation window is reduced to 0.2 seconds.
\proscons{The advantages of stereo over monocular (e.g., \cite{Ghosh22aisy} vs \cite{Rebecq18ijcv}) due to exploiting spatial parallax} are consistently clear in both tables: 
mean errors decrease by 30--45\%, and outliers also decrease (by more than half) while the number of recovered points remains. %
}

\findin{
Finally, it is worth noting that many stereo VO/SLAM systems do not directly report the performance of their depth-estimation (i.e., mapping) modules.} 
\descript{Instead they provide an alternative albeit indirect evaluation in terms of pose errors that does not require access to GT depth. 
Ideally, VO/SLAM systems should characterize the quality of their localization and mapping modules individually. 
However, because ($i$) both modules operate in an intertwined way (depth errors affect camera pose errors, and vice-versa), 
and ($ii$) GT localization information is considerably more compact (6-DOF) and easier to acquire than \challeng{accurate GT depth},
the result is that depth estimation errors are subsumed in the evaluation of camera trajectory errors \cite{Zhou20tro,Hadviger21ecmr,Wang23ral}.}

\descript{While it may seem that mean depth errors in \cref{tab:quant:deepmethods}-left and \cref{tab:model_eval_mvsec} are specified in the same format, note that they come from different observation windows and reconstructed points.
The dense disparity (depth) outputs from DNNs are evaluated on all pixels where GT depth is available, whereas the semi-dense outputs from model-based methods are evaluated on fewer pixels containing both input events and GT depth.
Hence, \proscons{a comparison should be made with caution}.
On MVSEC, \trend{learning-based instantaneous methods have considerable smaller errors (\cref{tab:quant:deepmethods}-left) than current model-based long-term methods.}
The gap may be due to the different observation windows and also to the fact that these \challeng{learning methods may be overfitting to the scenario} (MVSEC flying room), as they show \challeng{little generalization capabilities when they are applied to other datasets.} 
\proscons{Instead, model-based methods do not suffer from such an overfitting issue.}
Interpretation of DSEC values requires conversion of numbers, from disparity (\cref{tab:quant:deepmethods}-right) to depth (\cref{tab:model_eval_dsec}), which is not straightforward.
Nevertheless, \trend{a similar trend is expected}. 
}

\ifclearlook\cleardoublepage\fi \section{Summary of Insights}
\subsection{Trends}
\findin{The field of event-based stereo depth estimation is continuously evolving. 
Past research was dominated by instantaneous stereo pipelines from various labs working closer to the sensors (with access to the hardware and designing their own stereo rigs).
Current works include both instantaneous and long-term depth estimation (for SLAM) methods that are benchmarked on third-party public datasets recorded with the cameras mounted on vehicles, robots, etc. (i.e., moving camera scenario).} 
\trend{Even though the early works on stereopsis using event cameras championed the idea of fast asynchronous computations in dedicated analog hardware to mimic biological perception \cite{Mahowald92thesis}, most modern state-of-the-art algorithms are designed for CPUs with standard von Neumann hardware architectures, and recently GPUs (for deep learning), as these hardware are more commonplace.}

\findin{In the former category of asynchronous spike-based stereo,
\textbf{Cooperative Networks} are the dominant model as they can be run on efficient highly parallelized hardware with low-latency.}
\descript{They model 3D space using a grid with neurons that operate in a purely event-driven manner (without buffering) by matching timestamps of stereo events under epipolar constraints enforced with inter-neuronal connections.}
\trend{Through the years, there have been a constant flux of work on event-based cooperative stereo methods implemented on multiple specialized platforms like neuromorphic processors, FPGA, etc. as well as on general purpose computers.}
\proscons{However, results so far have been rather limited to simple scenes with low resolution cameras and constrained camera motion. 
The main promise of Cooperative Networks and SNN techniques lies in efficient (low power and fast) data processing on specialized neuromorphic platforms, but such ecosystem is not yet mature to tackle real-world scenarios robustly.}

\trend{In order to re-use established frame-based stereo matching pipelines on existing general purpose computers, there were initial efforts to convert sparse 4D event data (x-y-time-polarity) into 2D edge-like images.} %
\findin{Out of the many hand-crafted conversion strategies explored (e.g., 2D histograms, binary edge maps etc.), time surfaces seem to be the most successful one (i.e., \textbf{matching patches of (rectified) time surfaces} over epipolar lines), especially
for more complex scenes involving camera motion.
This is attributed to the fact that time surfaces include motion information and precise event generation timing, i.e., space-time information condensed into a 2D array compatible with traditional stereo methods.}
\proscons{However, time surfaces suffer in highly-textured scenes, where pixels are constantly overwritten by new events, causing loss of precise temporal information.
Grid-like 3D event representations \cite{Zhu19cvpr,Zhu18eccv,Ghosh22aisy} like DSIs have been shown to preserve finer details in depth estimation.}

\trend{Recent years have witnessed attempts to learn event representations for stereo matching using DNNs.}
\descript{Unlike SNNs that asynchronously process events, DNNs buffer events into multi-dimensional tensors that forego sparsity and are compatible with existing frame-based deep stereo architectures,} \findin{which form the backbone of current \textbf{deep-learning--based methods} for event-based stereo.
They reuse frame-based methods for feature extraction and matching, usually following the structure: 
1) feature extraction, 2) cost volume creation, 3) cost aggregation, and 4) cost refinement.}

\findin{The different DNNs primarily vary in the way input events are processed (sometimes in conjunction with intensity) to generate ``good-looking'' frames that are fed into the feature extraction module. 
This requires an input quantization phase, where events are converted into image-like representations (e.g., time surfaces, event stacks, or voxel grids \cite{Gallego20pami}) for compatibility.}
\descript{A list of the main grid-like representations used in event-based stereo matching is given in \cite{Bartolomei24eccv},} \opport{but finding the best representation is still an open question.}  
\findin{The key idea is to generate images suitable for stereo matching, either by reducing motion blur while filling up sparse areas (by using different event accumulation times) or reconstructing HDR frames (where standard cameras fail). 
\challeng{The motion dependency of events is seen as a nuisance} since the appearance of such images vary depending on the ego-motion, leading to variations in depth estimated for the same scene.
Deep learning is thus used to learn motion-invariant and illumination-invariant representations from raw events.}
\opport{A motion-decoupled event camera \cite{He24scirob} that always produces sharp, high frame-rate, HDR edge maps may be a better fit for such architectures.}

\descript{These methods estimate \textbf{dense} depth maps from stereo events,} 
\proscons{but using only (sparse) events to recover dense maps may produce erroneous results in} \challeng{regions without texture}, and dense maps are not always needed (sparse or semi-dense depth suffices for SLAM). 
\findin{To overcome this barrier, many methods try to generate data in textureless regions by using (or reconstructing) intensity frames and LiDAR.}

\descript{Most event-based deep stereo methods rely on high quality GT depth for supervised training, which \challeng{may not be readily available at the required spatial and temporal resolution.} 
In such cases, intensity images are used as an ancillary signal to train the networks in a self-supervised manner. 
\opport{There is a gap in the literature for end-to-end \textbf{deep unsupervised methods} that produce dense depth without using intensity frames or GT depth.}}

\opport{
There is a general need for more \textbf{explainability} in deep learning-based event stereo; current methods mostly rely on empirical analysis to get their point across, which limits theoretical understanding.
}

\trend{Most of the existing literature tackles the problem of instantaneous depth estimation. This is understandable since this task does not require the additional knowledge of the camera motion and can be used to showcase low-latency, high-speed (sparse) 3D reconstruction capabilities.}
\findin{However, for \textbf{static scenes}, the best results are obtained when depth estimates are fused over time to remove noise and improve accuracy. 
This is done via probabilistic filters in state-of-the-art stereo SLAM systems like ESVO and direct-ESVIO.
Yet, most deep learning methods for stereo depth estimation do not take \textbf{past information} into account when making predictions. 
They are mostly instantaneous, \descript{i.e., depth is estimated from small time windows at the current instance (i.e., no equivalent of ``depth fusion'').} 
Few recent efforts to add temporal context using attention~\cite{Chen24spl}, 3D scene flow~\cite{Cho24eccv} and SfM~\cite{Chen24wacv} have boosted performance by looking at local neighborhoods,} 
\proscons{but they still lack \challeng{global consistency} over longer time windows needed for \textbf{SLAM}}.
\opport{As future work, longer temporal context could be tightly incorporated into the network via camera motion inputs and recurrent connections}, to enable \appl{deep event-based stereo SLAM systems that include pose estimation, bundle adjustment and loop closure.}

\subsection{Pros and Cons of Event-based Stereo} 
\revise{
\textbf{Why event-based?}
The surveyed methods demonstrate that event cameras have clear advantages over their frame-based counterparts for stereo depth estimation.
One of the main advantages of using event cameras is that we can design algorithms to use precise spike timestamps for better stereo matching, thanks to their high temporal resolution.
Besides enabling efficient 3D perception in robotics, it can unlock research in novel scientific domains involving high speed imaging like reconstructing electric discharges %
\cite{Ilani22iaoc} and 3D fluid flow visualization \cite{Wang20eccv}.
}

\revise{
They also provide extensive hardware benefits like low power, low latency, low motion-blur and HDR sensing.
Depth estimation in HDR conditions is a property leveraged inherently by all event-based methods.
Without a global shutter, events are produced with minimal motion blur, which also translates to the estimated depth map. 
Thus, an event-based stereo pipeline has advantages over a frame-based pipeline during high-speed motion and/or HDR conditions; 
the latter would face issues not only because it takes non-trivial time for auto-exposure to kick in, but also because high exposure times in low light would cause motion blur.
This enables mapping scenes that are challenging for frame-based cameras, as shown in \cref{fig:hdr,fig:blur}.}

\begin{figure}[t]
    \centering
            \begin{tabular}{*{5}{>{\centering\arraybackslash}p{0.15\linewidth}}} 
             Intensity & Events & GT & E+I & I\\
        \end{tabular}
    \includegraphics[trim={0 0 0 220px},clip,width=\linewidth]{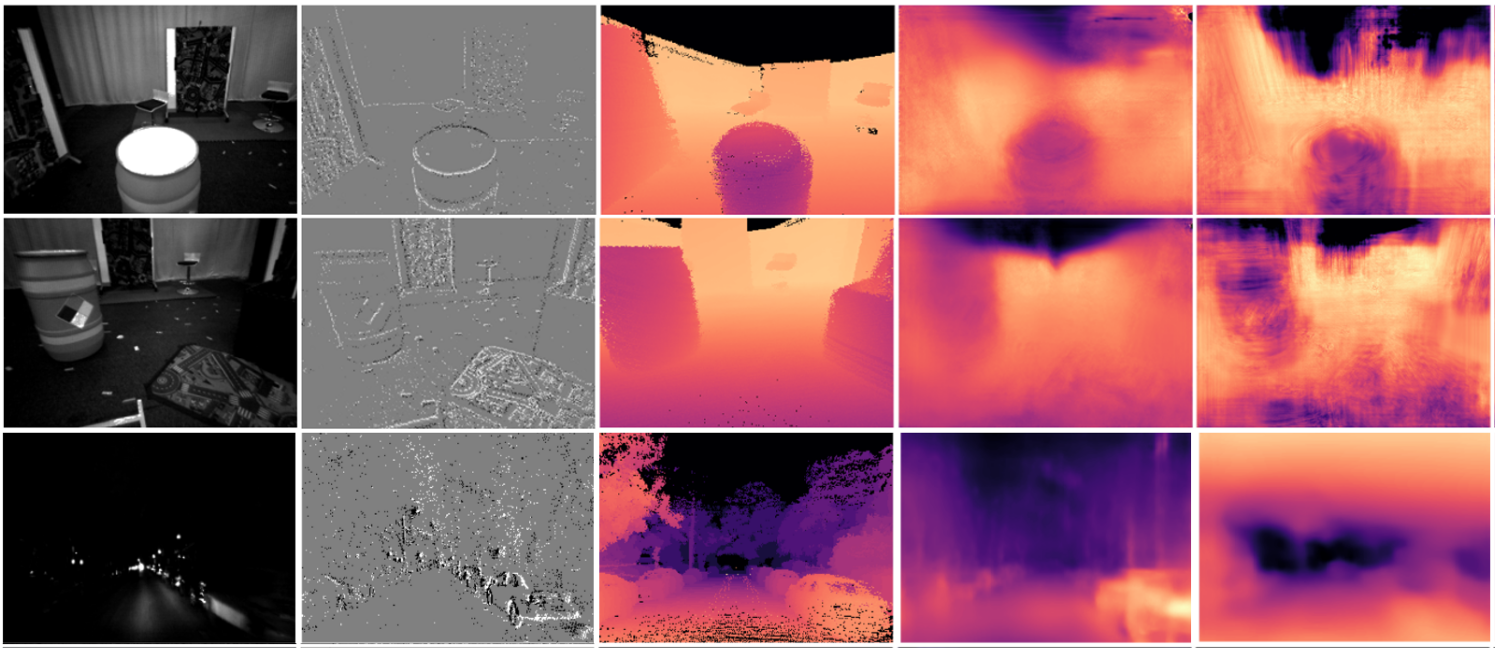}
    \caption{Comparing stereo depth estimated using Intensity images (I) and Events (E) by EIS \cite{Mostafavi21iccv}.
    Bottom: night scene.
    }
    \label{fig:EIS}
\end{figure}

\begin{figure}[t]
    \centering
    \begin{tabular}{*{4}{>{\centering\arraybackslash}p{0.2\linewidth}}} 
        Events & Intensity & E+I depth & I depth\\
    \end{tabular}
    \includegraphics[trim={0 660px 0 0},clip,width=\linewidth]{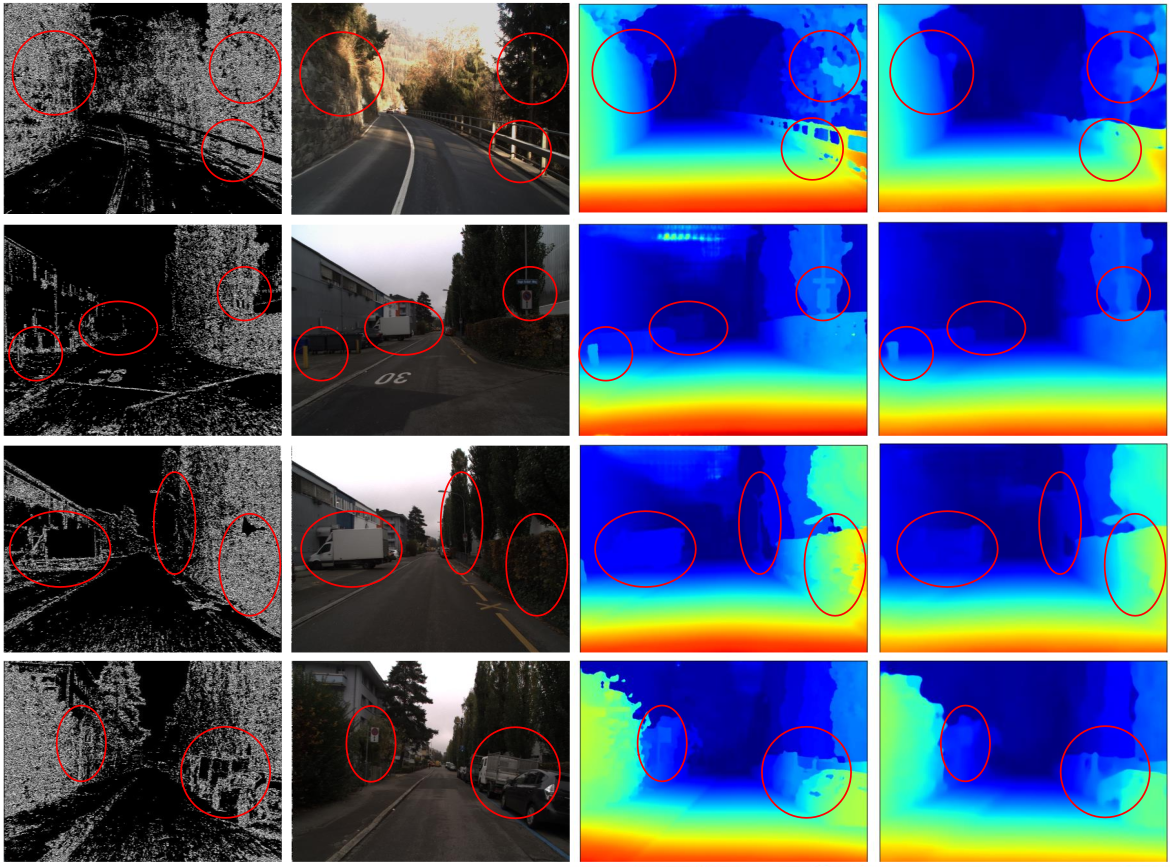}\\[0.5ex]
    \includegraphics[trim={0 720px 0 0},clip,width=\linewidth]{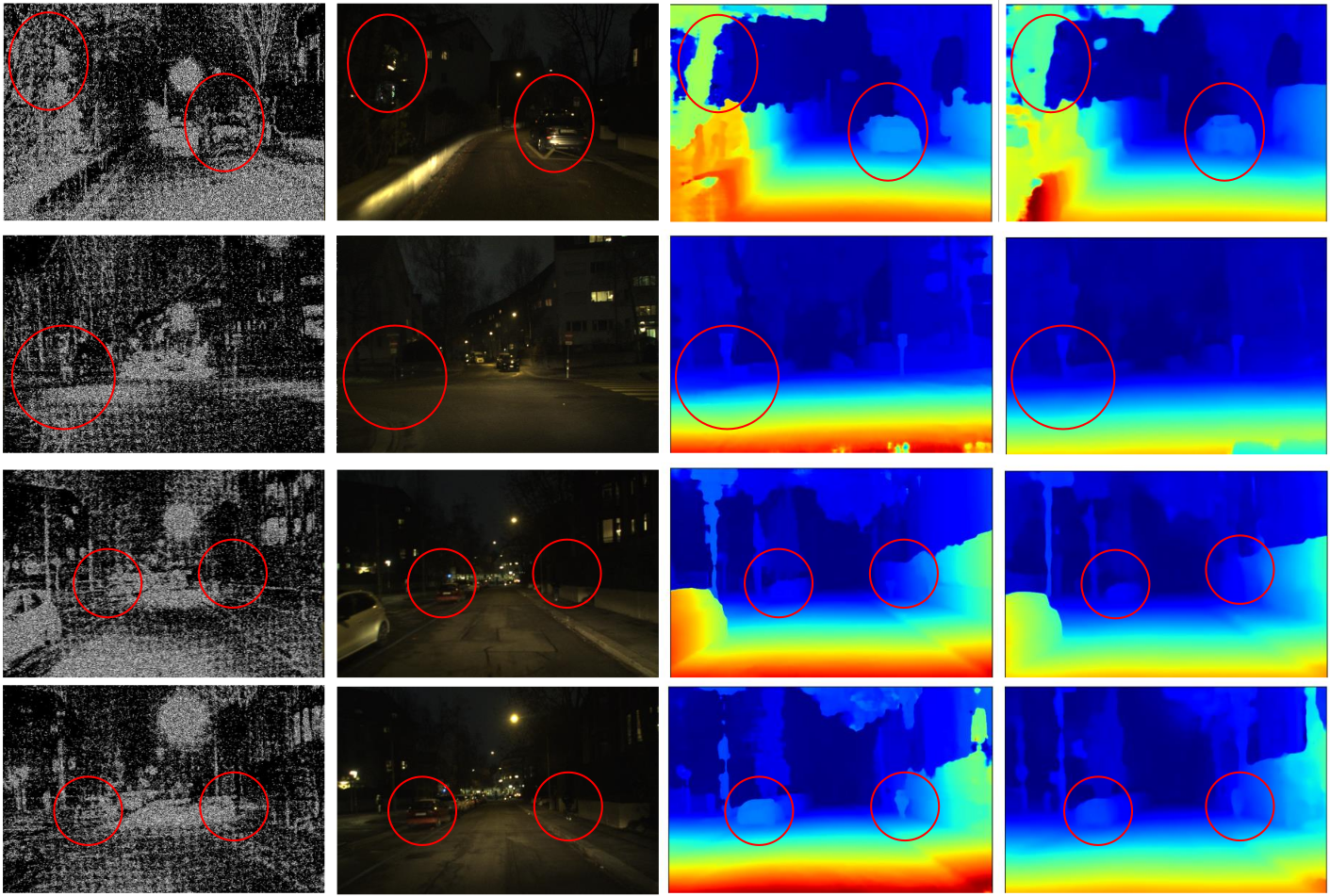}
    \caption{Comparing stereo depth estimated using Intensity images (I) and Events (E) by SCS-Net \cite{Cho22eccv}, in daylight (top) and at night (bottom).}
    \label{fig:SCSNet}
\end{figure}

\revise{
Moreover, events directly provide edge information which aid stereo matching by resolving high-fidelity depth discontinuity at object boundaries, which has been a problem in existing image-based stereo matching pipelines \cite{Cho22eccv,Zhao24eccv}. Deep event-based stereo pipelines \cite{Mostafavi21iccv,Cho22aaai,Cho22eccv} have demonstrated the benefits of using additional event-based information over pure intensity frames, both quantitatively and qualitatively.
For instance, SCS-Net~\cite{Cho22eccv} reduced the 1PE of DSEC~\cite{Gehrig21ral} sequences from 6.66 to 5.67 \% compared to the frame-based method GwC-Net \cite{Guo19cvpr}.
\Cref{fig:EIS,fig:SCSNet} demonstrate how events enhance stereo depth estimation, especially in low light (i.e., at night) \cite{Liu25pami,Shi23robio,Chen24arxiv}.}

\def\figWidth{0.315\linewidth}
\begin{figure}[t]
	\centering
    {\small
    \setlength{\tabcolsep}{2pt}
	\begin{tabular}{
	>{\centering\arraybackslash}m{\figWidth}
	>{\centering\arraybackslash}m{\figWidth}
	>{\centering\arraybackslash}m{\figWidth}}
	     Frame & Events & Depth from events\\
		\gframe{\includegraphics[trim={0px 130px 0 200px},clip,width=\linewidth]{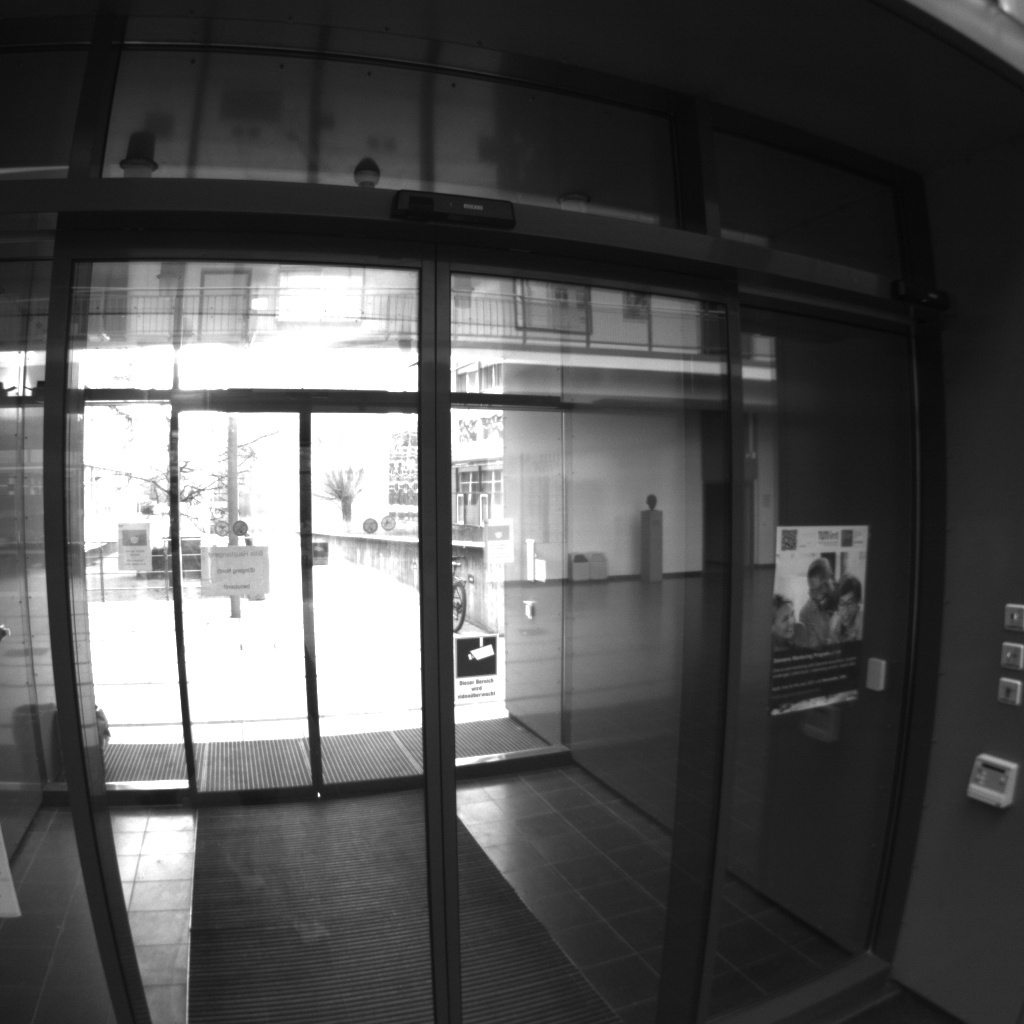}}
		&\gframe{\includegraphics[trim={0px 0 150px 0},clip,width=\linewidth]{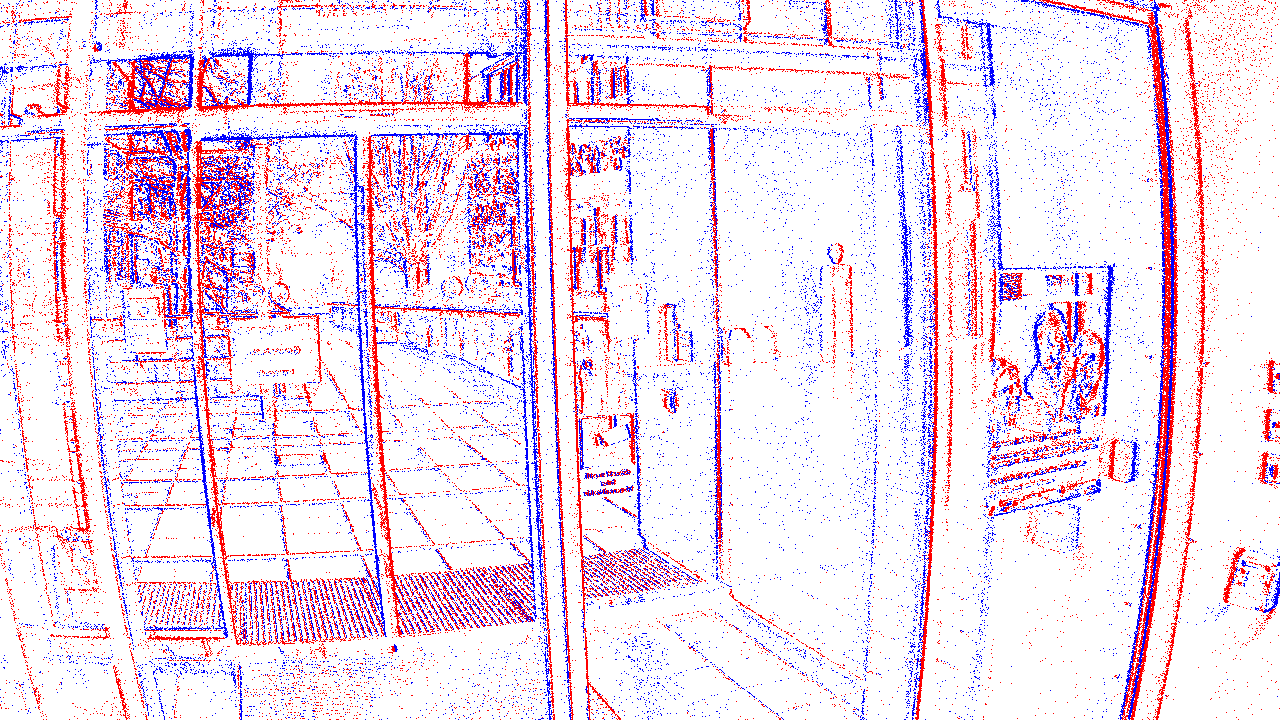}}
		&\gframe{\includegraphics[trim={0px 0 150px 0},clip,width=\linewidth]{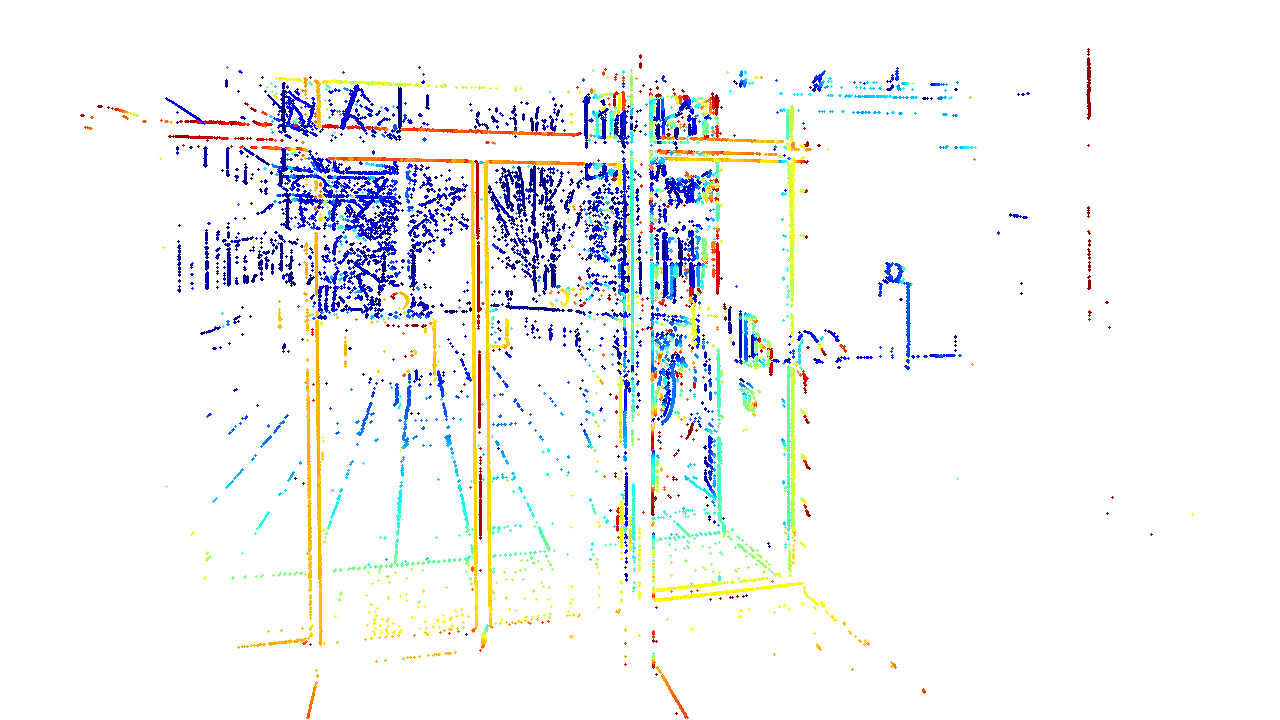}}
		\\
		
		\gframe{\includegraphics[trim={0px 150px 0 300px},clip,width=\linewidth]{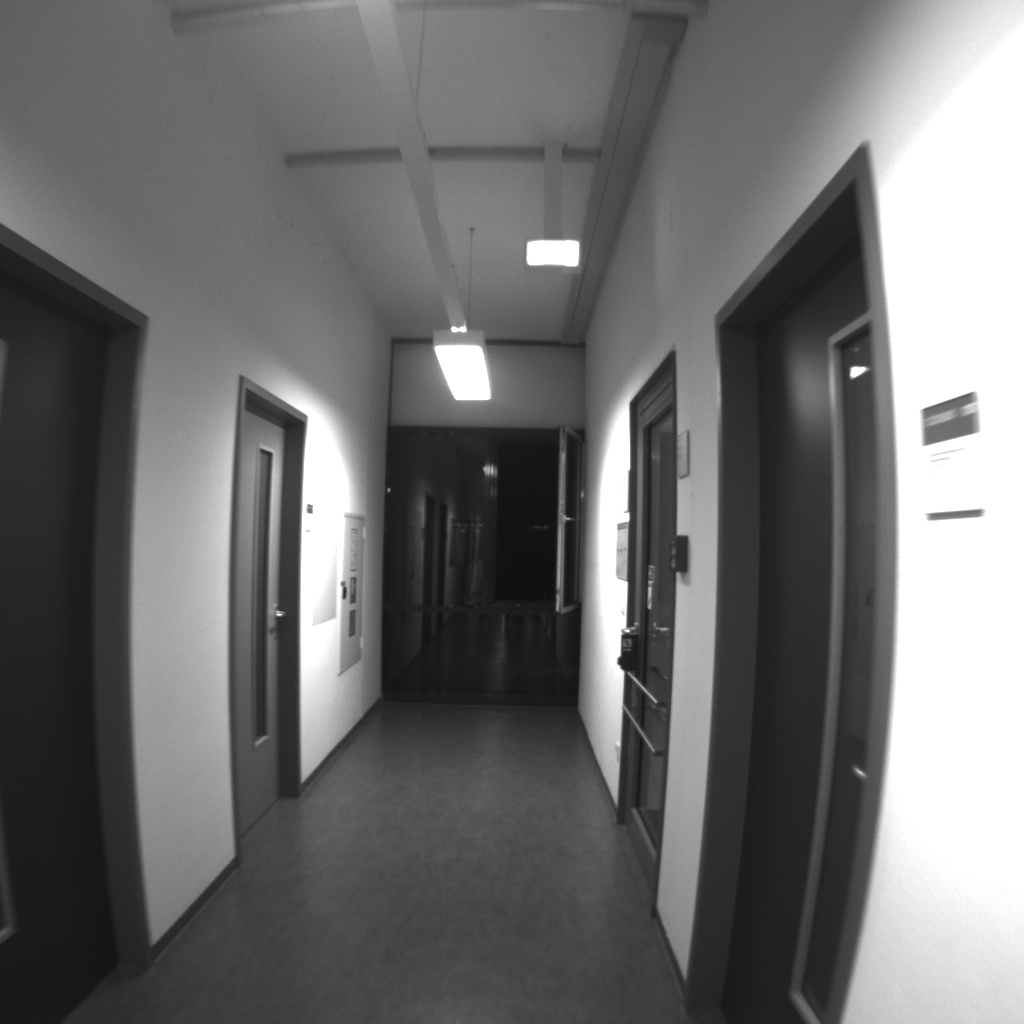}}
		&\gframe{\includegraphics[trim={0px 0 0 0},clip,width=\linewidth]{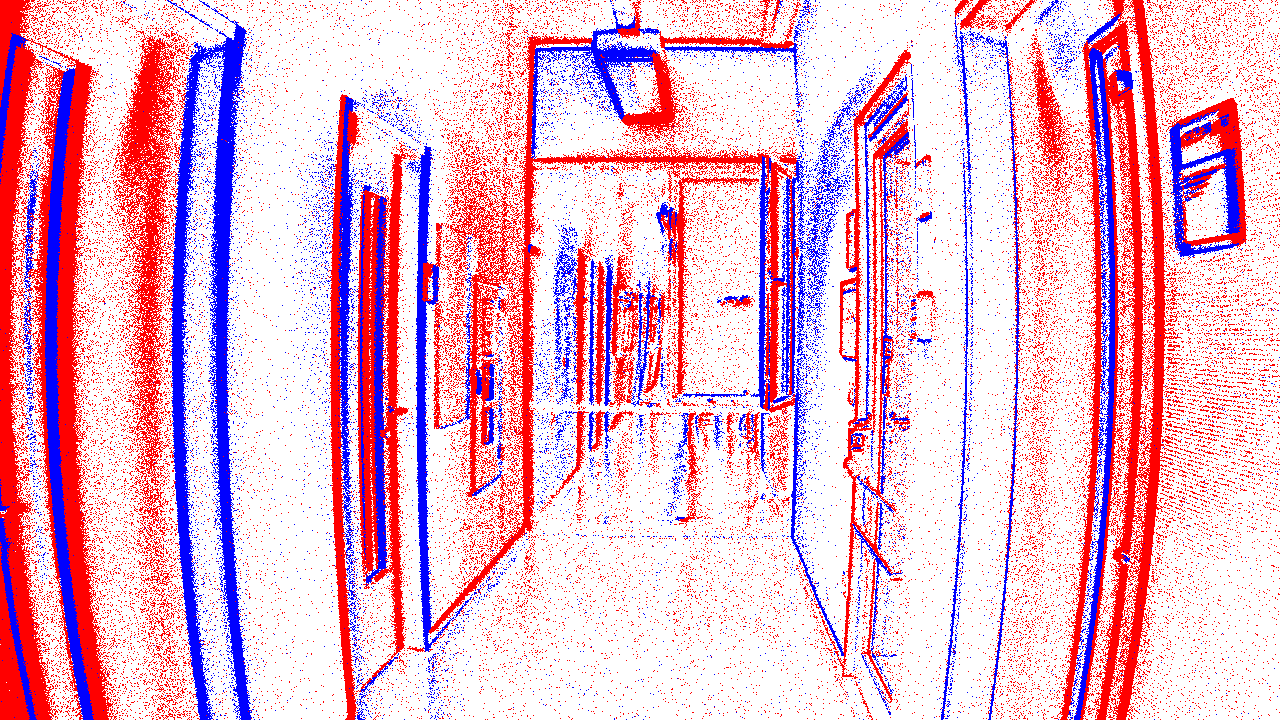}}
        &\gframe{\includegraphics[trim={0px 0 0 0},clip,width=\linewidth]{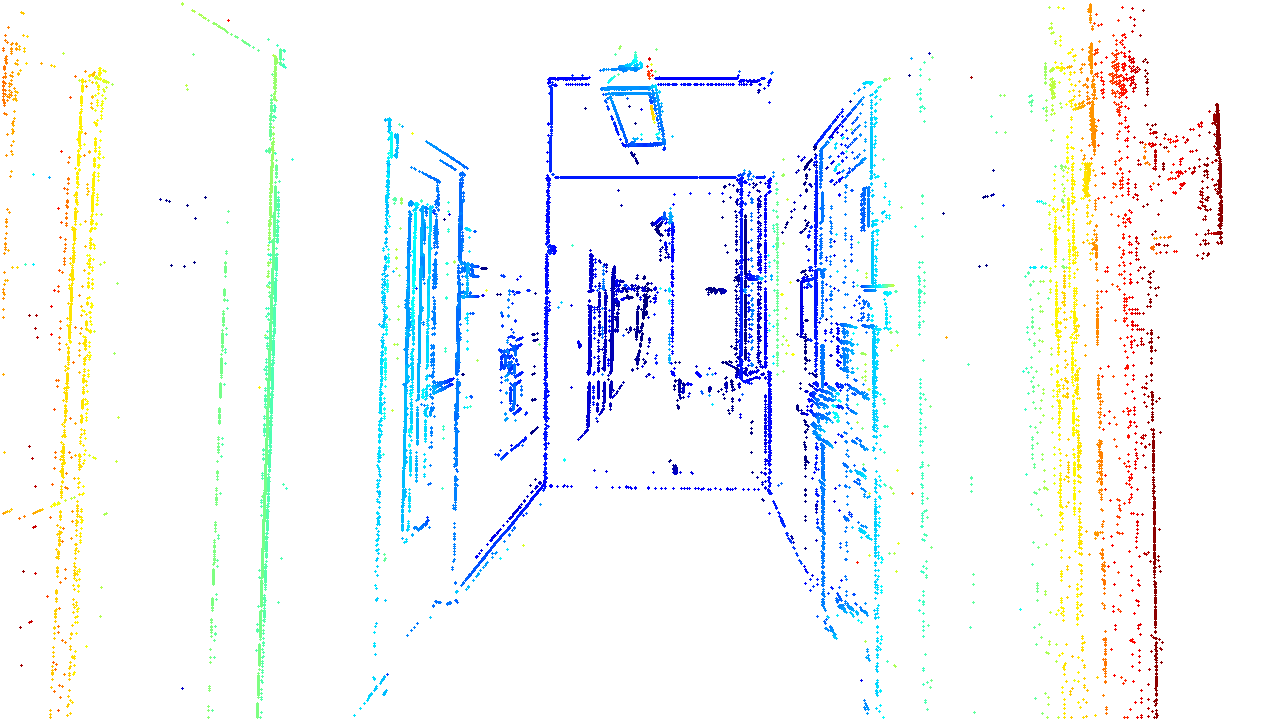}}
		\\
	\end{tabular}
	}
	\caption{\label{fig:hdr}
	\emph{HDR scenes.} Unlike frame-based cameras, event cameras can perceive both under- and over-exposed regions of the scene well, leading to good depth estimation throughout. Adapted from \cite{Ghosh22aisy}.
	}
\end{figure}

\begin{figure}[t]
	\centering
    {\small
    \setlength{\tabcolsep}{2pt}
	\begin{tabular}{
	>{\centering\arraybackslash}m{\figWidth}
	>{\centering\arraybackslash}m{\figWidth}
	>{\centering\arraybackslash}m{\figWidth}}
	     Frame & Events & Depth from events\\
		\gframe{\includegraphics[width=\linewidth]{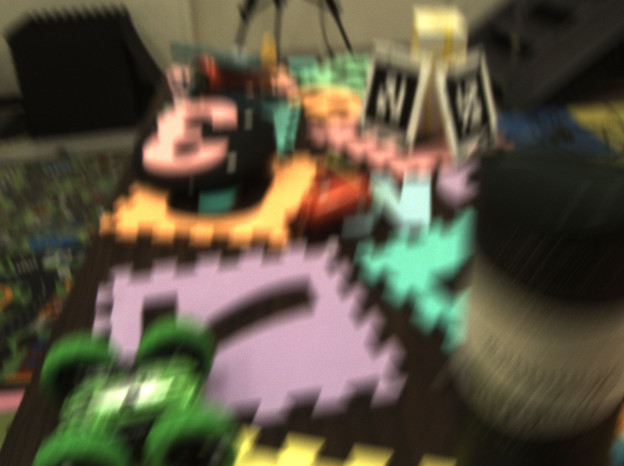}}
		&\gframe{\includegraphics[width=\linewidth]{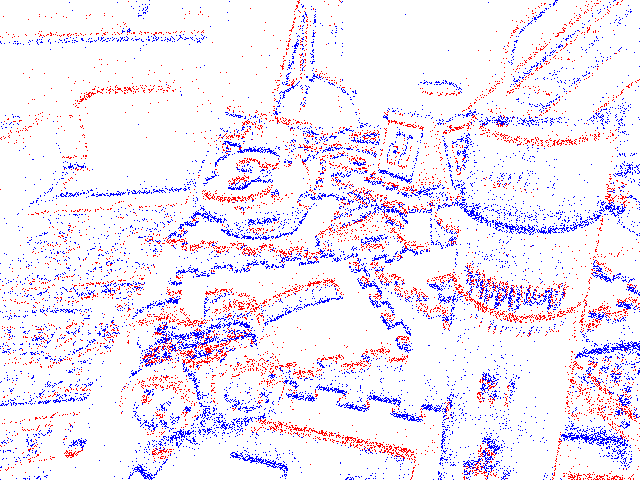}}
		&\gframe{\includegraphics[width=\linewidth]{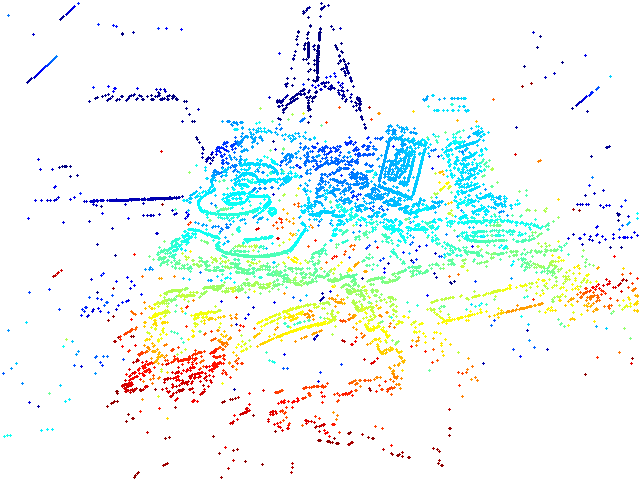}}
		\\
	\end{tabular}
	}
	\caption{\label{fig:blur}
	\emph{High-speed scene.} Unlike blurry frames, events enable 3D reconstruction during fast motion (here, with speeds between 1900 pix/s for objects in the far end and 3500 pix/s for objects close to the camera). 
    Adapted from \cite{Ghosh22aisy}.
	}
\end{figure}

\revise{
Algorithms using SNNs (running on neuromorphic processors) further leverage the sparsity and asynchronous nature of events for low latency and low power consumption, enabling efficient on-board deployment in robots.
Since a single event carries very little information, many algorithms buffer events before processing. 
Although this may introduce latency depending on the buffering window, the near-microsecond resolution of events allows 3D reconstruction at a very high rate.
For a frame-based camera to operate at a similar rate would require orders of magnitude higher bandwidth and power consumption \cite{Andreopoulos18cvpr,Gehrig24nature}.
}

\noindent
\revise{
\textbf{Why stereo?}
Event-based \emph{stereo} depth estimation has notable advantages over monocular (temporal) stereo \cite{Ghosh22aisy}:
higher accuracy, faster mapping, outlier removal and absolute scale recovery.
However, it is computationally more demanding, as double the events are processed, 
and it requires a more careful configuration and calibration.
}

\noindent\textbf{Challenges.} 
\findin{Most event-based stereo methods are based on some form of photometric consistency or time constancy assumption, 
which breaks down in case of \challeng{events not triggered by motion (but by flickering lights and noise)}.
Rampant flickering light (e.g., from LEDs) is especially problematic in urban settings \challeng{at night} because it generates an overwhelming number of noisy events. 
This may not only lead to undesired data drops due to \challeng{readout saturation}, but also overwhelm downstream algorithms leading to both high latency and poorer accuracy.
Due to the differential nature of the sensors, event-based stereo is also susceptible to other noise sources like reflections and glare.
}

\findin{\subsection{Evaluation and Benchmarking} 
\methodol{
By providing a comprehensive evaluation of stereo methods on common benchmarks}, this survey hopes to establish an accessible point of reference for future works in event-based stereo depth estimation. 
Our evaluation compilation shows that event-based stereo does not have an established benchmark like KITTI~\cite{Geiger13ijrr}. 
\opport{There is a need for concerted efforts towards defining benchmarks that could assess depth estimation over different observation windows (even on the same dataset).} 
\revise{Establishing useful benchmarks that will guide future research is challenging, since it is (i) dependent on specific applications and (ii) inherently speculative; it involves anticipating which benchmarks the community will adopt, often requiring both guesswork and foresight.}
Current benchmarks are still ``frame (snapshot)-based'' (due to the limitations in acquiring GT depth), even though event cameras unlock asynchronous depth estimation at high throughput.
Hence, works that focus on low-latency depth estimation often resort to qualitative evaluations. 
\opport{Additional tests on 3D point clouds and conversion of disparity benchmarks into actual depth (3D world) metrics would also help characterize the performance of an algorithm.}}

\findin{There is a lack of benchmarks specifically designed for instantaneous stereo methods. This need has been identified in the past \cite{Andreopoulos18cvpr,Steffen21frobt}, leading to proposal of a benchmark \cite{Steffen21frobt}, but it has not been widely adopted. 
The ego-motion datasets in computer vision and robotics communities have recently overshadowed previous efforts. 
Most such methods are evaluated on self-collected non-public data \trend{or more recently}, 
using SLAM datasets which include ego-motion and are designed for long-term stereo. 
\proscons{Although short instances of long-term mapping can be used for evaluating instantaneous methods, they do not target scenarios with stationary cameras and strong independent motion from dynamic objects.} 
\opport{This signals a need for benchmarks to foster steady progress in the ``instantaneous'' part.}
}

\trend{Since 2019, there has been some consistency in the evaluation data, protocol and metrics being reported,
allowing us to compare performance in~\cref{tab:quant:deepmethods,tab:model_eval_mvsec,tab:model_eval_dsec}.} 
\findin{It enables steady progress on the selected benchmark(s).
\opport{Yet, there remains gaps for metrics being reported for particular datasets.} 
Open source code is not available for many methods, which makes filling up these gaps harder. 
Moreover, many long-term stereo methods that are part of a SLAM pipeline do not report depth estimation errors; 
they evaluate performance via camera trajectory errors, making it harder to ablate the performance of individual tracking and mapping modules. 
Even though publication venues have strict page limits, %
\opport{supplementary materials providing populated tables with thorough evaluations would make it easier to compare to the existing state of the art while disseminating new work}.
}

\findin{Overall, good stereo event datasets are of tantamount importance not only to bolster objective results-driven research, but also to empower the promising rise of data-driven (deep learning) methods. 
\challeng{Recording a good stereo dataset with GT requires considerable engineering effort} (sensor synchronization and calibration, data validation, etc.). 
\opport{Providing easy-to-use benchmarks} (via APIs, good quality GT) also encourages its widespread use in the community.
While having established benchmarks helps ease comparison with existing works, \proscons{one caveat is that chosen benchmarks may not be representative and diverse enough (with respect to camera motions, scenes, etc.) for the real-world.} 
This may mislead us to chase metric superiority that does not translate into performance reliability in-the-wild, \challeng{limiting generalization capabilities of the algorithms} developed using them.
\organiz{Next we discuss existing stereo event datasets suitable for benchmarking depth estimation, most of which originate from the SLAM community.}
}

\ifclearlook\cleardoublepage\fi \begin{table*}[t]
\caption{
\descript{
Event camera datasets used for stereo depth estimation.}
\methodol{
The right part of the table describes the aspects of the recorded sequences, 
such as the number of sequences per scene (\# Seqs.), 
if they have grayscale frames (Frames), GT Poses (\LEFTcircle: Poses available partially, 3D: 3-DOF poses only), GT Depth, Outdoor scenes (Out), Indoor scenes (In) and Low light (LL).} 
\descript{`-' means no data available.
The last row is a dataset for hybrid (event-intensity) stereo.
\href{https://tinyurl.com/422ekj66}{Full spreadsheet link}.
\label{tab:datasets}}
}
\begin{adjustbox}{max width=\linewidth}
\setlength{\tabcolsep}{2.2pt}
\begin{tabular}{llllll|lrccccccccl}
\toprule 
\textbf{Dataset} & \textbf{Event sensors} & \textbf{Pixels} & \textbf{Baseline} & \textbf{FOV (V/H)} & \textbf{Events format} & \textbf{Sub-categories} & \textbf{Time [s]} & \textbf{\# Seqs} & \textbf{Frames} & \textbf{Poses} & \textbf{Depth} & \textbf{IMU} & \textbf{Out} & \textbf{In} & \textbf{LL} & \textbf{Motion type}\\
\midrule
{\textbf{RPG Stereo}}~\cite{Zhou18eccv} & 2 DAVIS240C & 240$\times$180 & 14.7 cm & 62.9$^\circ$ & ROS bag & indoor\_handheld & 257 & 8 & \CIRCLE & \CIRCLE & \Circle & \CIRCLE & \Circle & \CIRCLE & \Circle & 6-DOF handheld\\
(ECCV 2018) & 2 simulated & &  & 24.5 px &  &  simulated & 1 & 1 & \CIRCLE & \CIRCLE & \CIRCLE & \Circle & \Circle & \Circle & \Circle & 1D translation\\
\midrule
\tablecite{Zhu18ral}{\textbf{MVSEC}} & 2 mDAVIS346B & 346$\times$260 & 10 cm & 67$^\circ$/83$^\circ$ & ROS bag/ & indoor\_flying & 268 & 4 & \CIRCLE & \CIRCLE & \CIRCLE & \CIRCLE & \Circle & \CIRCLE & \Circle & 6-DOF hexacopter\\
(RAL 2018) &  &  &  &  & HDF5 & outdoor\_driving\_day & 915 & 2 & \CIRCLE & \CIRCLE & \CIRCLE & \CIRCLE & \CIRCLE & \Circle & \Circle & look-ahead driving\\
 &  &  &  &  &  & outdoor\_driving\_night & 916 & 3 & \CIRCLE & \CIRCLE  & \CIRCLE & \CIRCLE & \CIRCLE & \Circle & \CIRCLE & look-ahead driving\\
 &  &  &  &  &  & motorcycle & 1500 & 1 & \CIRCLE & \CIRCLE & \Circle & \CIRCLE & \CIRCLE & \Circle & \Circle & driving, tilted\\
\midrule
\tablecite{Gehrig21ral}{\textbf{DSEC}} & 2 Prophesee & 640$\times$480 & 60 cm & 60.1$^\circ$ & HDF5 & interlaken & 618 & 8 & \CIRCLE & \Circle & \CIRCLE & \CIRCLE & \CIRCLE & \Circle & \Circle & look-ahead driving\\
(RAL 2021) & Gen 3.1 &  &  &  &  & zurich\_city & 2499 & 42 & \CIRCLE & \Circle & \CIRCLE & \CIRCLE & \CIRCLE & \Circle & \CIRCLE & look-ahead driving\\
 &  &  &  &  &  & thun & 76 & 3 & \CIRCLE & \Circle & \CIRCLE & \CIRCLE & \CIRCLE & \Circle & \Circle & look-ahead driving\\
\midrule
\tablecite{Klenk21iros}{\textbf{TUM-VIE}} & 2 Prophesee & 1280$\times$720 & 11.84 cm & 65$^\circ$/90$^\circ$ & HDF5/ & mocap & 201 & 7 & \CIRCLE & \CIRCLE & \Circle & \CIRCLE & \Circle & \CIRCLE & \Circle & transl, 6-DOF handheld\\
(IROS 2021) & Gen 4 HD &  &  &  & TXT & running & 145 & 2 & \CIRCLE & \LEFTcircle & \Circle & \CIRCLE & \Circle & \CIRCLE & \CIRCLE & look-ahead+rotations, handheld\\
 &  &  &  &  &  & skate & 165 & 2 & \CIRCLE & \LEFTcircle & \Circle & \CIRCLE & \Circle & \CIRCLE & \CIRCLE & look-ahead+rotations, handheld\\
 &  &  &  &  &  & floor & 1189 & 5 & \CIRCLE & \LEFTcircle & \Circle & \CIRCLE & \Circle & \CIRCLE & \CIRCLE & walking, handheld\\
 &  &  &  &  &  & bike & 830 & 3 & \CIRCLE & \LEFTcircle & \Circle & \CIRCLE & \CIRCLE & \Circle & \Circle & look-ahead, head-mounted\\
 &  &  &  &  &  & slide & 196 & 1 & \CIRCLE & \LEFTcircle & \Circle & \CIRCLE & \Circle & \CIRCLE & \CIRCLE & high-speed, handheld\\
 &  &  &  &  &  & office & 160 & 1 & \CIRCLE & \LEFTcircle & \Circle & \CIRCLE & \Circle & \CIRCLE & \Circle & walking, handheld\\
\midrule
\tablecite{Gao22ral}{\textbf{VECtor}} & 2 Prophesee & 640$\times$480 & 17 cm & 67$^\circ$/82$^\circ$ & ROS bag/ & small-scale & 620 & 12 & \CIRCLE & \CIRCLE & \CIRCLE & \CIRCLE & \Circle & \CIRCLE & \CIRCLE & planar, 6-DOF handheld\\
(RAL 2022) & Gen 3 CD &  &  &  & HDF5 & large-scale & 637 & 6 & \CIRCLE & \CIRCLE & \CIRCLE & \CIRCLE & \Circle & \CIRCLE & \Circle & planar, walking, scooter riding\\
\midrule
\tablecite{Burner22evimo2}{\textbf{EV-IMO2v2}} & 2 Prophesee & 640$\times$480 & 22 cm & 70$^\circ$ (Proph.) & NPZ/ & SfM & 1117 & 62 & \CIRCLE & \CIRCLE & \CIRCLE & \CIRCLE & \Circle & \CIRCLE & \Circle & 6-DOF handheld\\
(arXiv 2022) & Gen 3 and  &  &  &  & TXT/ & SfM\_low\_light &236  & 11 & \CIRCLE & \CIRCLE & \CIRCLE & \CIRCLE & \Circle & \CIRCLE & \CIRCLE & 6-DOF handheld\\
 & 1 Samsung &  &  & 75$^\circ$ (Sams.) & ROS bag & sanity\_test & 418 & 31 & \CIRCLE & \CIRCLE & \CIRCLE & \CIRCLE & \Circle & \CIRCLE & \Circle & various rudimentary motions\\
 & Gen 3 &  &  &  &  & sanity\_test\_low\_light & 473 & 35 & \CIRCLE & \CIRCLE & \CIRCLE & \CIRCLE & \Circle & \CIRCLE & \CIRCLE & various rudimentary motions\\
\midrule
\tablecite{Chen23ral}{\textbf{HKU VIO}} & 2 DAVIS346 & 346$\times$260 & 6 cm & 55$^\circ$/69$^\circ$ & ROS bag & small-scale & 773 & 9 & \CIRCLE & \LEFTcircle & \Circle & \CIRCLE & \Circle & \CIRCLE & \CIRCLE & transl., rotation, 6-DOF drone\\
(RAL 2023) &  &  &  &  &  & large-scale & 2094 & 1 & \CIRCLE & \Circle & \Circle & \CIRCLE & \CIRCLE & \CIRCLE & \Circle & 6-DOF drone\\
\midrule
\tablecite{Chaney23cvprw}{\textbf{M3ED}} & 2 EVK4 & 1280$\times$720 & 12 cm & 38$^\circ$/63$^\circ$ & ROS bag/ & car & 6827 & 21 & \CIRCLE & \CIRCLE & \CIRCLE & \CIRCLE & \CIRCLE & \Circle & \CIRCLE & look-ahead driving\\
(CVPRW 2023) &  &  &  &  &  HDF5 & UAV & 2687 & 23  & \CIRCLE & \CIRCLE & \CIRCLE & \CIRCLE & \CIRCLE & \CIRCLE & \CIRCLE & 6-DOF drone flying\\
&  &  &  &  &  & legged & 2723 & 23 & \CIRCLE & \CIRCLE & \CIRCLE & \CIRCLE & \CIRCLE & \CIRCLE & \CIRCLE & quadruped walking, climbing stairs\\
\midrule
\tablecite{chen23tiv}{\textbf{ECMD}} & 2 DAVIS346 & 346$\times$260 & 30 cm & 48$^\circ$/61$^\circ$ & ROS bag & urban driving & 12078 & 81 & \CIRCLE & \CIRCLE & \CIRCLE & \CIRCLE & \CIRCLE & \Circle & \CIRCLE & look-ahead driving\\
(TIV 2023) & 2 DVXplorer & 640$\times$480 & 30 cm & 55$^\circ$/69$^\circ$ & & & & & & & & & & & \\
\midrule
\tablecite{hadviger23ecmr}{\textbf{SVLD}} & 2 DVXplorer & 640$\times$480 & 30 cm & & ROS bag & indoor & 420 & 8 &  \CIRCLE & \CIRCLE & \Circle & \CIRCLE & \Circle & \CIRCLE & \Circle & 6-DOF handheld\\
(ECMR 2023) & & & 57 cm & & & driving & 2760 & 7 & \CIRCLE & \CIRCLE & \Circle & \CIRCLE & \CIRCLE & \Circle & \CIRCLE & look-ahead driving \\
\midrule
\tablecite{Carmichael24ijrr}{\textbf{NSAVP}} & 2 DVXplorer & 640$\times$480 & 100 cm & 50$^\circ$/70$^\circ$ & HDF5 & urban driving & 10338 & 9 & \CIRCLE & \CIRCLE & \Circle & \CIRCLE & \CIRCLE & \Circle & \CIRCLE & look-ahead driving\\
(IJRR 2024) & & & & & & & & & & & & & & & \\
\midrule
\tablecite{wei24ijrr}{\textbf{FusionPortablev2}} & 2 DAVIS346 & 346$\times$260 & 73 cm & 67$^\circ$/83$^\circ$ & ROS bag & car & 4466 & 8 & \CIRCLE & \CIRCLE & \CIRCLE & \CIRCLE & \CIRCLE & \Circle & \Circle & high-speed look-ahead driving\\
(IJRR 2024) & & & 25 cm & & & handheld & 1274 & 6 & \CIRCLE & \CIRCLE (3D) & \CIRCLE & \CIRCLE & \CIRCLE & \CIRCLE & \Circle & 6-DOF walking \\
& & & & & & legged & 1336 & 5 & \CIRCLE & \CIRCLE (3D) & \CIRCLE & \CIRCLE & \CIRCLE & \CIRCLE & \Circle & jerky quadruped walking\\
& & & & & & UGV & 2011 & 8 & \CIRCLE & \CIRCLE & \CIRCLE & \CIRCLE & \CIRCLE & \CIRCLE & \Circle & low-speed  driving\\
\midrule
\tablecite{peng24arxiv}{\textbf{CoSEC}} & 2 EVK4 & 1280$\times$720 & - & - & - & driving & 3658 & 128 & \CIRCLE & \Circle & \CIRCLE & \CIRCLE & \CIRCLE & \Circle & \CIRCLE & look-ahead driving\\ 
(arXiv 2024) & & & & & & & & & & & & & & & \\

\midrule
{\textbf{SHEF}}~\cite{Wang21iros} & 1 Prophesee & 640$\times$480 & 6.54 cm & 46.8$^\circ$/60$^\circ$ & DAT/ & simple\_boxes & 753 & 21 & \CIRCLE & \CIRCLE & \CIRCLE & \Circle & \Circle & \CIRCLE & \Circle & 3D transl. using robot arm\\
(IROS 2021) & Gen 3 &  &  &  & TXT & complex\_boxes & 294 & 6 & \CIRCLE & \CIRCLE & \Circle & \Circle & \Circle & \CIRCLE & \Circle & \\
 &  &  &  &  &  & picnic & 377 & 6 & \CIRCLE & \CIRCLE & \Circle & \Circle & \Circle & \CIRCLE & \Circle & \\
\bottomrule
\end{tabular}

\end{adjustbox}
\end{table*}

\section{Datasets}
\label{sec:datasets}

\organiz{This section discusses stereo event datasets.
\ifarxiv 
We start by stating the requirements for a good dataset, and then describe the main publicly available ones.
\fi
\bg{Even though our main focus is stereo depth estimation, \findin{many of them are also used for benchmarking other tasks} like optical flow estimation, intensity image reconstruction, motion segmentation, semantic segmentation and camera localization.}
An overview is provided in \cref{tab:datasets} and \cref{fig:datasets}. 
\ifarxiv 
We conclude by discussing them, and listing good practices for new stereo event datasets.
\else
See the \textbf{supplementary material} for details on each dataset.
\fi
}

\ifarxiv \ifclearlook\cleardoublepage\fi \subsection{Requirements for a Good Stereo Dataset}
\label{sec:datasets:requirements}

\findin{Ideally, a good stereo dataset would comprise events and frames \challeng{perfectly aligned in space-time}, like in a DAVIS camera but with better quality frames and higher resolution (e.g., VGA or HD). 
There would be two or more of these devices (e.g., acting as left and right retinas).} 
\opport{However, such ``high-resolution DAVIS'' sensors (e.g. ALPIX-Edger\footnote{\url{https://alpsentek.com}}) are just starting to come up but are not yet widely available \cite{Guo23isscc,Kodama23isscc}}.
\findin{An alternative solution is using a beam splitter setup with separate frame-based and event-based cameras, like in \cite{peng24arxiv, Hidalgo22cvpr,Hamann22icprvaib,Tulyakov22cvpr}. 
\proscons{However, this limits the FOV of the sensors (59$^{\circ}$ HFOV/ 34$^{\circ}$ VFOV in \cite{Hidalgo22cvpr}) and the light incident on each pixel.} 
Without a beam splitter, many datasets try to place the frame-based and event-based cameras close to each other physically in the sensor rig such that far away points appear aligned on both pixel arrays.
Having good quality IMU and differential GPS measurements is a bonus.}

\findin{For evaluating and debugging SLAM pipelines, an ideal stereo dataset \challeng{should provide high quality GT poses and depth information in all sequences}. 
While motion capture systems provide accurate GT poses at a high rate (e.g., 200 Hz) compared to the hand-held speed of the motions (e.g., 6 Hz), they can produce many \challeng{IR noisy events}, which need to be filtered out (the best way is by using IR filters, to avoid generating them and risking interference and buffer saturation). 
Outdoors, GT pose could be provided with odometry via LiDAR, IMU, frame-based cameras or GPS.
Regarding GT depth, it is difficult to think of a sensor able to provide accurate ground truth at a rate that can be used to evaluate the depth provided by the asynchronous (i.e., high-speed) events of the camera (likewise for optical flow estimation). 
Current benchmarks evaluate depth at specific timestamps (about 10Hz) and interpolate (or are blind in the time between snapshots). 
}

\findin{The dataset \challeng{should also contain a good diversity of motions and scenarios for testing generalization capabilities}.
This involves recording in different speeds, scene textures and lighting conditions, with appropriate \challeng{bias tuning}.
The stereo baseline should be appropriate for the depth range of the scene being recorded (a rule of thumb is a baseline of 10\% of the expected depth range).}

\findin{Temporal alignment of all sensors is also critical due to high temporal resolution of event cameras.
Sensors should be \challeng{precisely calibrated} (ideally for each recording sequence) so they can be well-aligned in space and time. 
The quality of calibration should be validated and communicated using reprojection errors.}
\opport{Finally, making the data accessible by providing space-time--aligned (and redundant) measurements for each sensor in multiple commonly used data formats, along with convenient APIs for accessing, transforming and directly evaluating on the dataset is highly desirable.}
 \fi
\ifarxiv \def\figWidth{0.19\linewidth}
\begin{figure*}[ht!]
\centering
\begin{adjustbox}{max width=\linewidth}
    {\small
    \setlength{\tabcolsep}{1pt}
	\begin{tabular}{
	>{\centering\arraybackslash}m{0.4cm} 
	>{\centering\arraybackslash}m{\figWidth} 
    >{\centering\arraybackslash}m{\figWidth} 
	>{\centering\arraybackslash}m{\figWidth}
	>{\centering\arraybackslash}m{\figWidth}
    >{\centering\arraybackslash}m{\figWidth}
    >{\centering\arraybackslash}m{\figWidth}
    >{\centering\arraybackslash}m{\figWidth}
    >{\centering\arraybackslash}m{\figWidth}
    >{\centering\arraybackslash}m{\figWidth}
	}
        & UZH-RPG \cite{Zhou18eccv}
        & MVSEC \cite{Zhu18ral}
        & DSEC \cite{Gehrig21ral}
        & TUM-VIE \cite{Klenk21iros}
        & EV-IMO2 \cite{Burner22evimo2}
        & VECtor \cite{Gao22ral}
        & HKU VIO \cite{Chen23ral}
        & M3ED \cite{Chaney23cvprw}
        & SHEF \cite{Wang21iros} \\
        
        \rotatebox{90}{\makecell{Sensors}}
        & {\includegraphics[width=\linewidth]{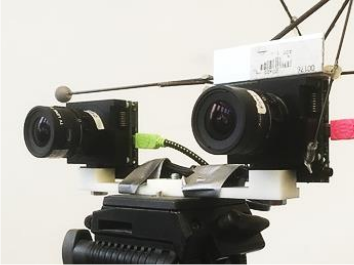}}
        & {\includegraphics[trim={0 .2cm 0 0},clip,width=\linewidth]{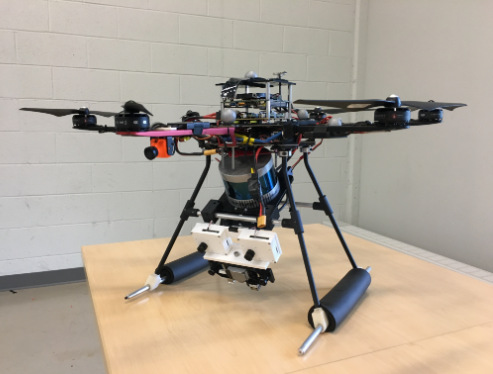}}
        & {\includegraphics[trim={1cm 0 .5cm 0},clip,width=\linewidth]{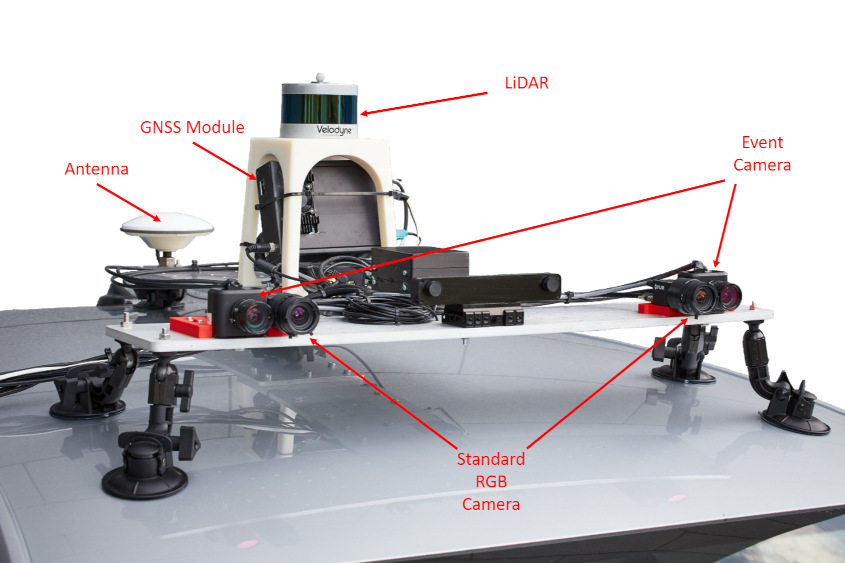}}
        & {\includegraphics[trim={0 6cm 0 3cm},clip,width=.65\linewidth]{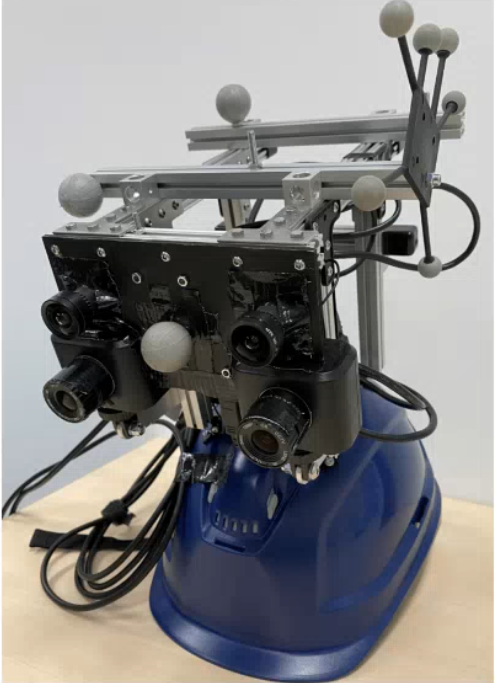}}
        & {\includegraphics[width=.75\linewidth]{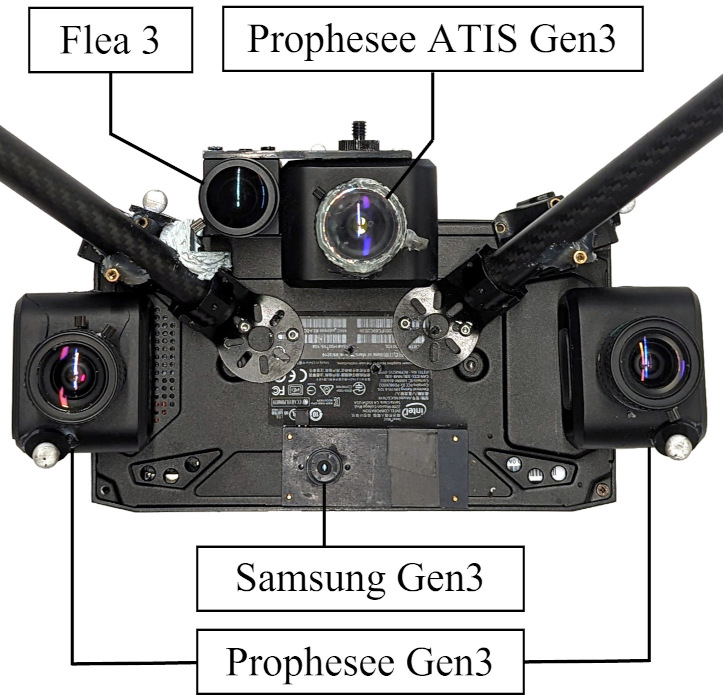}}
        & {\includegraphics[trim={1.2cm 0 1.2cm 0cm},clip,width=.65\linewidth]{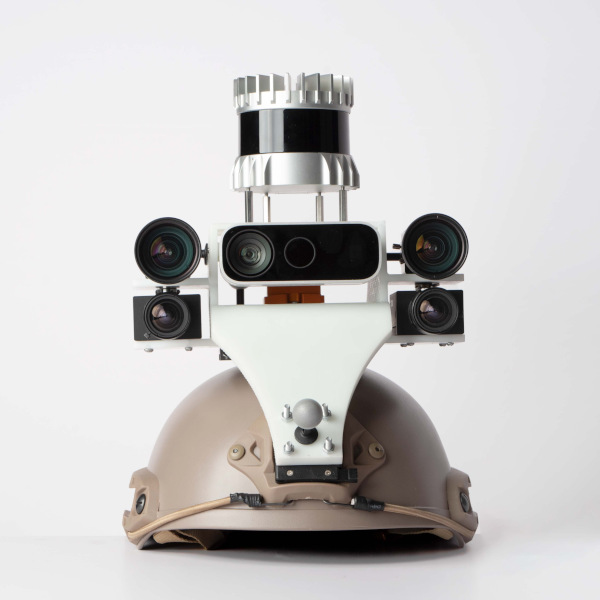}}
        & {\includegraphics[width=\linewidth]{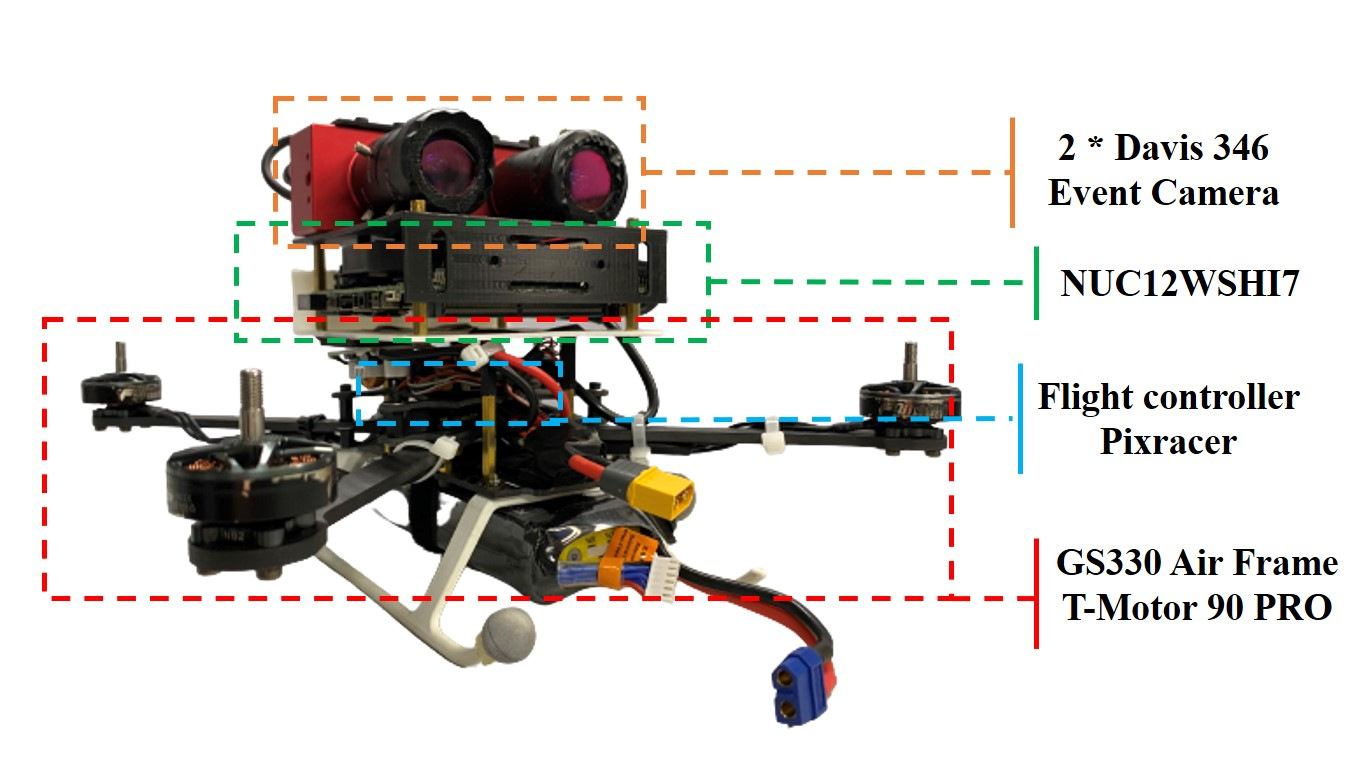}}
        & {\includegraphics[width=.65\linewidth]{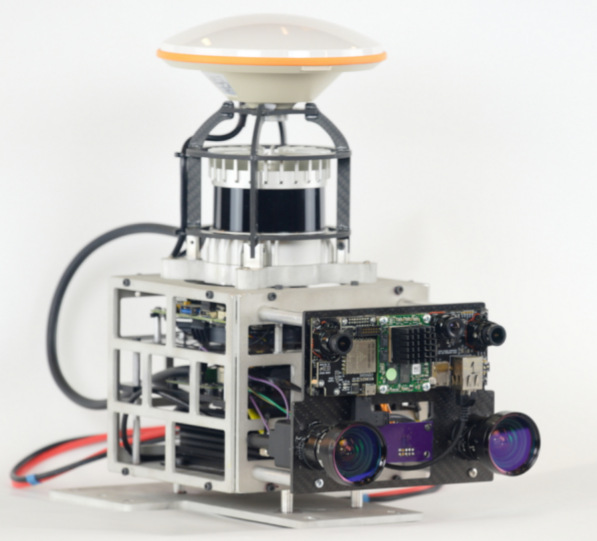}}
        & {\includegraphics[width=\linewidth]{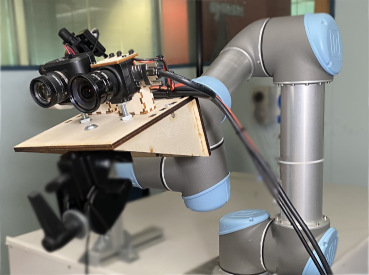}} \\

        \rotatebox{90}{\makecell{Frames}}
        & \gframe{\includegraphics[width=\linewidth]{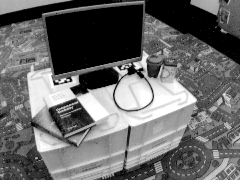}}
        & \gframe{\includegraphics[width=\linewidth]{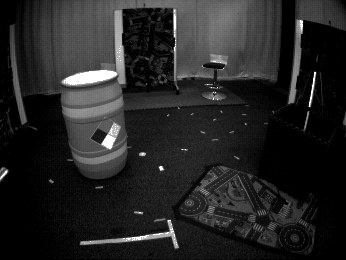}}
        & \gframe{\includegraphics[width=\linewidth]{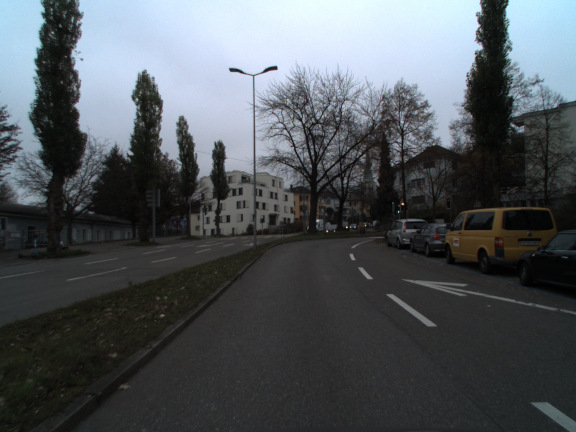}}
        & \gframe{\includegraphics[width=\linewidth]{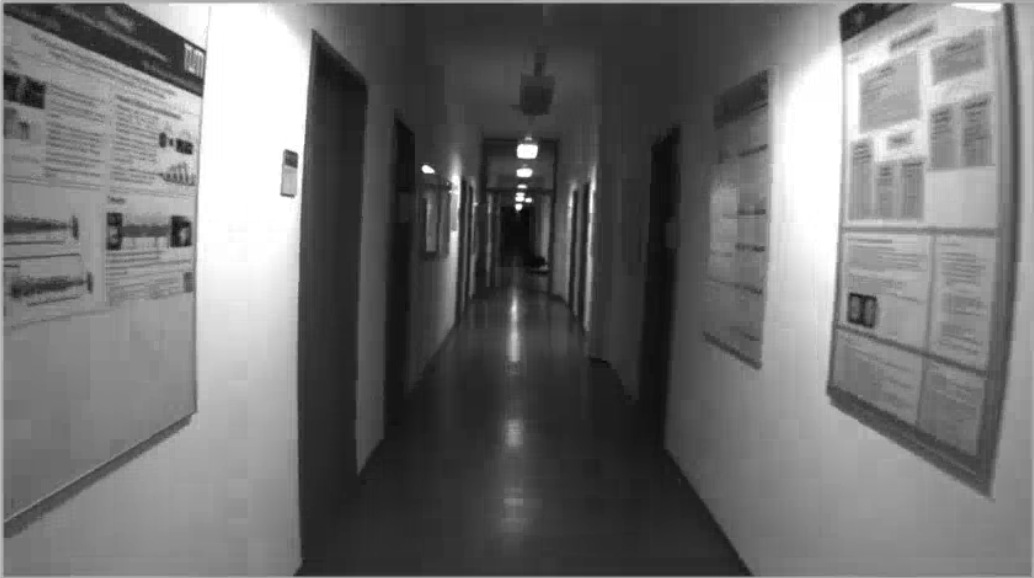}}
        & \gframe{\includegraphics[width=\linewidth]{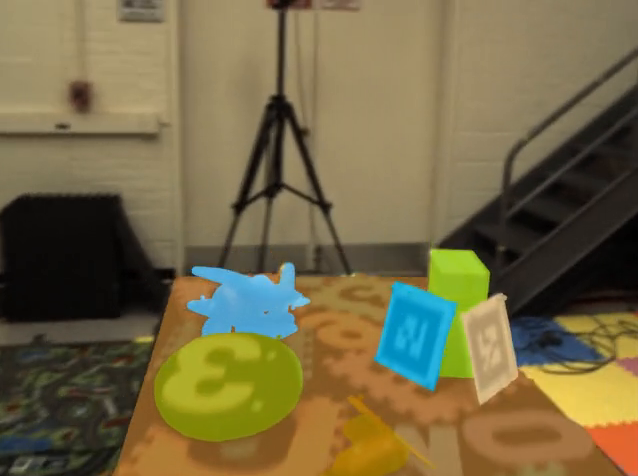}}
        & \gframe{\includegraphics[trim={0 .5cm .1cm .5cm},clip,width=\linewidth]{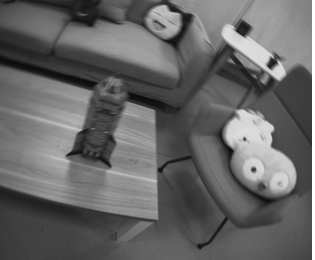}}
        & \gframe{\includegraphics[width=\linewidth]{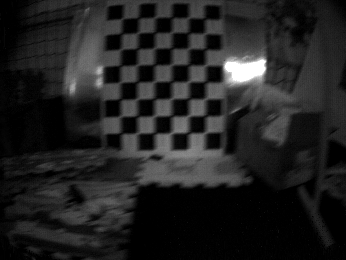}}
        & \gframe{\includegraphics[trim={0 .6cm 0 .6cm},clip,width=\linewidth]{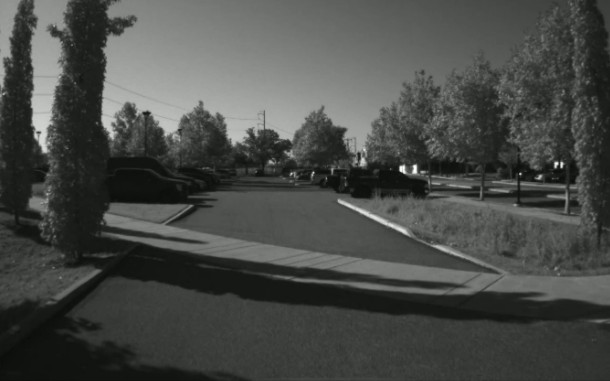}}
        & \gframe{\includegraphics[width=\linewidth]{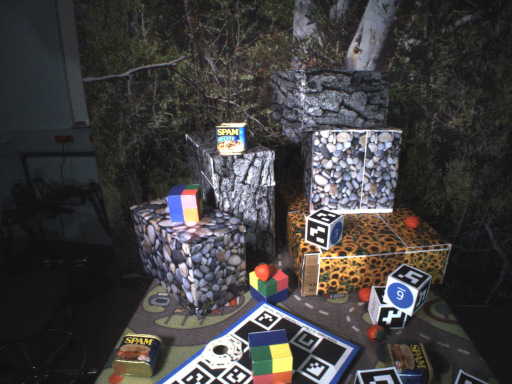}} \\

        \rotatebox{90}{\makecell{Events (left)}}
        & \gframe{\includegraphics[width=\linewidth]{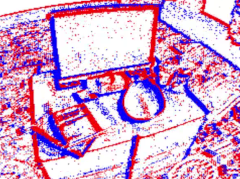}}
        & \gframe{\includegraphics[width=\linewidth]{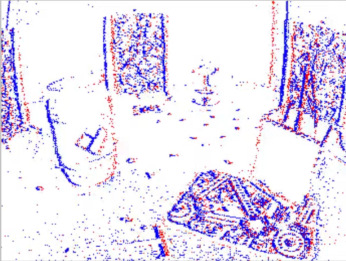}}
        & \gframe{\includegraphics[width=\linewidth]{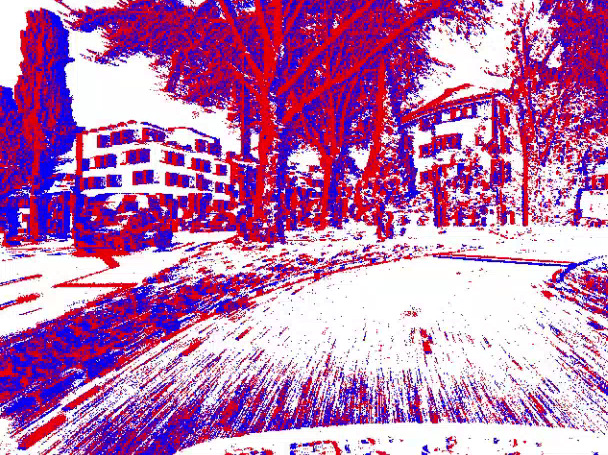}}
        & \gframe{\includegraphics[width=\linewidth]{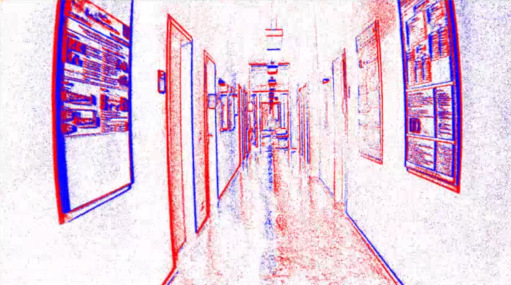}}
        & \gframe{\includegraphics[width=\linewidth]{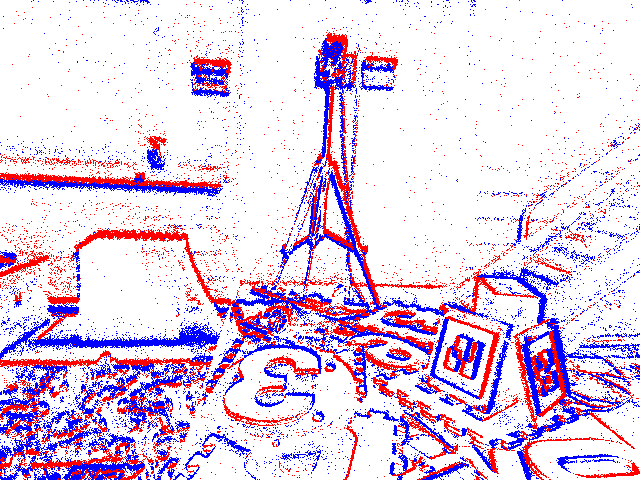}}
        & \gframe{\includegraphics[width=\linewidth]{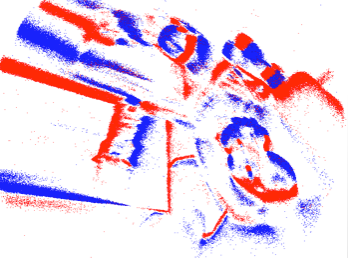}}
        & \gframe{\includegraphics[width=\linewidth]{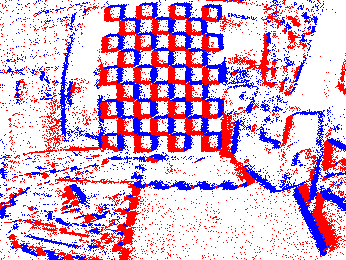}}
        & \gframe{\includegraphics[width=\linewidth]{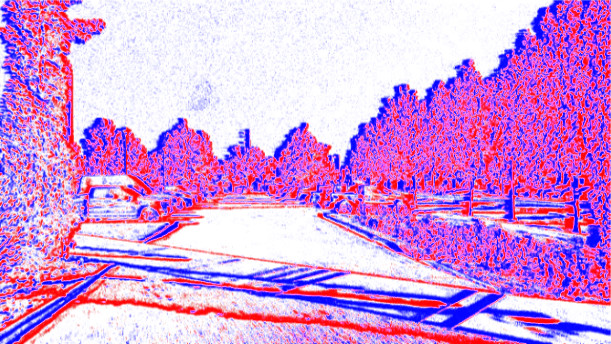}}
        & \gframe{\includegraphics[width=\linewidth]{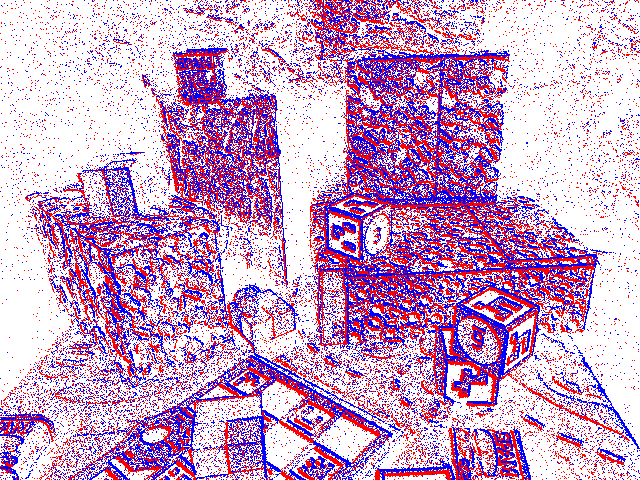}} \\

        \rotatebox{90}{\makecell{GT depth}}
        & N/A
        & \gframe{\includegraphics[width=\linewidth]{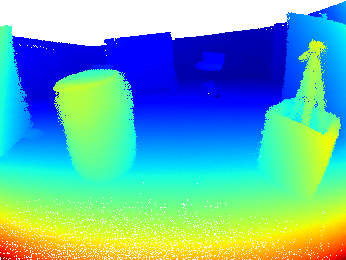}}
        & \gframe{\includegraphics[width=\linewidth]{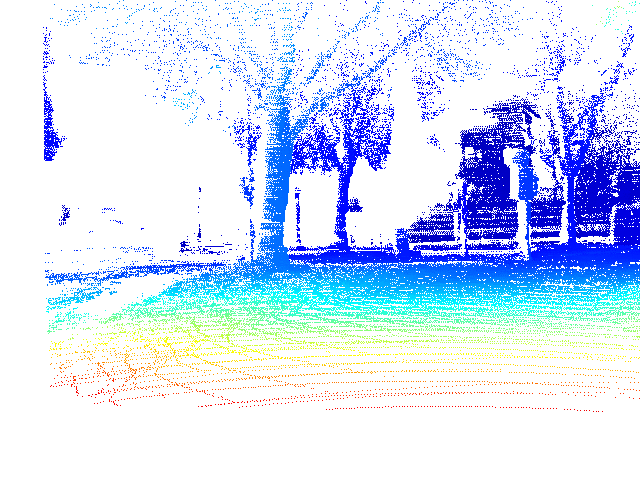}}
        & N/A
        & \gframe{\includegraphics[width=\linewidth]{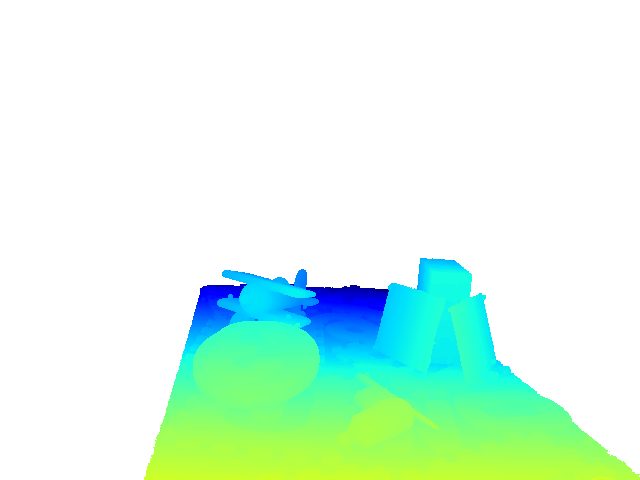}}
        & \gframe{\includegraphics[trim={0 0.8cm 0 1.6cm},clip,width=\linewidth]{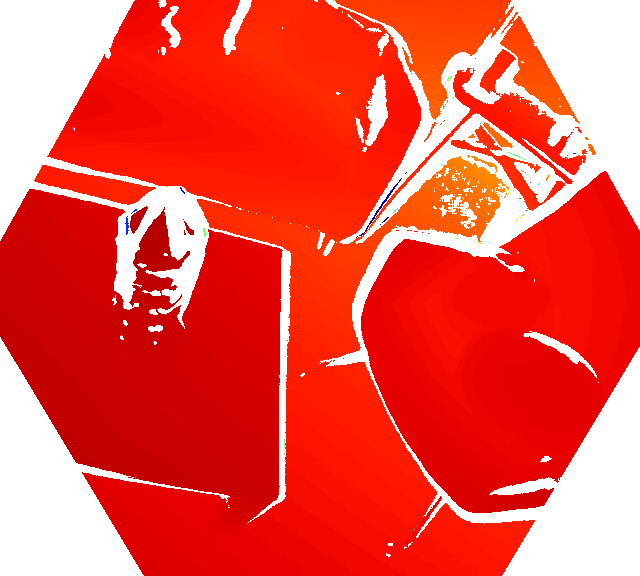}}
        & N/A
        & \gframe{\includegraphics[width=\linewidth]{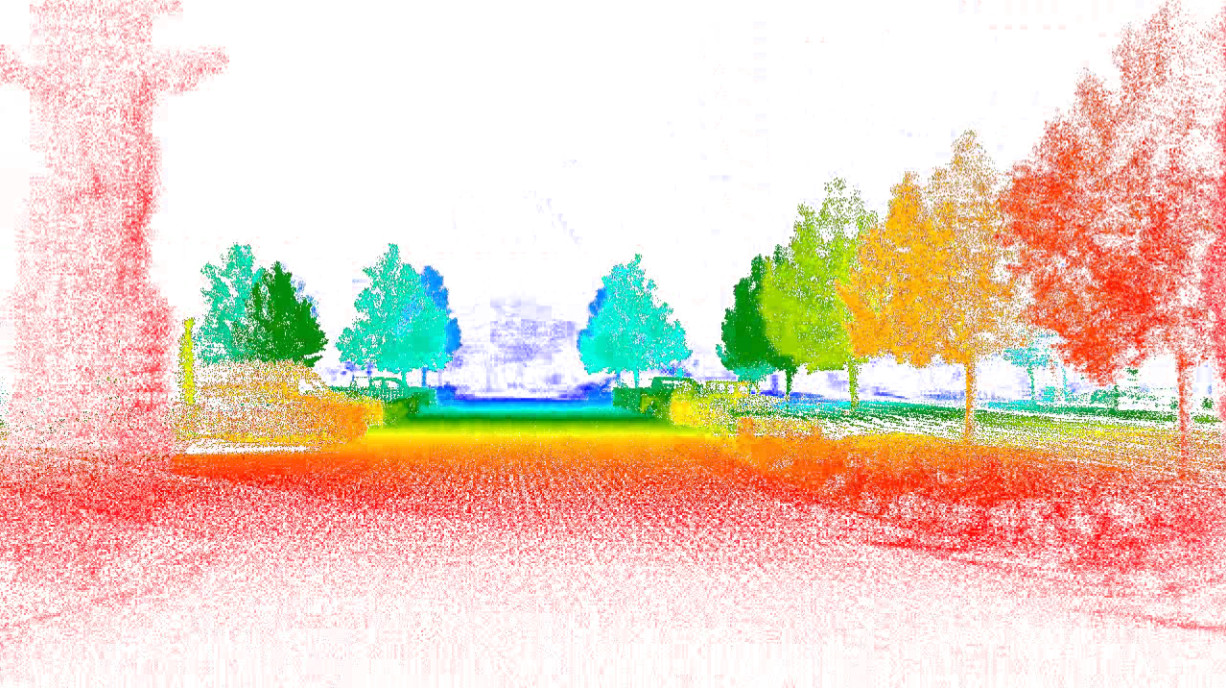}}
        & \gframe{\includegraphics[trim={.75cm 0 .75cm 0},clip,width=\linewidth]{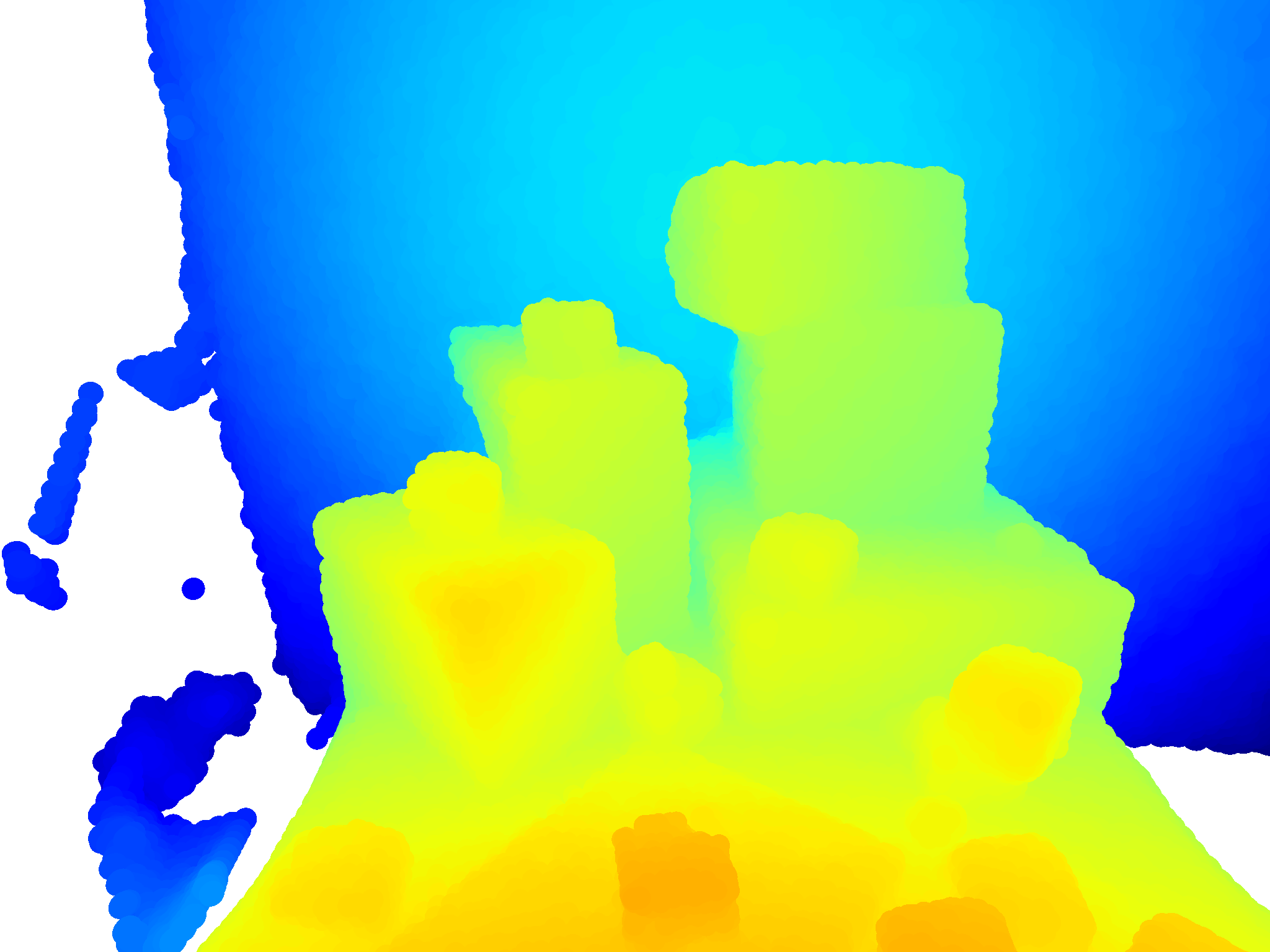}} \\
	\end{tabular}
	}
\end{adjustbox}
    \caption{\descript{Representative datasets used for event-based stereo depth estimation.}
    \label{fig:datasets}}
\end{figure*}
 \fi
\ifarxiv \ifclearlook\cleardoublepage\fi \subsection{Description of Existing Datasets}
\label{sec:datasets:description}

\hypertarget{RPGStereo}{\subsubsection{UZH-RPG Stereo Dataset}}

\descript{
This event stereo dataset %
\firstcite{Zhou18eccv} (\cref{fig:datasets}) from %
the University of Zurich comprises eight real-world sequences and a synthetic one.
The synthetic sequence was generated using an event camera simulator \firstcite{Mueggler17ijrr} and depicts three textured fronto-parallel planes at different depths while the camera translates in a circle. 
The real-world sequences were recorded with a handheld stereo rig in a motion capture room where accurate ground truth (GT) poses are available. 
Events were recorded using DAVIS240C cameras (a small 240$\times$180 px resolution) in well-lit conditions.} 
\proscons{
It serves as a good starting point for evaluating algorithms. 
However, it does not contain GT depth, so only qualitative evaluation of 3D reconstruction algorithms is possible.
}

\subsubsection{The Multi Vehicle Stereo Event Camera Dataset}

\descript{
The \textbf{MVSEC} dataset %
\firstcite{Zhu18ral} (\cref{fig:datasets}) from the %
University of Pennsylvania was the first comprehensive stereo event dataset including frames, GT poses and GT depth maps in various indoor and outdoor scenarios. 
Two prototype mDAVIS346-B event cameras were used with hardware synchronization. 
\findin{It is one of the most popular event camera datasets for SLAM, and is widely used for benchmarking event-based algorithms.} 
It comprises data acquired with the sensors mounted on a flying hexacopter, a car, and a motorcycle, with a variety of speeds, illumination levels and environments. 
For outdoor sequences, GT poses were obtained by visual-inertial odometry (VIO) using LiDAR, IMU and GPS (where available).
The GT for the different sensors is provided already pre-aligned in space and time, and therefore does not require any post-processing before use. 
For example, depth maps and camera poses are provided individually for each event camera.
}

\proscons{
\textbf{Limitations}. The authors observed an imbalance between event polarities when using default biases (positive events are 2.5 to 5 times more prevalent than negative ones).
Moreover, the motion capture and GPS are not hardware-synchronized with the rest of camera setup. 
They rely on synchronization via CPU clock by recording on the same computer, which may cause the clocks to drift for long sequences. 
Also, the calibration / alignment of some sensors was manually refined, which may suffer from some problems. 
\findin{The outdoor driving datasets are not well-suited for stereo applications because the baseline of 10cm is too small for the depth range in outdoor settings.} 
The relatively small camera resolution of 346$\times$260 px is also a disadvantage for large-scale scenarios~\firstcite{Gehrig21ral}.
Moreover, the mapping method used to generate GT depth maps assumes the scene is static, and thus provides erroneous depth (up to 2m error) on independently moving objects (\textbf{IMOs}) like cars. 
\findin{Thus, many stereo depth algorithms in the literature are evaluated only on the indoor flying sequences.}  
However, even in some of the flying sequences, the VI sensor data is missing. 
Moreover, it is inherently affected by shutter noise due to the DAVIS’ hybrid event-frame mode~\firstcite{Gao22ral}.
}

\descript{
Ground truth depth is provided by a LiDAR sensor operating at a limited rate of 20 Hz.} 
\proscons{The LiDAR points do not cover the full image plane (due to the limited field of view (FOV) of the LiDAR or due to the lower spatial sampling close to the sensor rig). 
Consequently, many pixels lack GT depth values, as seen in \cref{fig:datasets}. 
This effect of having pixels without GT depth becomes more noticeable in subsequent datasets with higher resolution event cameras but still the same LiDAR resolution, as seen in \cref{fig:datasets} (for DSEC or M3ED). 
The LiDAR also has a limited operation in range (depth), hence in outdoor scenarios, pixels may not have GT depth if objects are far away, as these values are beyond the reliable acquisition range of the LiDAR.
}

\subsubsection{Stereo Event Camera Dataset for Driving Scenarios}

\descript{
To address the gap in literature for autonomous driving datasets, the \textbf{DSEC} dataset %
\firstcite{Gehrig21ral} (\cref{fig:datasets}) was created by the UZH-RPG lab. 
It uses higher resolution Prophesee Gen 3.1 event cameras (640$\times$480 px) and a bigger baseline (60cm) than MVSEC to record data more suitable for stereo algorithms in driving scenarios. 
It aims to provide a standard stereo driving dataset similar to the well-known frame-based KITTI dataset. 
The dataset was recorded while driving through Switzerland in various illumination conditions. 
It also includes driving at night and through tunnels, both of which are challenging HDR perception scenarios. 
It provides GT depth measurements using LiDAR, like MVSEC. 
IMU data is also available. 
Hardware synchronization was achieved between the GPS, event- and RGB cameras using a microcontroller. 
A pulse wave was used to simultaneously inject special ``trigger events'' into the cameras and trigger RGB frame acquisition.}

\bg{
Competitions on optical flow estimation and disparity estimation on DSEC data (using events alone or with grayscale frames) have been held at CVPR Workshops.} 
\descript{The dataset contains some sequences for which GT flow disparity (and therefore, flow) has not been published, \challeng{to avoid overfitting in the above-mentioned competitions}.
}

\proscons{
Unlike MVSEC, the GT depth generation approach in DSEC handles occlusions and IMOs, but leads to increased sparsity of disparity labels.} %
\descript{The filtering approach is based on Semi-Global Matching (\textbf{SGM}) \cite{Hirschmuller08pami} using RGB images, \findin{thus resulting in sparser disparities in challenging conditions such as night driving or image overexposure.}
}

\proscons{
\textbf{Limitations}. A disadvantage of DSEC is that it does not provide GT poses. 
For short intervals, camera pose may be estimated with odometry using LiDAR and IMU. 
Although GPS was recorded, it is not publicly available. 
Since the motion in DSEC is predominantly look-ahead driving, often very few events are generated at the center of the image plane while driving on straight roads. 
The small parallax motion (except during turning) also makes it difficult to estimate depth with monocular methods like EMVS~\firstcite{Rebecq18ijcv}. 
LiDAR points cover less percentage of the image plane than in MVSEC due to the higher resolution event cameras used. 
Hence, many pixels lack GT depth. 
}

\proscons{During the night driving sequences, the flashing street lights produce a large number of spurious events that may be detrimental to algorithms that assume events are mostly caused by moving edges (brightness constancy assumption).}

\hypertarget{shef}{\subsubsection{Stereo Hybrid Event-Frame Dataset}}

\descript{
While datasets like MVSEC and DSEC primarily aim to solve the stereo correspondence problem between event cameras, the SHEF dataset %
\firstcite{Wang21iros} (\cref{fig:datasets}) by the Australian National University 
targets the problem of stereo 3D reconstruction between a frame-based and an event-based camera. 
\bg{Accurate stereo correspondence between the event and frame-based camera would enable better alignment between them, thus producing a similar effect as a DAVIS camera~\firstcite{Taverni18tcsii} but with higher image quality and resolution.}
}

\descript{
They use a Prophesee Gen3 VGA event camera and a FLIR RGB camera separated by a baseline of 6.54 cm, and mounted on the end effector of a UR5 robot arm manipulator.
GT poses are provided through the robot’s kinematics (computed from the motion of its joints). 
GT point clouds are generated for each environment by using the RGB camera in combination with a commercial multi-view stereo algorithm 3D Flow (Zephyr). 
GT depth projection for each camera frame is not provided but may be computed from the point cloud using the software provided.
This is, to the best of our knowledge, the only stereo event dataset acquired using a robot arm, 
which allows to \challeng{produce controlled repeatable motions in various surroundings}. 
Data is recorded with pure 3D translations drawing squares, circles and Lissajous curves.
}

\proscons{
\textbf{Limitations}. 
This dataset uses only a single event camera and a single frame-based camera, hence it is not suitable for pure event-driven stereo depth estimation. 
Paring an event camera with a frame-based camera can make the whole system suffer from the bottlenecks of the frame-based camera (low dynamic range, motion blur, etc).
}

\subsubsection{The TUM Stereo Visual-Inertial Event Dataset}

\descript{The TUM-VIE dataset %
\firstcite{Klenk21iros} (\cref{fig:datasets}) from %
TU Munich is the first stereo dataset using 1Mpx event cameras. 
It contains data from two Prophesee Gen4 HD event cameras, two frame-based grayscale cameras and an IMU. 
GT poses are provided via a motion capture system at the beginning and end of every sequence (whenever the sensor rig is in the motion capture room).
\proscons{However, no GT depth is provided.}  
With many long sequences that contain loops, it is primarily targeted for robust VIO applications. 
It is recorded with egocentric perception in mind with handheld and head-mounted sensor setups while the user walks, runs, skates and bikes through changing environments. 
}

\descript{
Compared to other stereo event datasets, they use wider FOV lenses (90 degrees horizontal), and include photometric calibration for all sequences. 
The event and frame-based cameras are hardware-synchronized with the IMU. 
\proscons{However, due to IMU readout delay the timestamps between events and IMU, the event camera clock and IMU clock need to be realigned in post-processing.}  
The time offset between the IMU and motion capture clock is computed by aligning the angular velocities. 
The published dataset has already been aligned in time. 
\findin{They also use IR-blocking filters to prevent interference from the motion capture system}.
}

\proscons{
\textbf{Limitations}. For sanity testing, it also contains some sequences recorded fully inside a motion capture room with simple motions. 
However, as noticed in~\firstcite{Ghosh22aisy}, the baseline of 11.84 cm is too big for the small depth range (nearest object $\approx$40cm away) in these sequences. 
The \emph{calib\_A} sequences were better calibrated than the \emph{calib\_B} ones~\firstcite{Ghosh22aisy}. 
A lot of events due to flashing artificial lights and surface reflections are also present in this dataset.
}

\subsubsection{Event Camera Motion Segmentation Dataset}

\descript{
The EV-IMO2 dataset %
\firstcite{Burner22evimo2} (\cref{fig:datasets}) from %
the University of Maryland 
is the only trinocular event camera dataset targeting structure from motion, VO, object recognition and
segmentation of IMOs. 
It is an evolved version of previous datasets recorded by the same lab \firstcite{Barranco16fns}, \firstcite{Mitrokhin19iros}. 
Its predecessor, EV-IMO~\firstcite{Mitrokhin19iros}, is a monocular dataset for independent motion segmentation. 
}
\descript{
EV-IMO2 has two versions: v2 has improves upon v1 in terms of post-processing the raw data. 
It features temporal synchronization between sensors, less jitter, and a more efficient data storage format (npz).
}

\descript{
The dataset contains VGA resolution events from two Prophesee Gen 3 and one Samsung DVS Gen 3 camera~\firstcite{Son17isscc} (like a DVXplorer), as well as RGB images from a single frame-based camera. 
The handheld camera rig is moved in a motion capture room while observing a textured tabletop with various objects. 
GT pose is provided via motion capture, whereas GT depth is provided by projecting the point clouds from a 3D scanner to various camera poses. 
The ``SfM subsets'' are relevant for SLAM since they comply with the static scene assumption; 
they comprise events recorded in both normal and low light scenarios.
The dataset also contains some sequences with rudimentary motion for sanity testing algorithms.
}

\proscons{
\textbf{Limitations}. 
Although EV-IMO2 contains trinocular event data, the FOVs have a smaller overlap than expected. 
This is primarily because the stereo Prophesee cameras are mounted in a portrait orientation. 
Moreover, since the depth maps are generated by 3D scanning individual objects, background pixels beyond the table-top have no GT depth information, as shown in \cref{fig:datasets}. 
Also, no information about hardware synchronization is provided.
}

\subsubsection{Versatile Event-Centric (VECtor) Benchmark Dataset}

\descript{
The VECtor benchmark dataset %
\firstcite{Gao22ral} (\cref{fig:datasets}) from %
ShanghaiTech University 
is the first event SLAM benchmark with a fully-synchronized hardware setup and full GT poses and depth
for many small-scale and large-scale indoor sequences under various illumination. 
}

\descript{
Stereo events are recorded using two Prophesee Gen3 VGA cameras.  
The authors opted for a VGA resolution rather than HD cameras because \challeng{they observed a ``smearing effect'' on top of the surface of active events, similar to problems} reported by~\firstcite{Hu21cvprw,Alzugaray18ral}, 
where motion blur in the event stream or timestamp delays was observed from sudden and significant contrast changes on DAVIS event cameras. 
The authors also do not use the highest resolution cameras \challeng{to reduce the number of artifacts and the data rate}. 
The dataset also comprises stereo RGB frames and IMU data for all sequences. 
For small-scale sequences, GT poses are provided via motion capture whereas GT depth is provided by a Kinect depth sensor. 
For large scale egocentric sequences, GT depth is provided by a LiDAR whereas camera poses are provided by aligning LiDAR point clouds with a pre-scanned 3D map of the environment. 
}

\descript{
The VECtor benchmark dataset contains diverse motions using head-mounted or handheld setups. 
It is fully hardware-synchronized using a microcontroller unit (MCU), whose setup and programming has been open-sourced. 
The dataset also contains sequences with simple motions and surroundings for sanity testing.
}

\proscons{\textbf{Limitations}. 
To control the event rate so that it does not exceed available bandwidth, a high contrast sensitivity threshold is used, which produces sparse event recordings. 
The bias values used in different sequences are reported. 
In this dataset, the lenses used for the event cameras are small for the sensor format, causing blind regions at the camera corners. 
In many sequences (e.g., robot-fast, units-dolly, school-dolly), a circular boundary of events around the event
camera’s FOV can be observed. 
Spurious events are generated at the edge of the blind regions probably due to lens artifacts. 
These spurious events are difficult to model and thus must be filtered out as a pre-processing step, further reducing the overall FOV.
Erroneous GT depth measurements from the Kinect camera are also noticeable in some sequences (\emph{sofa\_normal}, \emph{sofa\_fast} etc.). 
An example is shown in the GT depth of \cref{fig:datasets}, where a low depth value is displayed almost everywhere except at some object edges.}

\subsubsection{Univ.~of Hong Kong Visual-Inertial Odometry dataset}

\descript{
The stereo VIO dataset %
\firstcite{Chen23ral} (\cref{fig:datasets}) from the University of Hong Kong (HKU VIO) 
comprises events recorded using DAVIS346 cameras mounted on a drone. 
The sequences exhibit high-speed aggressive motions in indoor scenarios under normal and HDR conditions, as well as a large-scale outdoor scene.
In the indoor setting, GT poses are provided partially at the start and end of sequences via a motion capture system. 
This is a challenging VIO dataset (EVO~\firstcite{Rebecq17ral}, ESVO~\firstcite{Zhou20tro}, Ultimate SLAM~\firstcite{Rosinol18ral} failed in most sequences).
}

\descript{
\textbf{Limitations}. 
Having a stereo setup is sensible to estimate absolute scale more reliably than with an IMU. 
However, the stereo baseline of 6cm and camera resolution of 346$\times$260 px are rather small for stereo applications. 
The dataset does not have GT depth or time synchronization (to the best of our knowledge).
The stereo pair consists of a DAVIS monochrome and a DAVIS color camera; the difference in appearance between them might make it difficult for stereo matching than if both sensors were of the same type.
}

\subsubsection{Multi-Robot, Multi-Sensor, Multi-Environment Event Dataset (M3ED)}

\descript{
As the evolution of MVSEC from the GRASP Lab at UPenn, M3ED \firstcite{Chaney23cvprw} %
comprises data recorded using three distinct platforms -- a car, a drone and, for the first-time, a legged robot (Spot). 
The event-camera setup consists of Prophesee EVK4 HD cameras with infrared (IR) filters. 
It improves upon the other 1Mpx stereo event dataset TUM-VIE by providing ground truth depth from a LiDAR. 
Additionally, camera pose computed using a LiDAR-inertial odometry system (FasterLIO), is provided for all sequences. 
It has indoor and outdoor sequences in both urban settings and forests, including semantic segmentation masks for outdoor day sequences in urban scenes.
The forest sequences are particularly challenging for event-based perception due to exceptionally high event rates (upto 200 Mev/s) produced by high amounts of texture from the vegetation and light-and-shadow effects. 
M3ED was recorded using an Intel NUC with ROS nodelets, with full hardware-synchronization across sensors.
Handling high data rates while maintaining the integrity of the data stream is challenging. 
Through analysis of the synchronization signals through the camera, the authors observed an error of approximately 1 $\mu$s/s. 
}

\proscons{
\textbf{Limitations}.
The stereo baseline of 12cm may be insufficient for the depth range required in driving scenes. 
Moreover, there is soft mounting between the LiDAR and cameras to dampen vibrations, so extrinsic parameters are not always consistent. 
Even though the dataset encompasses several challenging scenarios for stereo perception like driving at night in the city and forests, it lacks the inclusion of simple motions for debugging and sanity testing. 
For 1Mpx event cameras, bias tuning and hardware-based noise filtering is of tantamount importance to control the high data throughput so that the sensors do not saturate. 
By recording all sequences with the same bias settings, the authors opted for inter-sequence comparability at the expense of data efficiency and SNR optimization.
}

\subsubsection{ECMD: An Event-Centric Multisensory Driving Dataset for SLAM}

\descript{
ECMD~\firstcite{chen23tiv} is another driving dataset from %
Hong Kong University that improves upon DSEC by using a thermal camera in addition to stereo event and frame-based cameras. 
The dataset contains sequences with two pairs of stereo event cameras (DAVIS346 and DVXplorer). 
Moreover, it contains multiple LiDARs mounted on the car to maximize coverage. 
It provides high accuracy GT poses with GNSS-RTK/INS module that combines GPS measurements with gyroscope at 1 Hz. 
It focuses on day- and night-time driving (aided by thermal camera) in urban settings.
}

\gpass{
\textbf{Limitations}.
Ground truth is available only at a slow rate of 1 Hz. 
It is however possible to compute pose measurements in between using LiDAR-inertial odometry.
}

\subsubsection{Stereo Visual Localization Dataset (SVLD)}

\descript{
Hadviger et al.~\firstcite{hadviger23ecmr} proposed this stereo event dataset for VO/SLAM. 
The primary focus is on high-quality GT camera poses in both indoor and outdoor scenarios. 
Indoor scenes use a motion capture system, whereas outdoor driving scenarios use a GNSS-RTK/INS system similar to ECMD, which combines GPS and IMU data to provide accurate poses at 40 Hz.
GT depth is provided outdoors by a LiDAR.}

\proscons{
\textbf{Limitations}.
The indoor scenes do not have any GT depth. 
There is no hardware clock synchronization between the event cameras and other sensors. 
It relies on software synchronization using CPU clock while recording on a common computer and then uses a tool called Calirad to estimate temporal shifts in post-processing. 
Thus, even though the aim is to provide high quality GT poses, they might suffer from temporal misalignment. 
}

\subsubsection{Novel Sensors for Autonomous Vehicle Perception Dataset (NSAVP)}
\descript{
NSAVP~\firstcite{Carmichael24ijrr} is a stereo event dataset that focuses on autonomous driving. 
It also uses VGA resolution (640$\times$480 px) event cameras similar to DSEC, ECMD and SVLD. 
It is the only dataset with stereo thermal cameras, with the aim of benchmarking vision-based depth estimation in low-light. 
In addition to the GNSS-RTK module, it uses wheel encoders to produce high quality GT poses at 200 Hz. 
It also has the biggest stereo baseline of 1m among all the driving datasets, significantly improving depth perception range. 
It includes precise temporal synchronization among all sensors using hardware triggers and Precision Time Control (PTP), and it is described in detail.
The sensor-mounted car was driven repeatedly along the same two 8km routes under different lighting conditions to enable robust algorithm development that works in a wide range of scenarios.
It was used for benchmarking the tasks of visual place recognition, which is critical for loop closure in SLAM.}

\proscons{
\textbf{Limitations}.
Even though this dataset aims to provide a benchmark for autonomous driving with non-conventional sensors, it does not contain any ground truth depth.
}

\subsubsection{FusionPortablev2 Dataset}
\descript{
FusionPortablev2~\firstcite{wei24ijrr} is a multi-sensor dataset for generalized SLAM across diverse environments and platforms. 
Similar to M3ED \firstcite{Chaney23cvprw}, it contains recordings from multiple platforms -- handheld, legged robot (Unitree Go1), Unmanned Ground Vehicle (UGV) and a car. 
However, they make useful platform-specific modifications for each case. 
For large scale scenes recorded with the car, the event stereo baseline is set to 73 cm, whereas it is kept to 25 cm in other cases. 
Moreover, they also provide kinematic data (joint positions/angles and wheel encoders) for the robots to aid sensorimotor learning.
Expanded from FusionPortable~\firstcite{jiao22iros}, it uses stereo DAVIS346 event cameras, stereo frame-based cameras, IMU and LiDAR as the primary sensors. 
Besides LiDAR, laser scanners provide high quality GT maps for the hanheld, UGV and legged sequences. 
GT poses are provided through GPS and GNSS-RTK/INS in the car sequences and some of the UGV sequences. 
The dataset covers a wide range of environments like urban roads, mountain roads, tunnels, rooms, (textureless) grasslands, garages and parking lots.
}

\proscons{
\textbf{Limitations}.
The event cameras are not hardware synchronized with the other sensors; they rely on ROS timestamps dictated by the CPU clock of the recording computer, leading to an Average Relative Time Latency (ARTL) of up to 15 ms with respect to LiDAR due to transmission delays.
For the handheld and legged robot, only 3-DOF GT poses are provided (using a laser tracker).
Moreover, some sequences have data gaps due to hardware failures.
Even though the dataset is aimed for robot control learning along with SLAM, its size is small compared to other such datasets recorded in constrained setups~\firstcite{brohan23arxiv}.
}

\subsubsection{CoSEC: A Coaxial Stereo Event Camera Dataset for Autonomous Driving}
\descript{
CoSEC~\firstcite{peng24arxiv} uses a pair of beam splitters to perfectly align pixels between frame-based and event cameras (1-Mpx EVK4s) for multi-modal fusion. 
This is needed because there are currently no combined event-intensity sensors for high resolution (VGA and HD) in the market; 
the widely used DAVIS cameras have a resolution of 346$\times$260~px. 
Previous multi-modal datasets tried to get around this problem by placing the frame-based and event cameras as physically close to each other as possible while recording far away objects to maximize pixel alignment. 
The dataset contains 128 sequences ($\approx$1 hour) recorded with cameras mounted on a car driving in city, parks and villages during both day and night.
GT depth is provided by combination of LiDAR and monocular depth-prediction DNN. 
While it does not contain GT poses, they can be derived using LiDAR-inertial odometry. 
In fact, the poses obtained in this way are combined with GT depth to provide GT optical flow.}

\proscons{
\textbf{Limitations}.
Similar to DSEC \firstcite{Gehrig21ral}, this dataset does not provide GT poses directly. 
Poses derived from LiDAR-IMU SLAM may be prone to errors in dynamic scenes. 
Beam splitters divide light intensity incident on the sensors by half (thus producing more noise) and have a reduced FOV compared to non-beam splitter setups. 
}

\subsubsection{Additional Stereo Event Datasets}
\descript{
Most of the aforementioned datasets address the task of VO/SLAM, but there are other stereo event datasets with the goal of instantaneous depth estimation for other downstream applications. 
For example, the DVS stereo dataset from Andreopoulos et al.~\firstcite{Andreopoulos18cvpr} comprises a setup of stereo DAVIS240C cameras that is stationary, and therefore only perceives dynamic IMOs in the scene like a rotating fan and a toy butterfly. 
Their goal is dynamic depth estimation of fast moving objects where finding stereo correspondences is difficult. 
A Kinect is used for dense GT depth.
}

\descript{
StEIC\firstcite{Lin25tcsvt} and SEID\firstcite{Ding24tmm} are hybrid event-intensity stereo datasets aimed to tackle the task of motion deblurring and video frame interpolation, respectively (i.e., their goal is not pure hetero event-stereo matching to estimate instantaneous depth).
They both use a handheld setup with a VGA resolution event camera (SilkyEvCam), a frame-based camera and Realsense. 
CPU clock and manual alignment is used for temporal synchronization. 
StEIC contains as input artificially blurred frames by combining multiple sequential frames, whereas SEID drops intensity frames to lower frame rate artificially and then enables depth map estimation in between intensity frames, thereby also addressing the task of depth map interpolation.
}
 \fi
\ifclearlook\cleardoublepage\fi \ifarxiv
\subsection{Dataset Discussion}
\fi
\label{sec:datasets:discussion}

\findin{While many stereo event datasets are currently available, they are still few compared to their frame-based counterparts, 
partially due to the novelty of the sensors (lack of established ecosystem, etc.) and their high cost.
Among them, the most popular ones are MVSEC, UZH-RPG stereo dataset and DSEC (with $\approx$1000 citations, they account for $\approx$80\% of all event-based stereo dataset citations (as of Nov.~2024)).
}

\findin{We observed that the datasets released so far comprise a rich diversity in terms of camera models and resolutions (QVGA to HD), recording platforms (cars, drones, legged robots, manipulator arm, ego-centric etc.), mode of GT acquisition (LiDAR, 3D scanners, depth cameras, motion capture, GPS, IMU, laser trackers etc.) and recording scenarios (urban, forests, garages, rooms, corridors, night, etc.).
}

\proscons{The UZH-RPG dataset is good for preliminary qualitative evaluation, due to its moderate motion, low resolution (fast prototyping), accurate poses and circular motions. 
MVSEC provides comprehensive GT depth and poses. 
The indoor drone sequences being widely used, but the outdoor driving sequences are not suitable for stereo due to the small baseline \cite{Zhu18eccv}.
The HKU-VIO dataset aims to push the limit of on-device stereo VO during high-speed drone flight in difficult lighting, but contains no GT depth and only partial pose data.
TUM-VIE was the first dataset using HD event cameras (with wide angle lenses), but provides no GT depth. 
The VECtor benchmark dataset provides comprehensive GT with diverse motions and illumination in indoor scenarios, and is also fully hardware synchronized. However, because of the high contrast sensitivity thresholds used, the event data is often sparse. 
It also contains artifacts at the edges, and contains incorrect GT depth from Kinect in some sequences.
EV-IMO2 is the only dataset with trinocular event camera sequences but the cameras are oriented in such a way that the overlap in the FOV among them is small.
The SHEF dataset uses a robot arm to generate controlled repeatable motions, which are useful for tuning camera biases. 
However, it employs only one event camera since the goal is event-intensity hybrid stereo. 
}

\findin{Recently, event-based SLAM datasets have heavily focused on \textbf{autonomous driving}. 
DSEC is currently the most popular event-based driving dataset, with VGA resolution event cameras, but it does not have GT poses. 
M3ED overcame these shortcomings by providing comprehensive GT depth and poses with HD stereo event data recorded in diverse scenarios (forests, cities, indoors) using multiple robot platforms including cars.
It contains challenging scenes with high texture, flashing lights etc. achieving data rates of up to 200 million events/s, fostering the development of real-time algorithms that can handle high throughput. 
ECMD, SVLD and NSAVP are newer driving datasets with a focus on accurate GT trajectories using state of the art GNSS-RTK/INS sensors, GPS and IMU. 
For better benchmarking during the night, they sometimes use thermal cameras.
Similar to M3ED, FusionPortablev2 records data across multiple robot platforms in diverse backgrounds, along with robot joint and wheel encoders to aid learning of sensorimotor control.
CoSEC is the latest stereo event dataset, containing optically aligned HD event and frame-based cameras using a pair of beam-splitters, also with a focus on driving.
}

\findin{\textbf{Calibration}. 
The UZH-RPG stereo, MVSEC, HKU-VIO, ECMD and FusionPortablev2 datasets use DAVIS cameras, so they use the readily available grayscale frames for calibration since grayscale intensities and events share the same pixel array. 
Other datasets that do not have shared pixel arrays generate frames containing checkerboards or AprilTag grids from events for calibration. 
While DSEC, CoSEC, M3ED, NSAVP and SVLD reconstruct intensity frames from events, the VECtor benchmark, SHEF and TUM-VIE datasets use a blinking screen (with a calibration pattern) and accumulate events for calibration.
The limits of event camera calibration can still be pushed to further improve accuracy of depth estimation methods.
}

\ifarxiv \else  \fi
\ifclearlook\cleardoublepage\fi \ifarxiv
\subsection{Checklist of Good Practices}
\label{sec:datasets:goodpractices}
\begin{figure}[t]
    \centering
    \includegraphics[trim={9.6cm 2.3cm 8.2cm 0.77cm},clip,width=.9\linewidth]{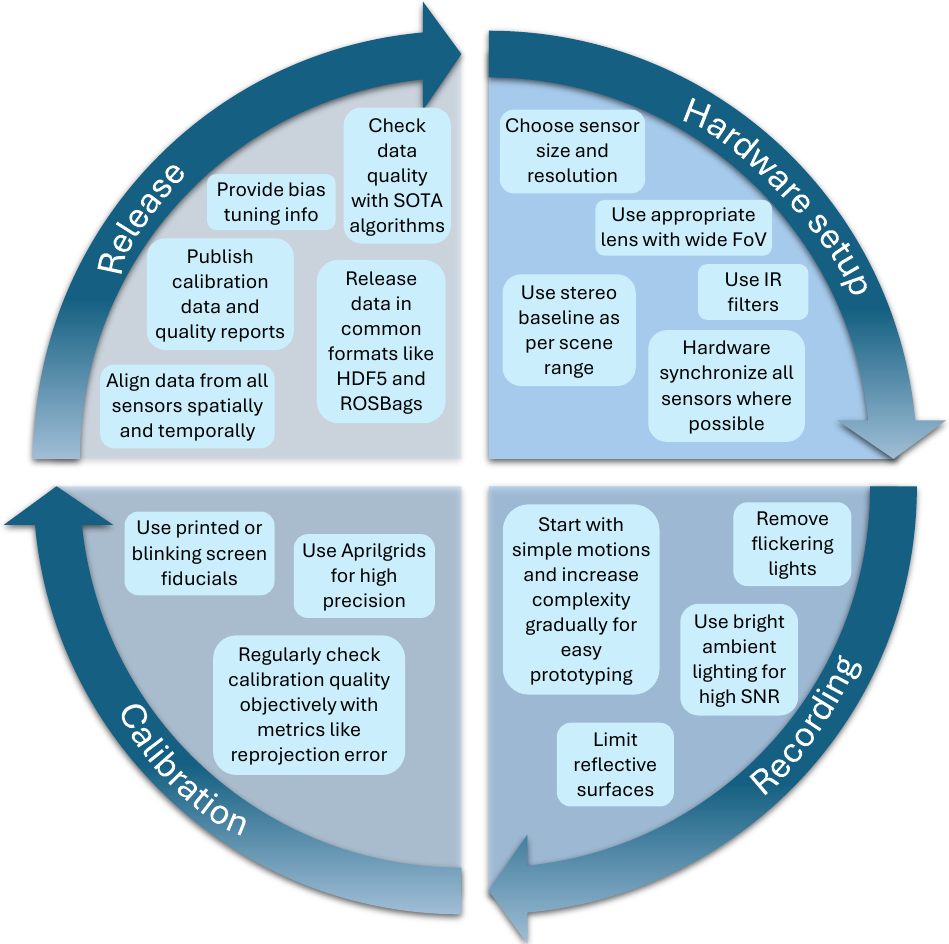}
    \caption{\label{fig:cycle}Event stereo data recording cycle.}
    \vspace{-2ex}
\end{figure}

\rammif{
Surveying existing datasets and given our own experience, we suggest iterating through the steps outlined in \cref{fig:cycle} to produce high-quality data for event stereo research.}  
\rammif{A concrete checklist with recommendations is given below.}
\begin{itemize}[wide, labelwidth=!, labelindent=0pt]
\item Consider what the appropriate \emph{spatial resolution} is for the target task:  
low resolution event cameras are good for quick prototyping, and HD cameras are preferred for application with high accuracy demands. 
Most comprehensive and ambitious is providing data with different spatial resolution event cameras, 
possibly aligned with frame-based cameras.
\item Choose a \emph{lens} suitable for the camera's sensor size, so that the full sensor size is used and no artifacts are due to artificial boundaries between the lens and the sensor.
\item When recording with high-resolution stereo event cameras like Prophesee's EVK4, use a powerful computer to ensure \emph{no data drops or timestamping issues} during high data rates. Even though it is possible to limit data rates in the firmware of these cameras, it leads to undesired artifacts.
\item Use wide angle lenses (but not fish-eye) for the task of (stereo) VO/SLAM, to be able to disambiguate translational and rotational motions.
\item Choose a stereo \emph{baseline} in accordance with the scene depth range. 
Wider baselines are better for stereo, but one must ensure that there is enough overlap between the FOVs.
\item Use \emph{infrared (IR) filters} to mitigate flashes from a motion capture and LiDAR.
\item Remove or limit \emph{flickering light sources} as much as possible, especially artificial lighting at night.
\item Remove \emph{reflections} from the scene were possible.
\item \emph{Noisy events} are more abundant in darker areas than in brighter areas (due to the logarithmic response of the event camera’s photoreceptors), hence if a background color is to be chosen, bright is preferable.
\item \emph{Hardware-synchronize} the sensors whenever possible.
\item Produce data with \emph{aligned timestamps} for sensors so that minimal post-processing is required.
\item \revise{Use rigid \emph{printed fiducials for calibration} since their patterns can be easily reconstructed from events as in \cite{Gehrig21ral, wei24ijrr, peng24arxiv, Carmichael24ijrr, Chaney23cvprw, hadviger23ecmr, chen23tiv}. 
While a blinking pattern on a screen is an alternative, it is less portable for large-scale scenes like with cameras mounted on cars where we need a big pattern to be visible from further away.}
\item Use AprilTags for \emph{camera calibration} since they are detected even when the grid is partially visible (going out of the camera’s FOV), unlike checkerboards.
\item Provide \emph{calibration data and calibration quality results}, so that others may calibrate using alternative tools and/or check the quality of the calibration. 
\item It is important to have quick calibration quality check pipelines. While NN methods like E2VID provide best reconstruction quality, less accurate but faster reconstruction techniques like \emph{simple\_image\_recon}~\cite{Pfrommer22recon} are equally good at detecting high-contrast corners.
\item \emph{Align data} spatially from different sensors as much as possible using calibrated extrinsic parameters and scene~depth.
\item Provide data with different event camera \emph{bias settings}, and report such values.
\item Check \emph{data quality} using state-of-the-art event-based and frame-based SLAM algorithms or other developed tools (visualization, statistical analysis, etc.).
\item Record simple and \emph{increasingly complex motions} that help prototype and debug algorithms.
\item Until an event-based \emph{data format} standard is developed (e.g., JPEG XE), use compressed HDF5 files for efficiently storing events. 
If possible, also provide converted ROS bags that include data from all sensors, for easy compatibility with
existing ROS-based SLAM algorithms (\cref{tab:datasets}).
\end{itemize}

\else
\subsection{Data Recording Recommendations}
\label{sec:datasets:goodpractices}
\rammif{
Surveying existing datasets and given our own experience, we suggest iterating through the steps outlined in \cref{fig:cycle} to produce high-quality data for event stereo research.}  
\rammif{A concrete checklist with recommendations is given in the \textbf{supplementary material}.}
\begin{figure}[t]
    \centering
    \includegraphics[trim={9.6cm 2.3cm 8.2cm 0.77cm},clip,width=.9\linewidth]{images/stereo_event_data_cycle.pdf}
    \caption{\label{fig:cycle}Event stereo data recording cycle.}
    \vspace{-2ex}
\end{figure}

\fi

\ifclearlook\cleardoublepage\fi

\section{Future Research Directions}
\opport{
Let us outline some potential future research directions in event-based stereo depth estimation:
\begin{itemize}[wide, labelwidth=!, labelindent=0pt]
    \item \textbf{Unsupervised learning.}
    While supervised learning methods for depth estimation will continue to improve, 
    investigating better unsupervised/self-supervised methods is desirable to remove the need for GT or auxiliary data (e.g., frames). 
    Explainability is also required for making progress and understanding. 
    \item \textbf{Optimal event representations.} 
    Most of the current methods involve forming ``good-looking'' images that have HDR and low motion-blur properties of events, while trying to populate data in texture-less and motion-less pixels, for feature-based stereo matching. 
    In this type of methods, finding the best such representation (that efficiently exploits sparsity while maximizing accuracy) for feature extraction is an open problem.
    \item \textbf{SNNs on efficient hardware.} 
    To realize the promise of low-power, low-latency edge computing of event cameras, SNNs shall be implemented on efficient neuromorphic hardware that can compete in accuracy with the best performing non-spiking approaches. 
    Promising methods like StereoSpike are currently implemented on GPU. 
    Neuromorphic hardware development needs to catch up, and would benefit from co-design with the algorithm software \cite{Li24ieee}. 
    \item \textbf{Longer temporal context}. 
    Incorporating longer context in deep learning pipelines via camera motion input or recurrent connections will improve accuracy in static scenes (SLAM).
    It will also improve dense depth estimation in regions where few events are generated.
    \item \textbf{Radiance Fields for 3D representations.} 
    Beyond prevalent cost volume voxel grids and Disparity Space Images, implicit representations like Neural Radiance Fields (NeRFs) and explicit representations like 3D Gaussian Splatting and  may be used for encoding the scene during multi-view stereo matching, for efficient long-term fusion and view-agnostic rendering. 
    Recent works have used these representations with event cameras in a monocular setup but with the focus on blur-free, HDR intensity renderings, sometimes in combination with complementary sensors (e.g., traditional cameras).
    \revise{
    \item \textbf{Foundational Models for Event-based Stereo.}
    An unexplored direction in event-based stereo is using modern foundational networks like generative Diffusion models, Vision Language Models and discriminative Vision Transformers.
    While these models have recently shown remarkable results on dense depth estimation from monocular RGB frames \cite{Bochkovskii25iclr}, they are still unexplored for event-based depth estimation.
    Recent work on stereo foundation models \cite{wen25cvpr,izquierdo25cvpr} have shown remarkable results using RGB images, which could also be translated to events.}
    \item \textbf{Joint Estimation.} 
    By combining depth estimation over different observation windows, algorithms can be developed to jointly solve independent motion segmentation and long-term depth estimation for robust SLAM. 
    Many other joint problems that involve depth estimation shall be considered (to improve robustness).
    \item \textbf{Anti-flicker.} 
    To enable wide adaptation of event cameras in night conditions, solutions for anti-flicker are needed, either with explicit filters or intermediate representations that are immune to them.
    \item \textbf{Low-latency algorithms for HD cameras.} 
    Pushing for resolution and efficiency, there is a need to develop real-time algorithms that can handle high event rates from modern megapixel event cameras, by intelligent sub-sampling or intermediate representations.
    \item \textbf{Beyond frame-based benchmarking.} 
    Depth estimation benchmarks need to improve via evaluations beyond synchronous depth/disparity frames and provide high-rate depth evaluations. 
    For example, in M3ED, an API enables depth readout at arbitrary timestamps by projecting point cloud at interpolated camera poses.
    \item \textbf{Instantaneous stereo benchmarks.} 
    There is a need for strong benchmarks for instantaneous stereo involving stationary cameras observing dynamic scenes.
    \item \textbf{Accessible benchmarking and comparisons.} 
    The community will benefit from comprehensive reporting of multiple performance metrics (accuracy, completion, power consumption, latency, etc.) across multiple datasets (and platforms). 
    Making a dataset accessible by providing APIs and data conversion tools also promotes its adoption as a standard benchmark. 
    To aid such standardization, this survey extensively discusses existing datasets (\cref{sec:datasets} and supplementary), collates performance metrics on them (\cref{sec:eval}), and guides building of new stereo benchmarks (\cref{sec:datasets:goodpractices}).
\end{itemize}
}

\section{Conclusion}
\label{sec:conclusion}
In this paper, we have surveyed the current landscape on the topic of event-based stereo depth estimation, providing the most in-depth as well as extensive coverage of the subject to date.
We have traced its journey from the origins in the early 90's, classifying the main approaches and discussing changing trends through the years.
We have comprehensively covered existing algorithms, providing insights into their principles of operations, pros and cons.
We have also compared them empirically using common benchmarks.
Through this analysis, we have identified current leading approaches, performance gaps and future research directions.
To support results-based development and data-driven algorithms, we have extensively surveyed relevant stereo event datasets, as well as provided recommendations for best practices in collecting data and establishing benchmarks.
Stereo event-based depth perception unlocks the potential for low-power, on-device spatial AI in challenging high-speed motion and HDR lighting conditions. 
The literature is growing in this relatively new field, and there are plenty of opportunities to innovate. 
We hope this survey serves as an accessible entry point for newcomers diving into this exciting topic, as well as a practical guide for seasoned experts.

\ifarxiv
\section*{Acknowledgments}
Funded by the SONY Research Award Program 2021; project ``ESSAI''.
Funded by the Deutsche Forschungsgemeinschaft (DFG, German Research Foundation) under Germany’s Excellence Strategy -- EXC 2002/1 ``Science of Intelligence'' -- project number 390523135.
\fi

\ifclearlook\cleardoublepage\fi
{\footnotesize
\ifarxiv
\bibliographystyle{IEEEtran_link}
\else 
\bibliographystyle{IEEEtran}
\fi

}

\vspace{-4ex}
\begin{IEEEbiography}[{\includegraphics[width=1in,height=1.25in,clip,keepaspectratio]{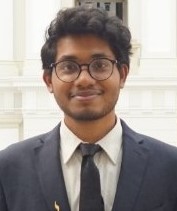}}]{Suman Ghosh} 
is PhD student in the Dept. of EECS %
at Technische Universit\"at Berlin, Germany. 
Previously, he was a Research Fellow %
at Istituto Italiano di Tecnologia, Italy. 
He obtained his dual M.Sc. degree through the Erasmus Mundus Master on Advanced Robotics (EMARO+) program in 2019 from the University of Genoa (Italy) and Warsaw University of Technology (Poland) with full scholarship. 
In 2016, he received a Bachelor's degree in Electronics and Telecommunication Engineering from Jadavpur University, India. 
His research interests include (biological and computer) vision, robotics and embodied intelligence.
\end{IEEEbiography}
\vspace{-4ex}
\begin{IEEEbiography}[{\includegraphics[width=1in,height=1.25in,clip,keepaspectratio]{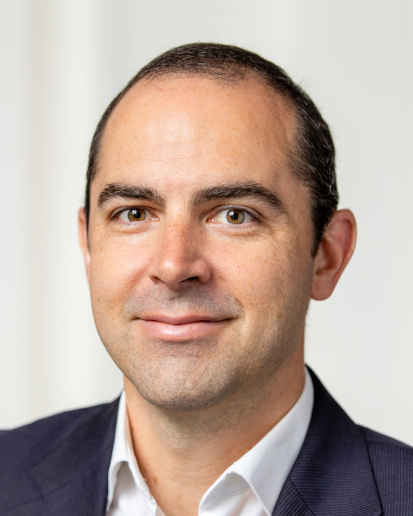}}]{Guillermo Gallego} is Full Professor at Technische Universit\"at Berlin, in the Dept. of EECS, %
and at the Einstein Center Digital Future, %
where he leads the Robotic Interactive Perception Laboratory.
He is also a Principal Investigator at the Science of Intelligence Excellence Cluster. %
and the Robotics Institute Germany.
He received the Ph.D. degree in Electrical and Computer Engineering from the Georgia Institute of Technology, USA, in 2011, supported by a Fulbright Scholarship.
He serves as Associate Editor for IEEE T-PAMI, RA-L and IJRR, and guest Editor for IEEE T-RO.
His research interests include robotics, perception, %
optimization and geometry. 
\end{IEEEbiography}

\vfill

\end{document}